%% file: root.tex
\documentclass[letterpaper, 10 pt, conference]{ieeeconf}  

\IEEEoverridecommandlockouts                              

\usepackage{cite}
\usepackage{amsmath,amssymb,amsfonts}
\usepackage{algorithmic}
\usepackage{textcomp}

\usepackage{enumerate}
\usepackage{amsmath}
\usepackage{amssymb}
\usepackage{algorithm,algorithmic}
\usepackage{graphicx}
\usepackage{upgreek}
\usepackage{bm}
\usepackage{blindtext}
\usepackage{tikz}
\usetikzlibrary{shapes,arrows,positioning,calc}
\usepackage[export]{adjustbox}
\usepackage{dblfloatfix} 
\usepackage{comment}
\usepackage{hyperref}
\usepackage{graphicx}
\usepackage{subcaption}
\usepackage[skins]{tcolorbox}

\newcommand{\vect}[1]{\boldsymbol{\mathbf{#1}}}

\DeclareMathAlphabet{\mathpzc}{OT1}{pzc}{m}{it}

\newcommand{\ie}{{\em i.e.}}
\newcommand{\eg}{{\em e.g.}}
\usepackage{colortbl}
\usepackage{float}

\usepackage{collcell}
\newcommand*{\MinNumber}{0.0}%
\newcommand*{\MidNumber}{50.0} %
\newcommand*{\MaxNumber}{100.0}%
\usepackage{xstring}

\newcommand{\ApplyGradient}[1]{%
     \IfDecimal{#1}{
        \ifdim #1 pt > \MidNumber pt
            \pgfmathsetmacro{\PercentColor}{max(min(100.0*(#1 - \MidNumber)/(\MaxNumber-\MidNumber),100.0),0.00)} %
	    \edef\x{\noexpand\cellcolor{green!\PercentColor}}\x\textcolor{black}{#1}
        \else
            \pgfmathsetmacro{\PercentColor}{max(min(100.0* (\MidNumber - #1)/(\MidNumber-\MinNumber),100.0),0.00)} %
            \edef\x{\noexpand\cellcolor{red!\PercentColor}}\x\textcolor{black}{#1}
        \fi
        }
        {
        #1
        }
}

\newcolumntype{R}{>{\collectcell\ApplyGradient}c<{\endcollectcell}}

\usepackage{flushend}
\usepackage{cite}
\def\check{\pmb{\tikz\fill[scale=0.3](0,.35) -- (.25,0) -- (1,.7) -- (.25,.15) -- cycle;}}

\usepackage{booktabs}

\title{\LARGE \bf 
FlightGoggles: A Modular Framework for \\  Photorealistic Camera, Exteroceptive Sensor, and Dynamics Simulation
}

\author{Winter Guerra, Ezra Tal, Varun Murali, Gilhyun Ryou and Sertac Karaman
\thanks{All authors are with the Laboratory for Information and Decision Systems (LIDS), Massachusetts Institute of Technology. 
{\tt\footnotesize \{winterg, eatal, mvarun, ghryou, sertac\}@mit.edu}}%
}

\begin{document}

\maketitle
\thispagestyle{empty}
\pagestyle{empty}

\begin{abstract}
FlightGoggles is a photorealistic sensor simulator for perception-driven robotic vehicles.
The key contributions of FlightGoggles are twofold.
First, FlightGoggles provides photorealistic exteroceptive sensor simulation using graphics assets generated with photogrammetry. 
Second, it provides the ability to combine {\em (i)} synthetic exteroceptive measurements generated {\em in silico} in real time and {\em (ii)} vehicle dynamics and proprioceptive measurements generated {\em in motio} by vehicle(s) in a motion-capture facility. 
FlightGoggles is capable of simulating a virtual-reality environment around autonomous vehicle(s).
While a vehicle is in flight in the FlightGoggles virtual reality environment, exteroceptive sensors are rendered synthetically in real time while all complex extrinsic dynamics are generated organically through the natural interactions of the vehicle. 
The FlightGoggles framework allows for researchers to accelerate development by circumventing the need to estimate complex and hard-to-model interactions such as aerodynamics, motor mechanics, battery electrochemistry, and behavior of other agents. 
The ability to perform vehicle-in-the-loop experiments with photorealistic exteroceptive sensor simulation facilitates novel research directions involving, \eg, fast and agile autonomous flight in obstacle-rich environments, safe human interaction, and flexible sensor selection.
FlightGoggles has been utilized as the main test for selecting nine teams that will advance in the AlphaPilot autonomous drone racing challenge. 
We survey approaches and results from the top AlphaPilot teams, which may be of independent interest.
\end{abstract}

\section*{SUPPLEMENT: SOFTWARE, ASSETS, \& VIDEOS}
FlightGoggles is distributed as open-source software along with the photorealistic graphics assets for several simulation environments, under the MIT license. Software, graphics assets, and videos showcasing FlightGoggles as well as videos showcasing the results of the AlphaPilot simulation challenge and recent work using FlightGoggles can be found at \url{http://flightgoggles.mit.edu}.


\input{intro.tex} %

\input{sysarch.tex} %

\input{photogrammetry.tex} %

\input{visualrendering.tex} %

\input{vehicledynamics.tex} %

\input{experiments.tex}

\input{alphapilot.tex}

\section{CONCLUSIONS}
\label{sec:conclusions}
This paper introduced FlightGoggles, a new modular framework for realistic simulation to aid robotics testing and development. 
FlightGoggles is enabled by photogrammetry and virtual reality technologies. 
Heavy utilization of photogrammetry helps provide realistic simulation of camera sensors. 
Utilization of virtual reality allows direct integration of real vehicle motion and human behavior acquired in motion capture facilities directly into the simulation system. 
FlightGoggles is being actively utilized by a community of robotics researchers. 
In particular, FlightGoggles has served as the main test for selecting the contestants for the AlphaPilot autonomous drone racing challenge. 
This paper also presented a survey of approaches and results from the simulation challenge, which may be of independent interest.

\bibliographystyle{IEEEtran} 

\bibliography{references}

\end{document}

%% file: intro.tex

\section{INTRODUCTION}

Simulation systems have long been an integral part of the development of robotic vehicles. 
They allow engineers to identify errors early on in the development process, and allow researchers to rapidly prototype and demonstrate their ideas. 
Despite their success in accelerating development, many researchers view results generated in simulation systems with skepticism,
as any simulation system is some abstraction of reality and will disagree with reality at some scale. 
\begin{figure}[H]
	\includegraphics[width=\columnwidth]{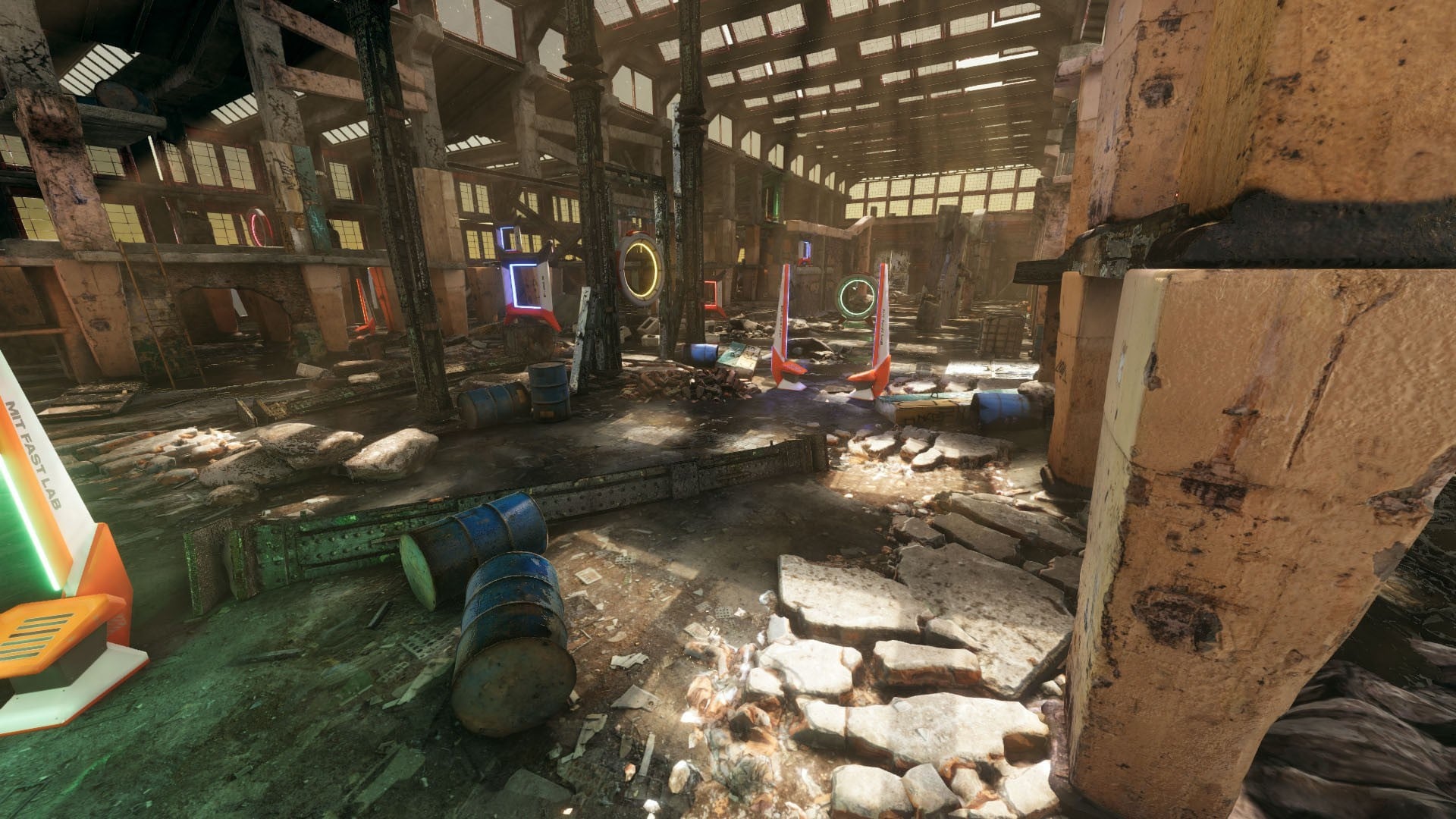} \\[0.3em]
	\includegraphics[width=\columnwidth]{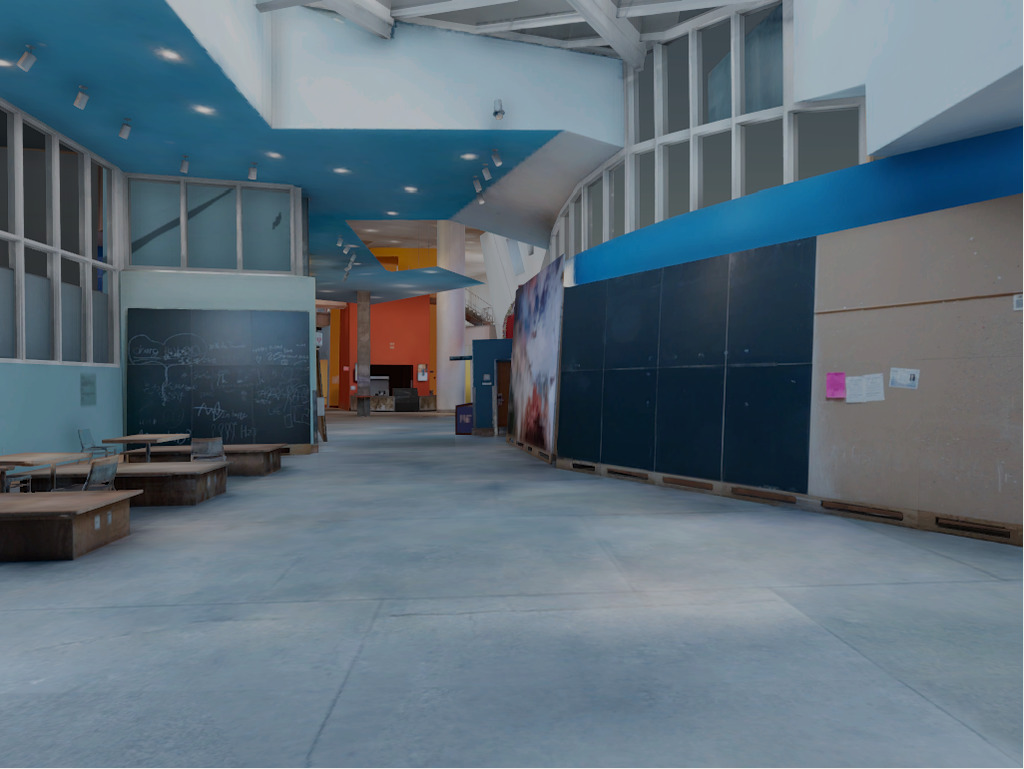} \\[0.3em]
	\includegraphics[width=\columnwidth]{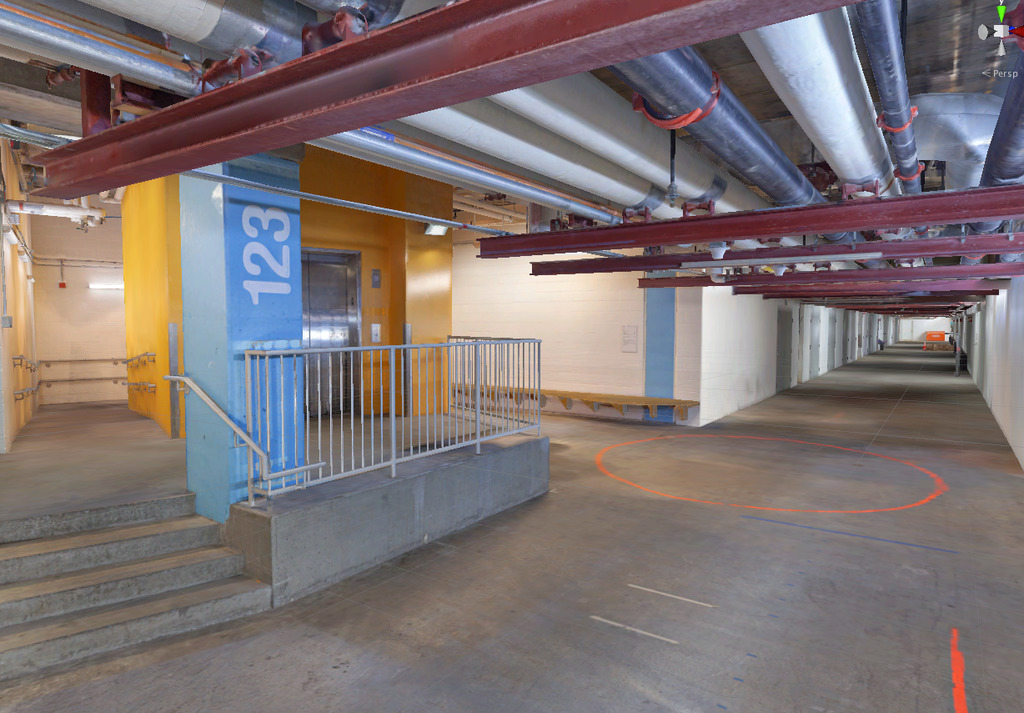}
	\caption{FlightGoggles renderings of the available environment. The top picture shows the \textit{Abandoned Factory} environment and the \textit{Ground floor} and \textit{Basement} of the Ray and Maria Stata Center at MIT in the middle and bottom respectively.}
	\label{fig:unity3d}
\end{figure}
This skepticism towards results generated exclusively in simulations studies is exemplified by Rodney Brooks' well-known quote from 1993: ``[experiment] simulations are doomed to succeed ... [because] simulations cannot be made sufficiently realistic''\cite{brooks1993real}.

Despite the skepticism towards simulation results, several trends have emerged in recent years that have driven the research community to develop better simulation systems out of necessity. 
A major driving trend towards realistic simulators stems from the emergence of data-driven algorithmic methods in robotics, for instance, based on machine learning methods that require extensive data.
Simulation systems provide not only massive amounts of data, but also the labels required for training algorithms.
For example, simulation systems can provide a safe environment for reinforcement learning methods~\cite{tan2018sim}. 
This driving trend has posed a critical need to develop better, more realistic simulation systems.

Several enabling trends have also recently emerged that allow for better, more-realistic simulation systems to be developed.
The first enabling trend is the development of new computing resources that enable realistic rendering.
The rapid evolution of game engine technology, particularly 3D graphics rendering engines,
has made available advanced features such as improved material characteristics, real-time reflections,
volumetric lighting, and advanced illumination through deferred rendering pipelines.
Particularly, the maturation of off-the-shelf software packages such as Unreal Engine~\cite{unreal} and Unity~\cite{unity3d},
makes them suitable for high-fidelity rendering in applications beyond video games, such as robotics simulation.
Simultaneously, next-generation graphics processors simply pack more transistors, and the transistors are better organized for rendering purposes,
\eg, for real-time ray tracing. In addition, they incorporate computation cores that utilize machine learning, for instance, trained with pictures of real environments to generate realistic renderings~\cite{nvidia-turing}.
This trend is an opportunity to utilize better software and hardware to realize realistic sensor simulations.
The second enabling trend stems from the proliferation of motion capture facilities for robotics research, enabling precise tracking of robotic vehicles and humans through various technologies, such as infrared cameras, laser tracking, and ultra-wide band radio.
These facilities provide the opportunity to incorporate real motion and behavior of vehicles and humans into the simulation in real time.  
This trend provides the potential to combine the efficiency, safety, and flexibility of simulation with real-world physics and agent behavior.

Traditionally, simulation systems embody ``models'' of the vehicles and the environment,
which are used to emulate what the vehicles sense, how they move, and how their environment adapts. 
In this paper, we present two concepts that use ``data'' to drive realistic simulations. 
First, we heavily utilize {\em photogrammetry} to realistically simulate exteroceptive sensors.
For this purpose, we photograph real-world objects, and reconstruct them in the simulation environment.
Almost all objects in our simulation environment are, in fact, a rendering of a real-world object.
This approach allows realistic renderings, as shown in Figure~\ref{fig:unity3d}.
Second, we utilize a novel {\em virtual-reality system} to realistically embed inertial sensors, vehicles dynamics, and human behavior into the simulation environment.
Instead of modeling these effects, we place vehicles and human actors in motion-capture facilities. We acquire the pose of the vehicles and the configuration of the human actors in real time, and create their avatars in the simulation environment. 
For each autonomous vehicle, its proprioceptive measurements are acquired using on-board sensors, \eg, inertial measurement units and odometers; while exteroceptive sensors are rendered photorealistically in real time. 
In addition, the human behavior observed by the vehicles is generated by humans reacting to the simulation. 
In other words, vehicles embedded in the FlightGoggles simulation system experience real dynamics, real inertial sensing, real human behavior, and synthetic exteroceptive sensor measurements rendered photorealistically effectively by transforming photographs of real-world objects. 

The combination of real physics and data-driven exteroceptive sensor simulation that FlightGoggles provides is not achieved in traditional simulation systems. 
Such systems are typically built around a physics engine that simulates vehicles and the environment based on a ``model'', most commonly a system of ordinary or partial differential equations.
While these models may accurately exemplify the behavior of a general vehicle or actor, this is not sufficient to ensure that simulation results transfer to the real world.
Complicated aspects of vehicle dynamics, \eg, vibrations and unsteady aerodynamics, and of human behavior may significantly affect results,
but can be very challenging to accurately capture in a physics model.
For an overview of popular physics engines, the reader is referred to \cite{erez2015simulation}.
In order to generate exteroceptive sensor data, robotics simulators employ a graphics rendering engine in conjunction with the physics engine. 
A popular example is Gazebo \cite{koenig2004design}, which lets users select various underlying engines.
It is often used in combination with the Robot Operating System (ROS) to enable hardware-in-the-loop simulation.
However, Gazebo is generally not capable of photorealistic rendering.
Specifically, for unmanned aerial vehicles simulation, two popular simulators that are built on Gazebo are the Hector Quadrotor package~\cite{meyer2012comprehensive} and RotorS~\cite{furrer2016rotors}.
Both simulators include vehicle dynamics and exteroceptive sensor models, but lack the capability to render photorealistic camera streams.
Habitat AI \cite{habitat19iccv} and AI2-Thors \cite{kolve2017ai2} generate photorealistic environments using photogrammetry, but target the indoor mobile robot which has a simple dynamics model.
AirSim, on the other hand, is purposely built on the Unreal rendering engine to enable rendering of photorealistic camera streams from autonomous vehicles,
but still suffers from the shortcomings of typical physics engines when it comes to vehicle dynamics and inertial measurements~\cite{shah2018airsim, ratnesh2020airsim}.
Flightmare \cite{song2020flightmare} enables users to flexibly use custom physics engines or real-world flight data to overcome this shortcoming.

The rise of data-driven algorithms for autonomous robotics, has bolstered the need for extensive labeled data sets.
Simulation offers an alternative to experimental data gathering.
Clearly, there are many advantages to this approach, \eg, cost efficiency, safety, repeatability, and essentially unlimited quantity and diversity.
In recent years, several synthetic, or virtual, datasets have appeared in literature.
For example, Synthia~\cite{ros2016synthia} and Virtual KITTI~\cite{Gaidon:Virtual:CVPR2016} use Unity to generate photorealistic renders of an urban environment while \cite{wang2020tartan} use the Unreal Engine to generate data for visual inertial odometry.
Similiarly, generating high-fidelity images using a rendering engine is used in various visual SLAM datasets such as \cite{straub2019replica, wang2020tartan}.
The 3D images or point cloud data are often used to enhance the quality of the rendered environments. 
For instance, Matterport3d~\cite{chang2017matterport3d} utilizes the Matterport 3D camera sensor~\cite{matterport}, and Stanford BuildingParser~\cite{armeni20163d} utilizes the large-scale point cloud to generate fine-grained dataset.
ICL-NUIM~\cite{handa:etal:ICRA2014} provides synthetic renderings of an indoor environment based on pre-recorded handheld trajectories.
The Blackbird Dataset~\cite{antonini2018blackbird, blackbird2020ijrr} includes real-world ground truth and inertial measurements of a quadcopter in motion capture,
and photorealistic camera imagery rendered in FlightGoggles.
The open-source availability of FlightGoggles and its photorealistic assets enables users to straightforwardly generate
additional data, including real-time photorealistic renders based on real-world vehicles and actors.

This paper is an extension of \cite{guerra2019FGIROS}. The extensions presented in this paper are
\begin{itemize}
\item Two new environments: \textit{Ground Floor of the Stata Center} and the \textit{Basement of the Stata Center}
\item New camera types \ie depth, segmentation, surface normals. 
\item A python API to facilitate offline processing and machine learning applications.
\end{itemize}

This paper is organized as follows. Section \ref{sec:sysarch} provides an overview of the FlightGoggles system architecture, including interfacing with real-world vehicles and actors in motion capture facilities.
Section \ref{sec:exteroceptive} outlines the photogrammetry process and the resulting virtual environment. This section also details the rendering engine and the exteroceptive sensor models available.
Section \ref{sec:sim_vehicles} describes a physics engine for simulation of multicopter dynamics and inertial measurements.
Section \ref{sec:useCases} describes several applications of FlightGoggles, including results of the AlphaPilot qualifications. 
Finally, Section \ref{sec:conclusions} concludes with remarks.

%% file: sysarch.tex

\section{SYSTEM ARCHITECTURE}
\label{sec:sysarch}

\input{sysoverview.tex}

FlightGoggles is based on a modular architecture, as shown in Figure \ref{fig:sysarch}.
This architecture provides the flexibility to tailor functionality for a specific simulation scenario involving real and/or simulated vehicles, and possibly human interactors.
As shown in the figure, FlightGoggles' central component is the Unity game engine.
It utilizes position and orientation information
to simulate camera imagery and exteroceptive sensors, and to detect collisions. Collision checks are performed using
polygon colliders, and results are output to be included in the dynamics of simulated vehicles.

A description of the dynamics and measurement model used for multicopter simulation is given in Section \ref{sec:sim_vehicles}.
Additionally, FlightGoggles includes a simulation base class that can be used to
simulate user-defined vehicle equations of motion, and measurement models.
Simulation scenarios may also include real-world vehicles through the use of a motion capture system.
In this case, Unity simulation of camera images and exteroceptive sensors, and collision detection are based on the real-world vehicle position and orientation.
This type of {\em vehicle-in-the-loop simulation} can be seen as an extension of customary hardware-in-the-loop configurations.
It not only includes the vehicle hardware, but also the actual physics of processes that are challenging to simulate accurately, such as aerodynamics (including effects of turbulent air flows), and inertial measurements subject to vehicle vibrations.
FlightGoggles provides the novel combination of real-life vehicle dynamics and proprioceptive measurements, and simulated photorealistic exteroceptive sensor simulation.
It allows for real-life physics, flexible exteroceptive sensor configurations, and obstacle-rich environments without the risk of actual collisions.
FlightGoggles also allows scenarios involving both humans and vehicles, colocated in simulation but placed in different motion capture rooms, \eg, for safety.

Dynamics states, control inputs, and sensor outputs of real and simulated vehicles, and human interactors are available to the user through the FlightGoggles API.
In order to enable message passing between FlightGoggles nodes and the API, the framework can be used with either ROS \cite{quigley2009ros} or LCM \cite{huang2010lcm}. The FlightGoggles simulator can be run headlessly on an Amazon Web Services (AWS) cloud instance to enable real-time simulation on systems with limited hardware. 

Dynamic elements, such as moving obstacles, lights, vehicles, and human actors, can be added and animated in the environment in real-time.
Using these added elements, users can change environment lighting or simulate complicated human-vehicle, vehicle-vehicle, and vehicle-object interactions in the virtual environment.

In Section \ref{sec:useCases}, we describe an use case involving a dynamic human actor. In this scenario, skeleton tracking motion capture data is used to render a 3D model of the human in the virtual FlightGoggles environment. The resulting render is observed in real-time by a virtual camera attached to a quadcopter in real-life flight in a different motion capture room (see Figure \ref{fig:human_render}).

\subsection{Cloud Simulation}
\label{sec:cloud_simulation}

FlightGoggles can be operated with multiple machines or cloud environments.
In order to obtain photorealistic camera images with high frequency, FlightGoggles often requires high computational resources.
Such computational cost can be achieved by parallel operation of computational resources, e.g. computing machines with multiple GPU or headless servers such as AWS.
The tasks of rendering each camera and updating the vehicles’ dynamics model can be manually assigned to each computational resource.
This parallelization of computation enables FlightGoggles to simulate multiple vehicles with high-resolution camera sensors without significant latency.
The manual to setup cloud simulation environments is included in our confluence page.

\subsection{Python Interface}
\label{sec:python_api}
FlightGoggles provides Python API for applications such as deep learning and offline testing and tuning of vision algorithms.
The high-fidelity simulation has been required in the field of machine learning due to the recent success in deep reinforcement learning.
In order to provide convenient integration with existing deep learning framework, e.g. Pytorch or Tensorflow, we implement Python wrappers for all components of FlightGoggles libraries, including the dynamics models and the rendering client.
All scenarios including simulating multiple vehicles or running cloud simulation also can be used with Python API.

Besides utilizing FlightGoggles in machine learning research, we also expect that the Python API can provide a convenient environment for virtual robotics experiments.
The vision algorithms such as visual odometry can be evaluated and tuned on FlightGoggles before operating on the real-world environment.
Moreover, based on FlightGoggles Python API, the online robotics experiments are designed for MIT lecture of autonomous racing cars.

\section{MULTIPLE VEHICLE SIMULATION}
\label{sec:multiple_vehicle_simulation}
FlightGoggles provides a parallel simulation of multiple vehicles. 
Recently, many research has been conducted in the field of swarm navigations which focuses on coordinating multiple vehicles to achieve complicated robotics tasks.
On the other hand, research on non-cooperative motion planning has been conducted which can be applied to racing games.
FlightGoggles is able to operate these experiments online by simulating multiple vehicles with corresponding sensors such as inertial measurement units and cameras.
We provide the dynamics simulation of a dubins car and a multicopter aircraft, and all physical parameters for each vehicle are configurable.

\begin{figure*}
  \centering
  \includegraphics[width=\textwidth]{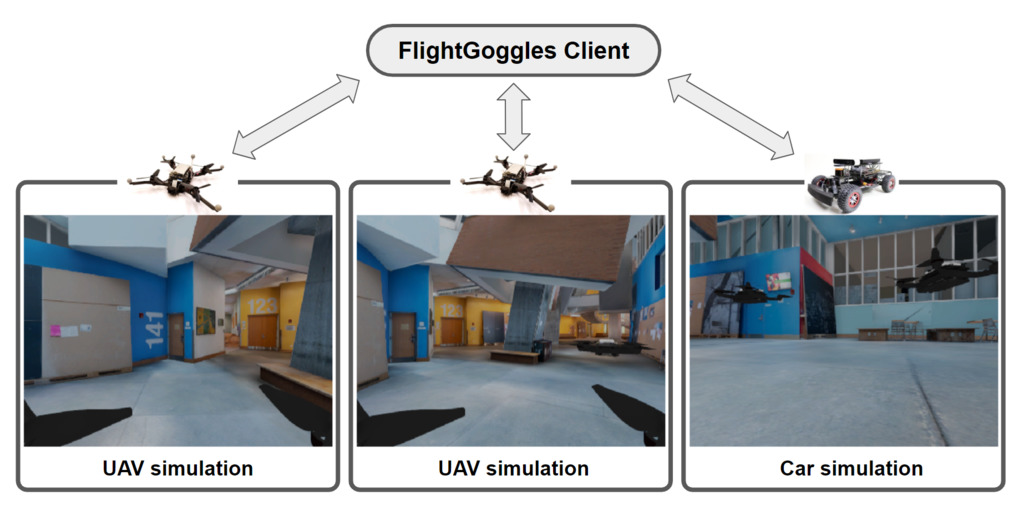}
	\label{fig:multi_vehicle}
	\caption{Camera images of multiple vehicles can be parallelly rendered.}
\end{figure*}

%% file: sysoverview.tex

\tikzstyle{smallrect} = [rectangle, rounded corners = 5, minimum height=6em, minimum width = 7.5em, font = \footnotesize, draw, text depth = -8em]

\tikzstyle{largerect} = [rectangle, rounded corners = 5, minimum height=20em, minimum width = 26.5em, font = \small, text depth = 17 em, draw]

\tikzstyle{medrect} = [rectangle, rounded corners = 5, minimum height=11.5em, minimum width = 18em, font = \small, text depth = 8.5 em, draw]

\tikzstyle{largenarrowrect} = [rectangle, rounded corners = 5, minimum height=11.5em, minimum width = 9.5em, font = \small, text depth = 7.5 em, align = center, draw]

\tikzstyle{largenarrowrect2} = [rectangle, rounded corners = 5, minimum height=11.5em, minimum width = 9.5em, font = \small, text depth = 8.5 em, align = center, draw]

\tikzstyle{fullwidthrect} = [rectangle, rounded corners = 5, minimum height=25em, minimum width = 0.95\linewidth, draw, font = \LARGE, text depth = 22 em]
\tikzstyle{fullwidthrect2} = [rectangle, rounded corners = 5, minimum width = 0.95\linewidth, draw, font = \LARGE]

\tikzstyle{smalltext} = [font = \footnotesize]
\tikzstyle{medtext} = [font = \Large]
\tikzstyle{headertext} = [font = \LARGE]

\tikzstyle{mypoint} = [coordinate]

\definecolor{myblue}{rgb}{0.81, 0.87, 1.0}
\definecolor{mygreen}{rgb}{0.7, 0.93, 0.36}
\definecolor{myred}{rgb}{1.0, 0.71, 0.76}
\definecolor{myyellow}{rgb}{0.98, 0.93, 0.37}
\definecolor{myorange}{rgb}{0.95,0.95,0.95}

\begin{figure*}
	\centering
	\footnotesize
	{\begin{tikzpicture}[auto, node distance=6em]
		
	\node[fullwidthrect,fill=myblue](Renderer){FlightGoggles Renderer};
	\node[largerect,below of=Renderer,node distance = 1.25 em,fill=myorange](Camera){Camera Simulation};
	\node[largenarrowrect,below left= -21.25 em and -13.25 em of Renderer,fill=myorange](Beacons){Beacons and \\ Range Sensors};
	\node[largenarrowrect,below right= -21.25 em and -13.25 em of Renderer,fill=myorange](Collisions){Collision \\ Detection};
	
	\node[smallrect, above right = -9em and -8.5em of Camera, fill overzoom image*={clip,trim=0em 0em 0em 0em}{images/depth}](Cam3){};
	\node[smalltext, below of=Cam3, node distance = 4 em](Cam3Text){Depth Map};	
	\node[smallrect, above right = -9em and -17em of Camera, fill overzoom image*={clip,trim=0em 0em 0em 0em}{images/gray}](Cam2){};
	\node[smalltext, below of=Cam2, node distance = 4 em](Cam2Text){Grayscale Video};
	\node[smallrect, above left = -9em and -8.5em of Camera, fill overzoom image*={clip,trim=0em 0em 0em 0em}{images/RGB}](Cam1){};
	\node[smalltext, below of=Cam1, node distance = 4 em](Cam1Text){RGB Video};
	
	\node[smallrect, above right = -18em and -8.5em of Camera, fill overzoom image*={clip,trim=0em 0em 0em 0em}{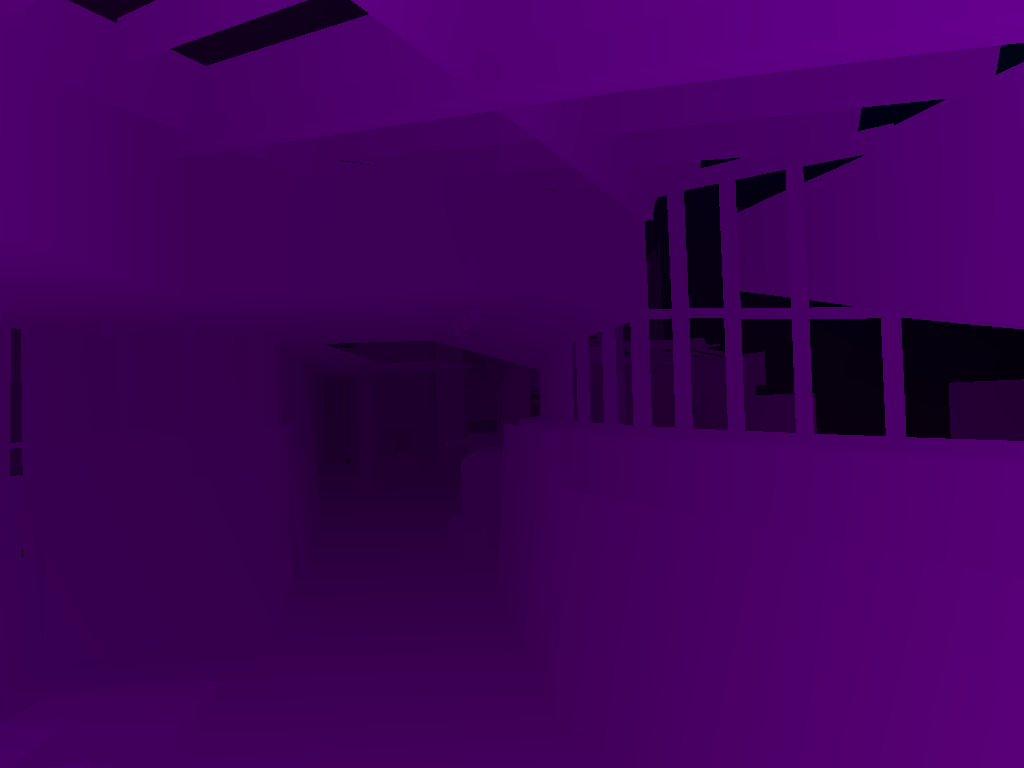}](Cam6){};
		\node[smalltext, below of=Cam6, node distance = 4 em](Cam6Text){Optical Flow};		
	\node[smallrect, above right = -18em and -17em of Camera, fill overzoom image*={clip,trim=0em 0em 0em 0em}{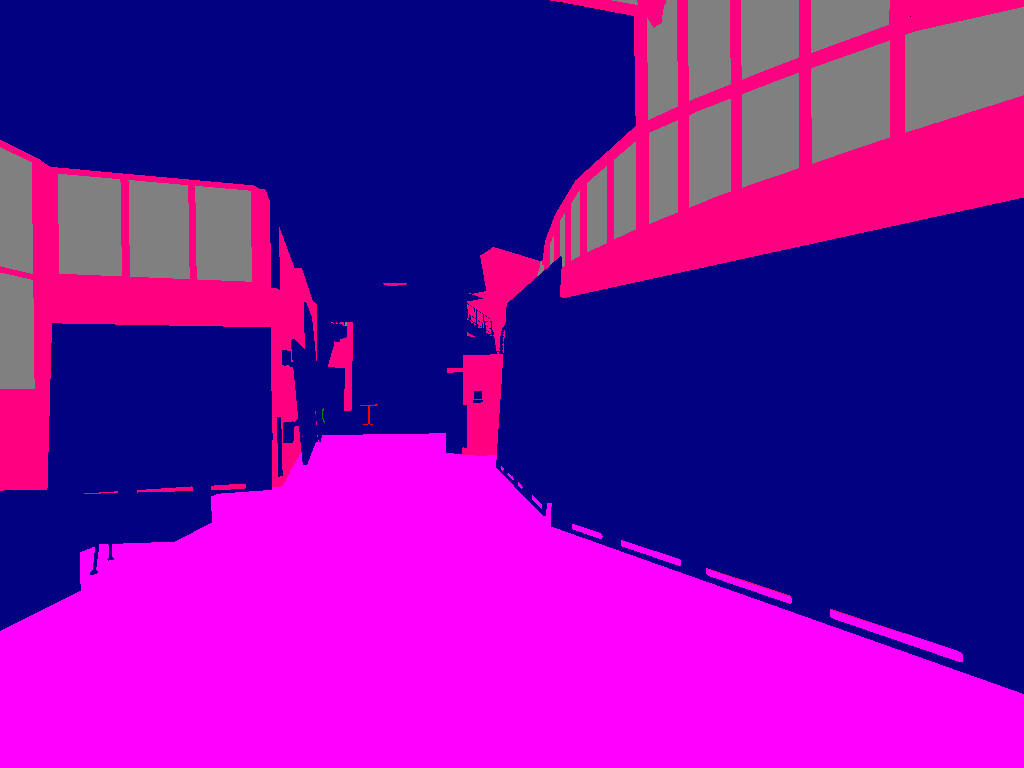}](Cam5){};
		\node[smalltext, below of=Cam5, node distance = 4 em,align=center](Cam5Text){Semantic Segmentation};	
	\node[smallrect, above left = -18em and -8.5em of Camera, fill overzoom image*={clip,trim=0em 0em 0em 0em}{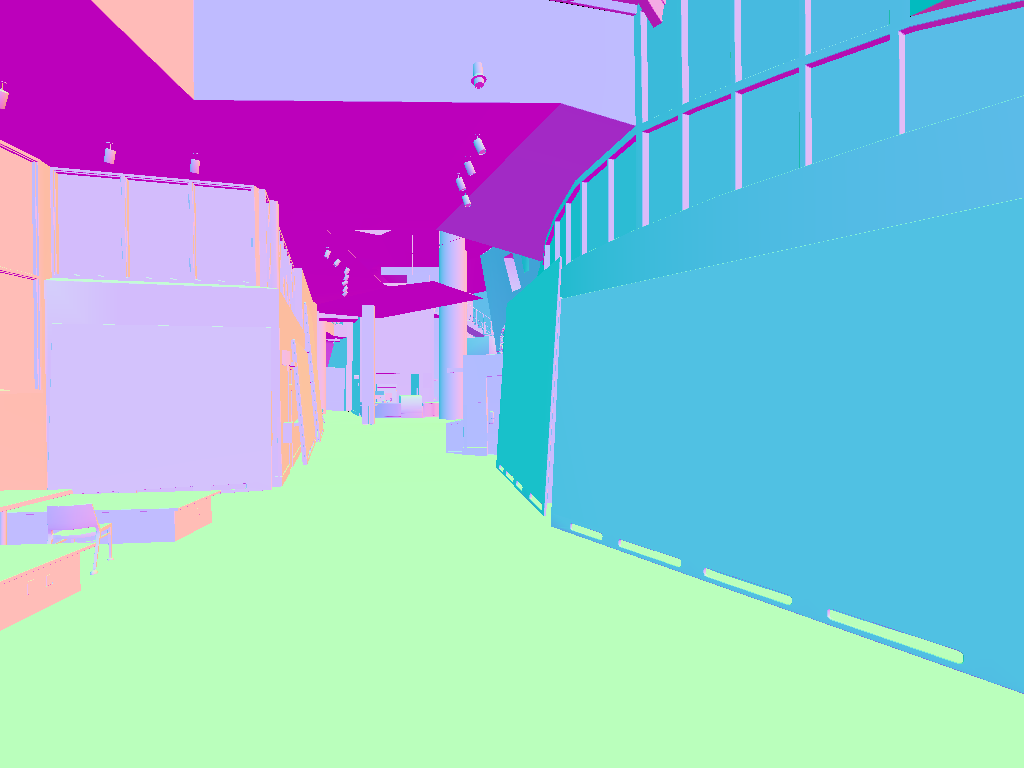}](Cam4){};
	\node[smalltext, below of=Cam4, node distance = 4 em](Cam4Text){Surface Normals};	
	
	\node[smallrect, below of= Beacons, node distance = 1.5 em, fill overzoom image*={clip,trim=0 0 25em 0}{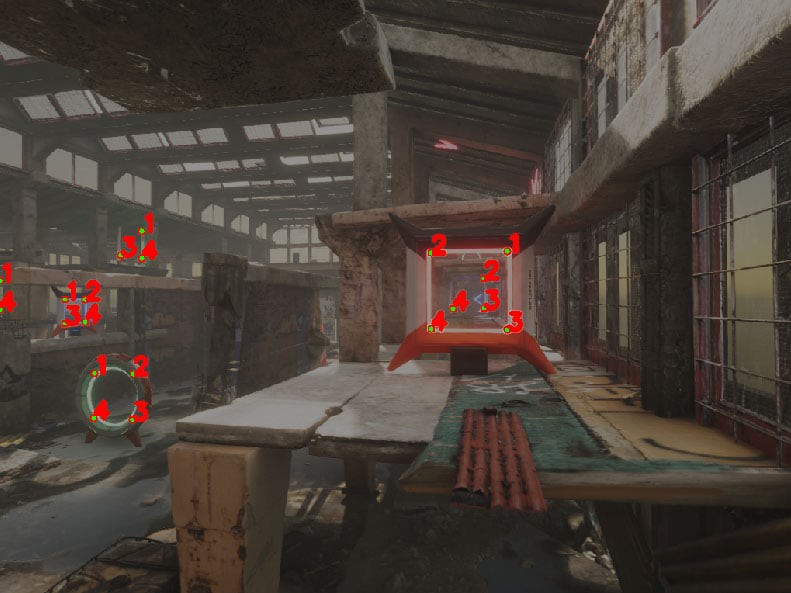}] (Beaconsinner){};
	\node[smallrect, below of= Collisions, node distance = 1.5 em, fill overzoom image*={clip,trim=0em 0em 0em 0em}{images/collision}] (Collisionsinner){};
	
	\node[fullwidthrect2, below of=Renderer, node distance = 25 em, minimum height = 23.25 em,fill=myblue](Client){};
	
	\node[headertext, above of=Client, node distance =10 em](clienttitle){FlightGoggles Client};
	
	\node[mypoint,above right= -25 em and 0 em of Client](Clientpr){};
	\node[mypoint,above left= -25 em and 0 em of Client](Clientpl){};
	\node[mypoint,below right = -17.5 em and -0.5em of Clientpr](rtop-pyapi){};
	\node[mypoint,below left = -18.25 em and -0.5em of Clientpl](ltop-rosapi){};

	\draw[rounded corners=5,fill=myyellow](ltop-rosapi) |- ++(48.5em,3em) -- ++(0,-14.25em) -- ++(-38em,0) -- ++(0,-4em) -| (ltop-rosapi);
	\draw[rounded corners=5,fill=mygreen,opacity=0.5](rtop-pyapi) |- ++(-30em,3em) -- ++(0,-12.75em) -- ++(18em,0) -- ++(0,-4.75em) -| (rtop-pyapi);
	
	\node[medtext,below right= -5.5 em and 1.5 em of Clientpl](rostitle){ROS API};
	\node[medtext,below left= -5.5 em and 1.5 em of Clientpr](pytitle){Python API};
	\node[largenarrowrect2,below left= -20. em and -29 em of Clientpl,fill=myorange](rtinputs){Real-Time Inputs};
	\node[largenarrowrect2,below right= -20. em and -10.5 em of Clientpr,fill=myorange](pbindings){Python Bindings};
	\node[medrect,below left = -20.em and -48.25em of Clientpl,fill=myorange](libs){C++ Simulation Libraries};
	\node[medrect,below left = -20.em and -19em of Clientpl,fill=myorange](mocap){Motion Capture System(s)};

	\node[smallrect, below of= rtinputs, node distance = 0.25 em, fill overzoom image*={clip,trim=0em 0em 0em 0em}{images/realtimeinputs}] (rtinputsinner){};
	\node[smallrect, below of= pbindings, node distance = 0.25 em, fill overzoom image*={clip,trim=0em 10em 0em 0em}{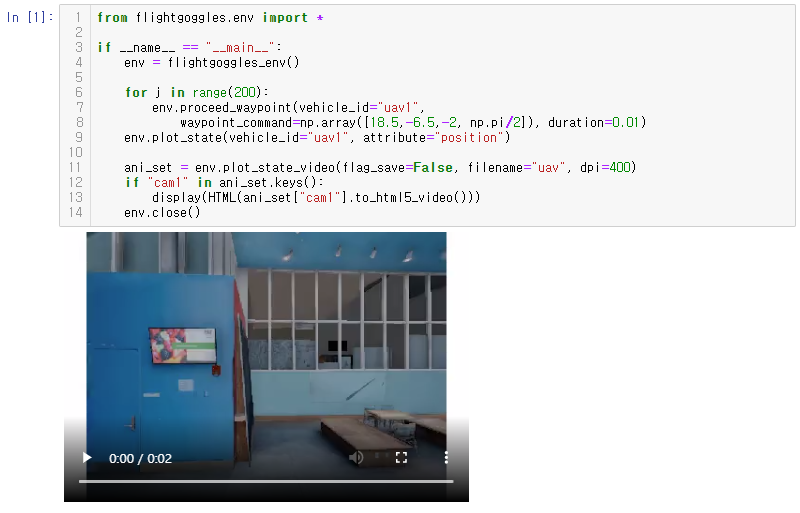}] (pbindingsinner){};
	
	\node[smallrect, above left = -9em and -8.5em of libs, fill overzoom image*={clip,trim=10em 0em 25em 5em}{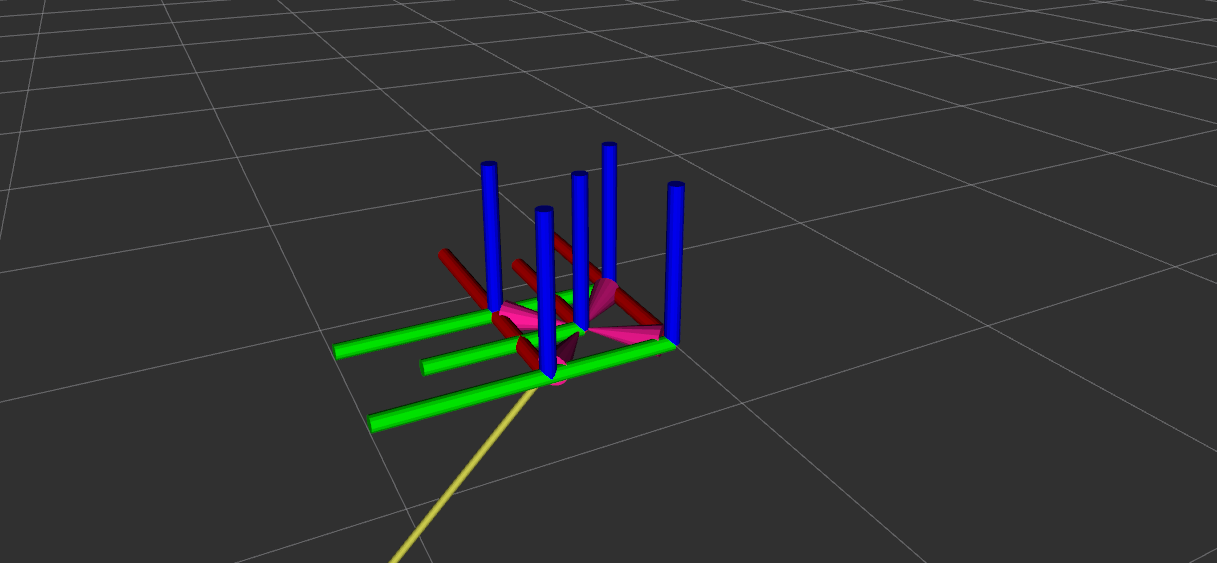}](lib1){};
	\node[smalltext, below of=lib1, node distance = 4 em,align=center,](lib1Text){Vehicle Dynamics};	
	\node[smallrect, above right = -9em and -8.5em of libs, fill overzoom image*={clip,trim=0em 0em 0em 0em}{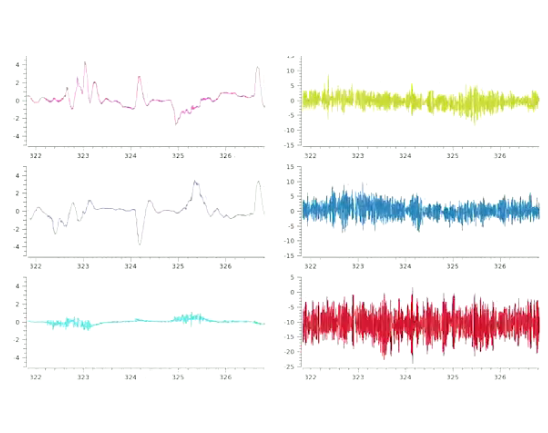}](lib2){};
	\node[smalltext, below of=lib2, node distance = 4 em](lib2Text){IMU Measurements};
	
	\node[smallrect, above left = -9em and -8.5em of mocap, fill overzoom image*={clip,trim=0em 0em 0em 0em}{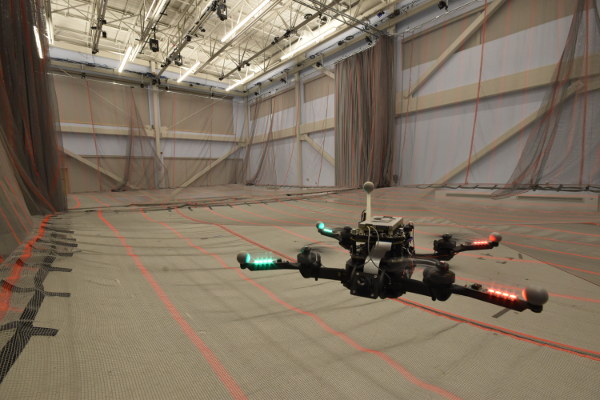}](mocap1){};
	\node[smalltext, below of=mocap1, node distance = 4 em](mocap1Text){Vehicle in Flight};	
	\node[smallrect, above right = -9em and -8.5em of mocap, fill overzoom image*={clip,trim=0em 0em 0em 0em}{images/humaninteractor}](mocap2){};
	\node[smalltext, below of=mocap2, node distance = 4 em](mocap2Text){Human Interactor};

	\draw [latex'-latex',double,thick,font=\large] (rtop-pyapi)++(-4em,13em) -- node [left,pos=0.4]{ZeroMQ} 
	++ (0,-8em); 
	
	\node[smalltext,  above left = 9em and -14.5em of ltop-rosapi, align = left](renderdescr){One or multiple Renderer\\ instances run on (cloud-based)\\ Windows or Ubuntu systems.};
	
	\node[smalltext,  above left = -15em and -36.5em of ltop-rosapi, align = left](renderdescr){One or multiple Client instances run on Ubuntu systems,\\ each using the ROS or Python API.};	
		
	\end{tikzpicture}}
	\caption{Overview of FlightGoggles system architecture. The Client(s) send pose data of human interactors, and real and simulated vehicles to the Renderer(s), and receive simulated camera imagery, collision detections, and beacon and range measurements.} \label{fig:sysarch}
\end{figure*}

%% file: photogrammetry.tex
\section{PHOTOGRAMMETRIC ASSET CREATION}
\label{sec:photogrammetry}

\begin{figure*}[tbp]
	\centering
	\begin{subfigure}[b]{\textwidth}
		\includegraphics[width=\textwidth, trim=0 0 0 10.5cm, clip]{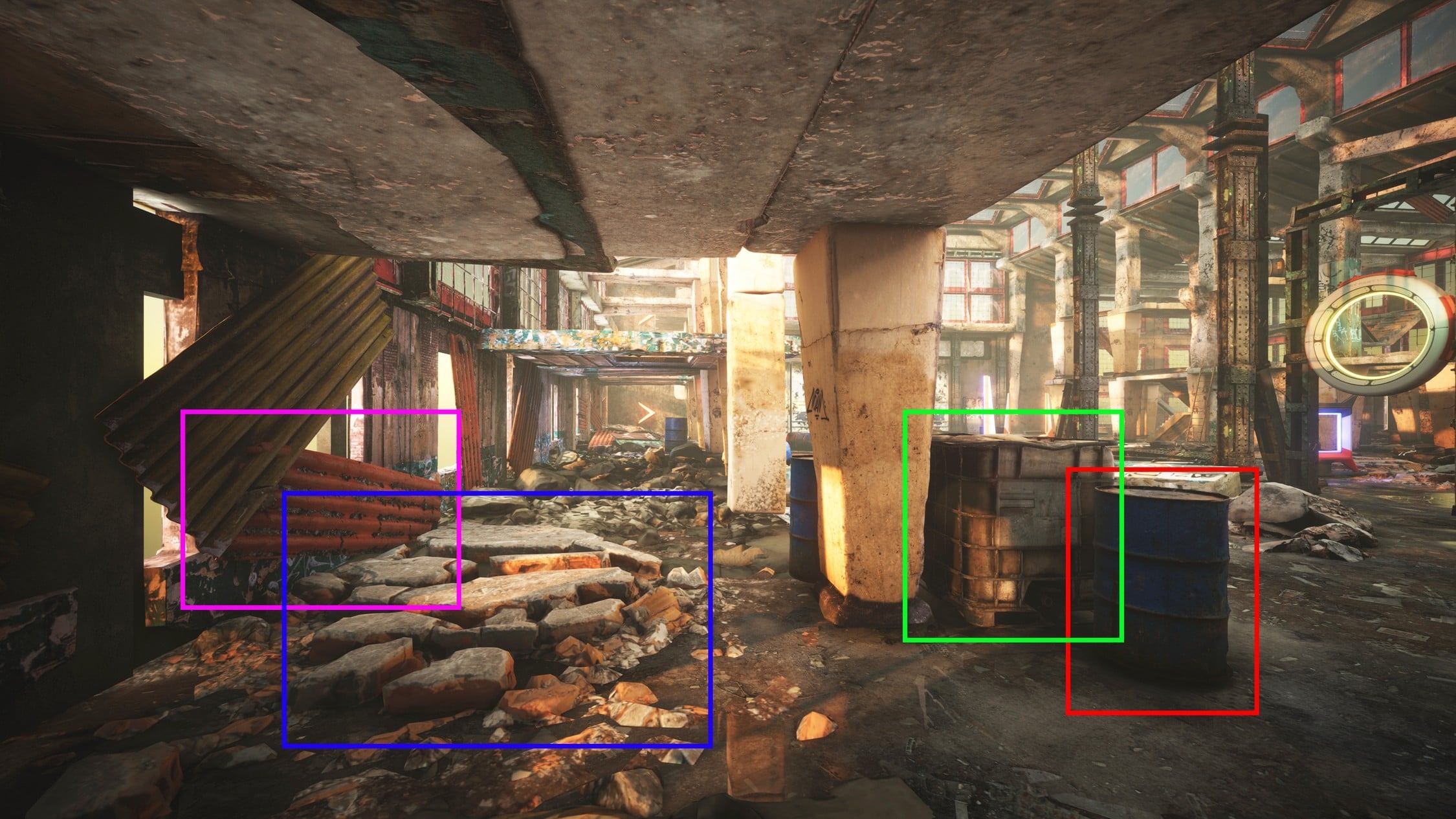}
		\caption{Virtual environment in FlightGoggles with barrel (red), rubble (blue), corrugated metal (magenta), and caged tank (green).}
		\label{fig:envasset_overview}
	\end{subfigure}
	\vspace{0.01in}

	\begin{subfigure}[b]{0.235\textwidth}
		\includegraphics[width=\textwidth]{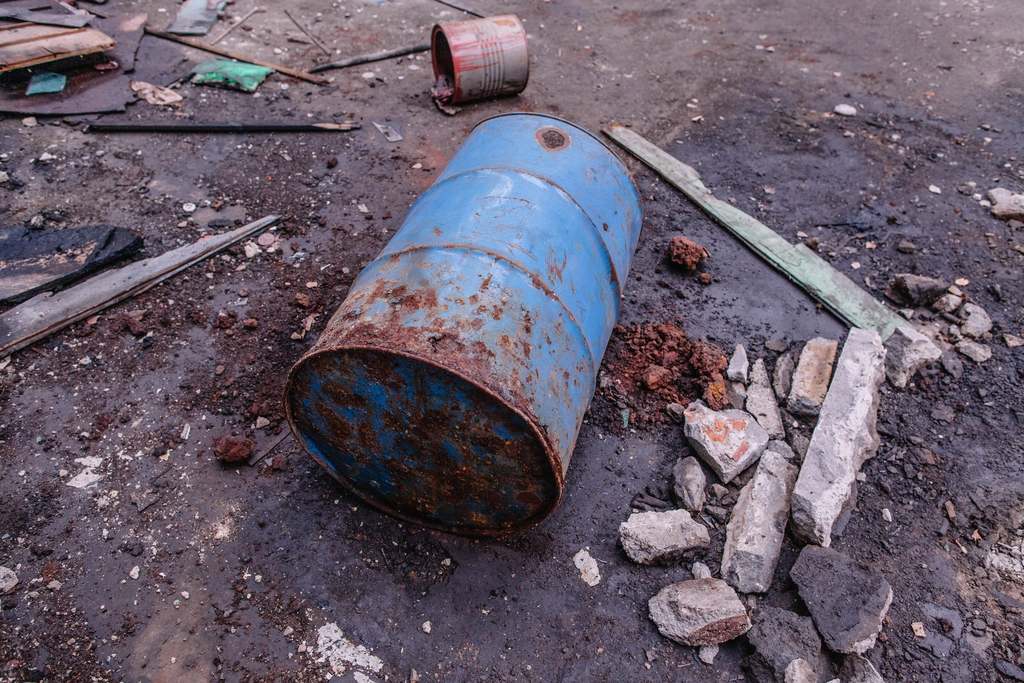}
		\caption{Photograph of barrel.\vspace{1em}}
		\label{fig:envasset_barrel_pic}
	\end{subfigure}
	~ 
	\begin{subfigure}[b]{0.235\textwidth}
		\includegraphics[width=\textwidth]{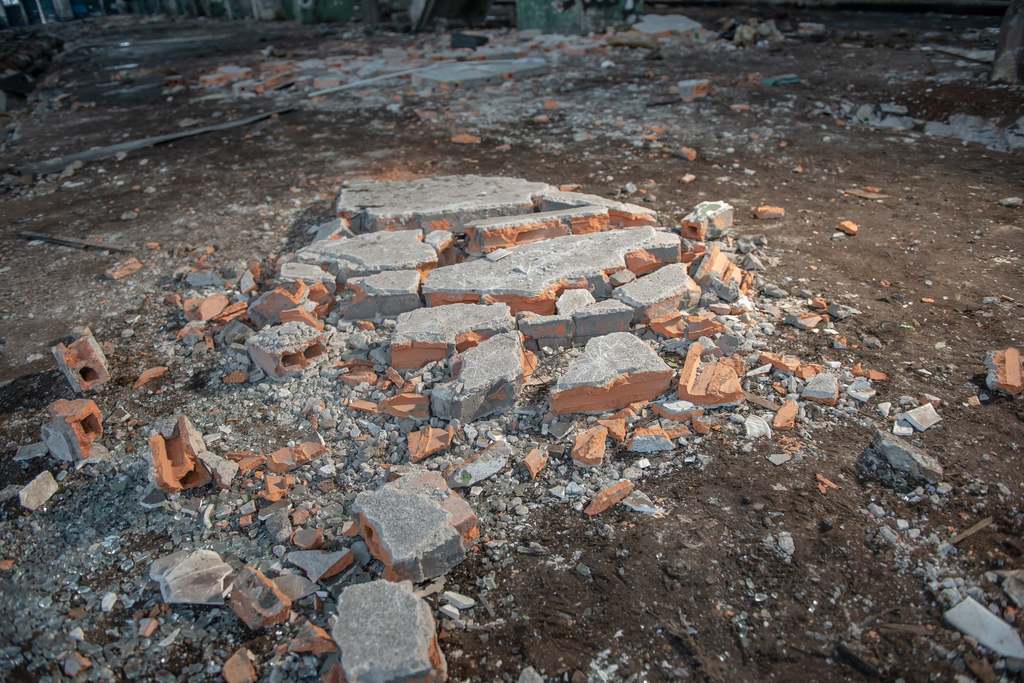}
		\caption{Photograph of rubble.\vspace{1em}}
		\label{fig:envasset_rubble_pic}
	\end{subfigure}
	~ 
	\begin{subfigure}[b]{0.235\textwidth}
		\includegraphics[width=\textwidth]{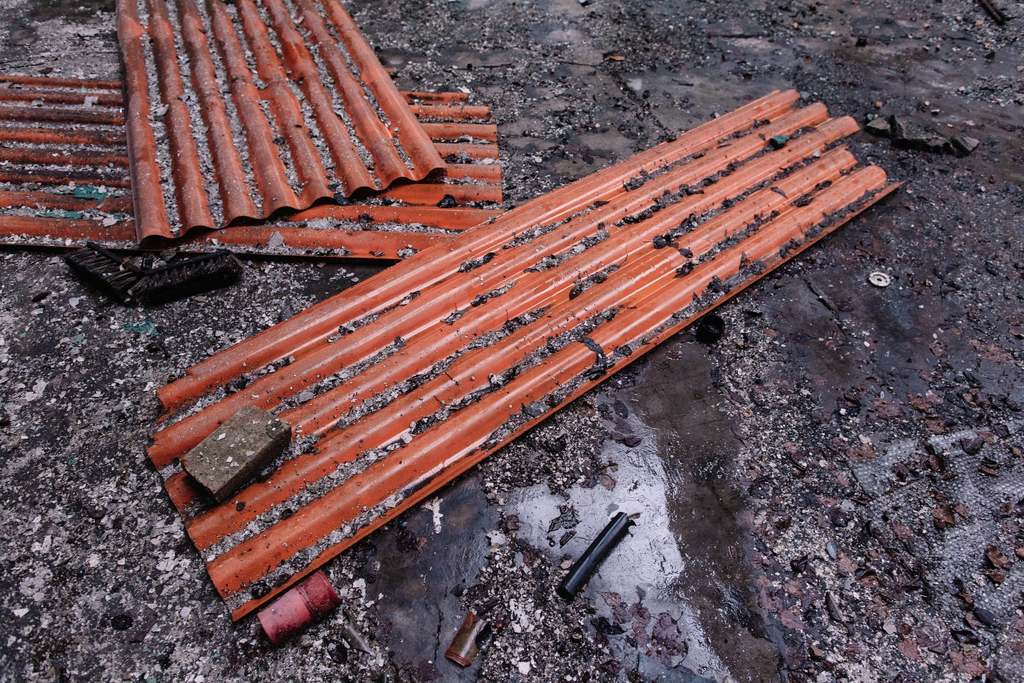}
		\caption{Photograph of corrugated metal.}
		\label{fig:envasset_corrugated_pic}
	\end{subfigure}
	~ 
	\begin{subfigure}[b]{0.235\textwidth}
		\includegraphics[width=\textwidth]{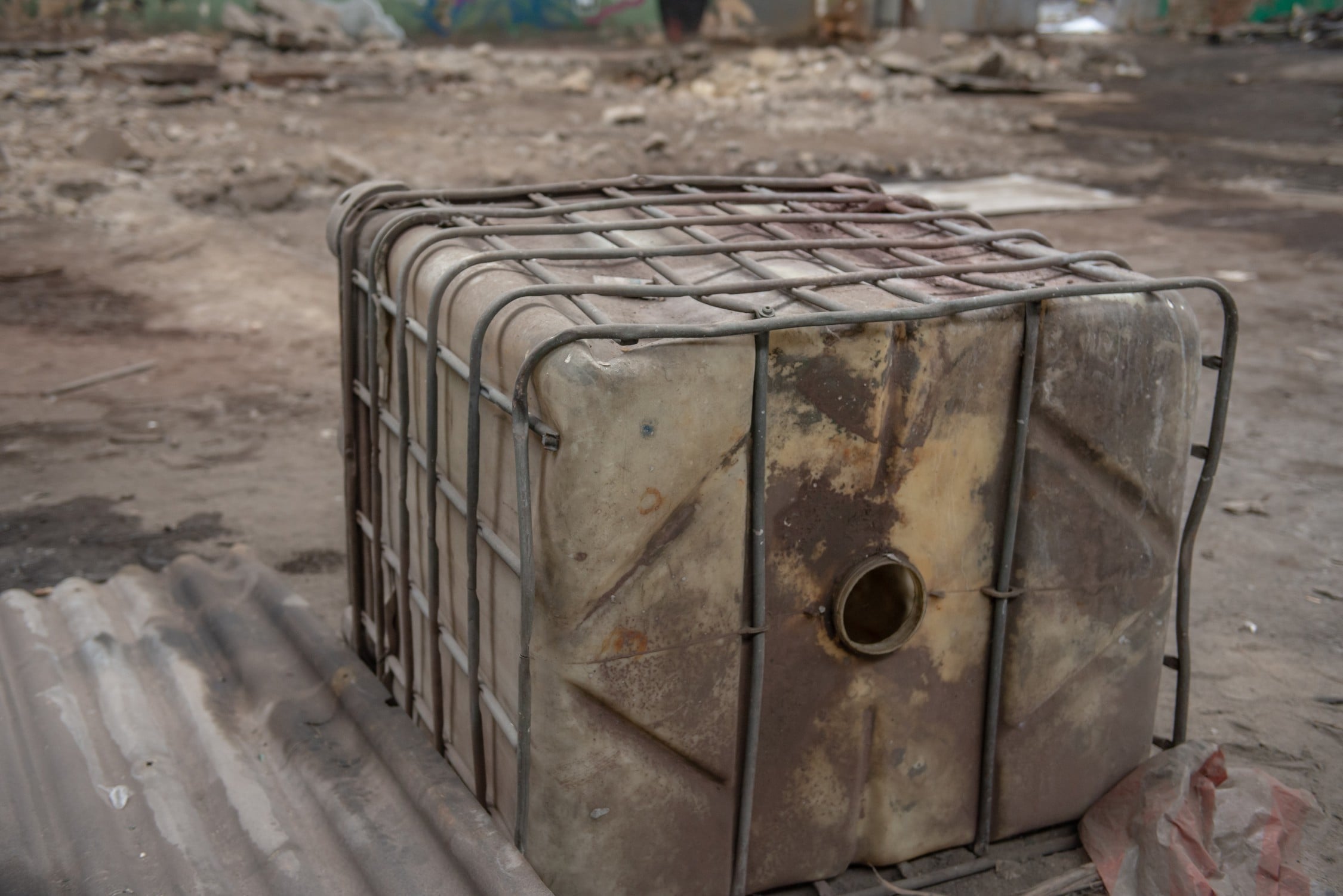}
		\caption{Photograph of caged tank.\vspace{1em}}
		\label{fig:envasset_tank_pic}
	\end{subfigure}

	\begin{subfigure}[b]{0.235\textwidth}
		\includegraphics[width=\textwidth]{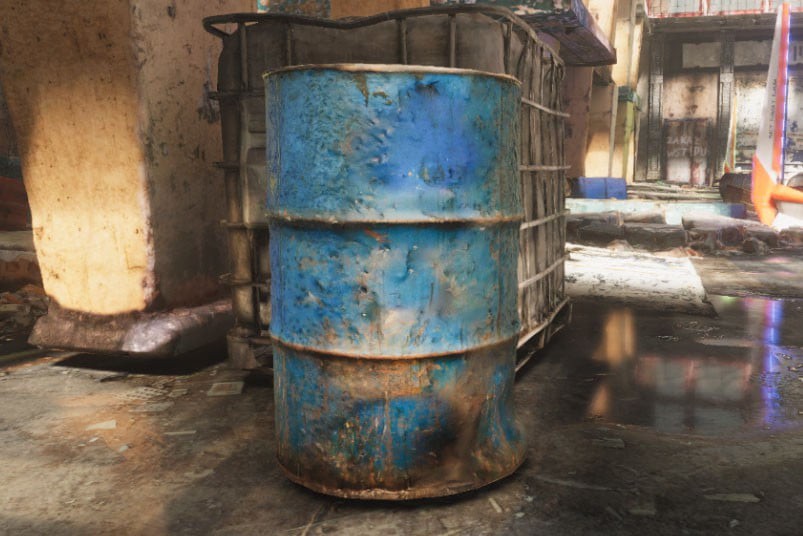}
		\caption{Rendered image of barrel.\vspace{1em}}
		\label{fig:envasset_barrel_render}
	\end{subfigure}
	~ 
	\begin{subfigure}[b]{0.235\textwidth}
		\includegraphics[width=\textwidth]{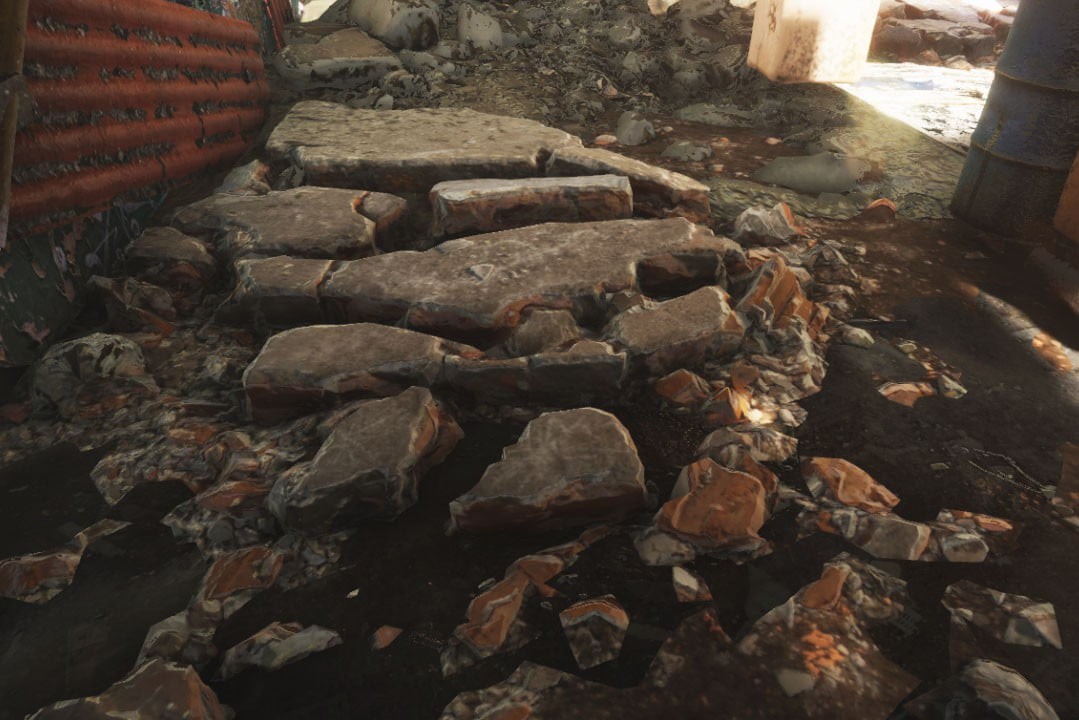}
		\caption{Rendered image of rubble.\vspace{1em}}
		\label{fig:envasset_rubble_render}
	\end{subfigure}
	~ 
	\begin{subfigure}[b]{0.235\textwidth}
		\includegraphics[width=\textwidth]{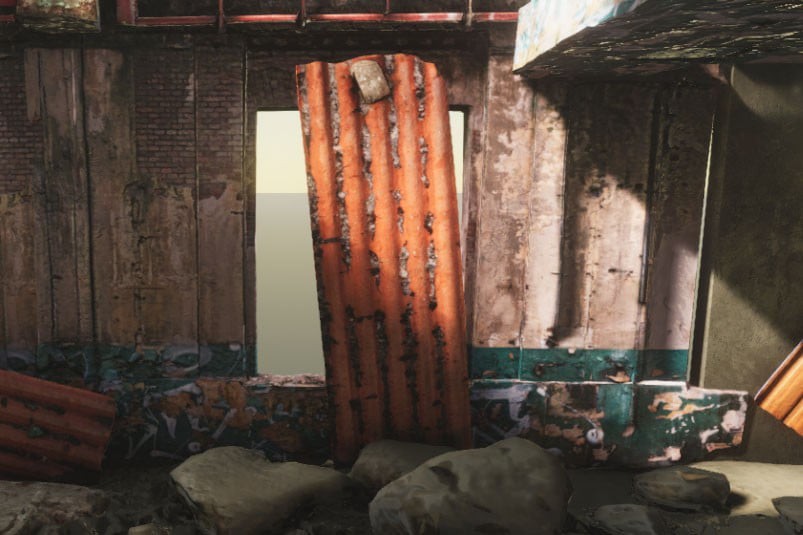}
		\caption{Rendered image corrugated metal.}
		\label{fig:envasset_corrugated_render}
	\end{subfigure}
	~ 
	\begin{subfigure}[b]{0.235\textwidth}
		\includegraphics[width=\textwidth]{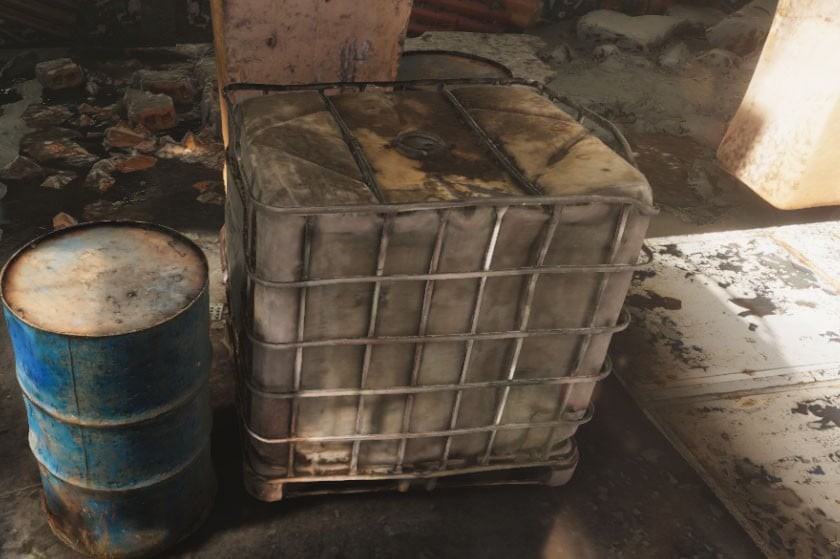}
		\caption{Rendered image of caged tank.}
		\label{fig:envasset_tank_render}
	\end{subfigure}
	\caption{Object photographs that were used for photogrammetry, and corresponding rendered assets in FlightGoggles.}
	\label{fig:envasset}
\end{figure*}

\begin{figure*}
   \centering
	\begin{subfigure}[t]{0.22\textwidth}
		\includegraphics[width=\textwidth]{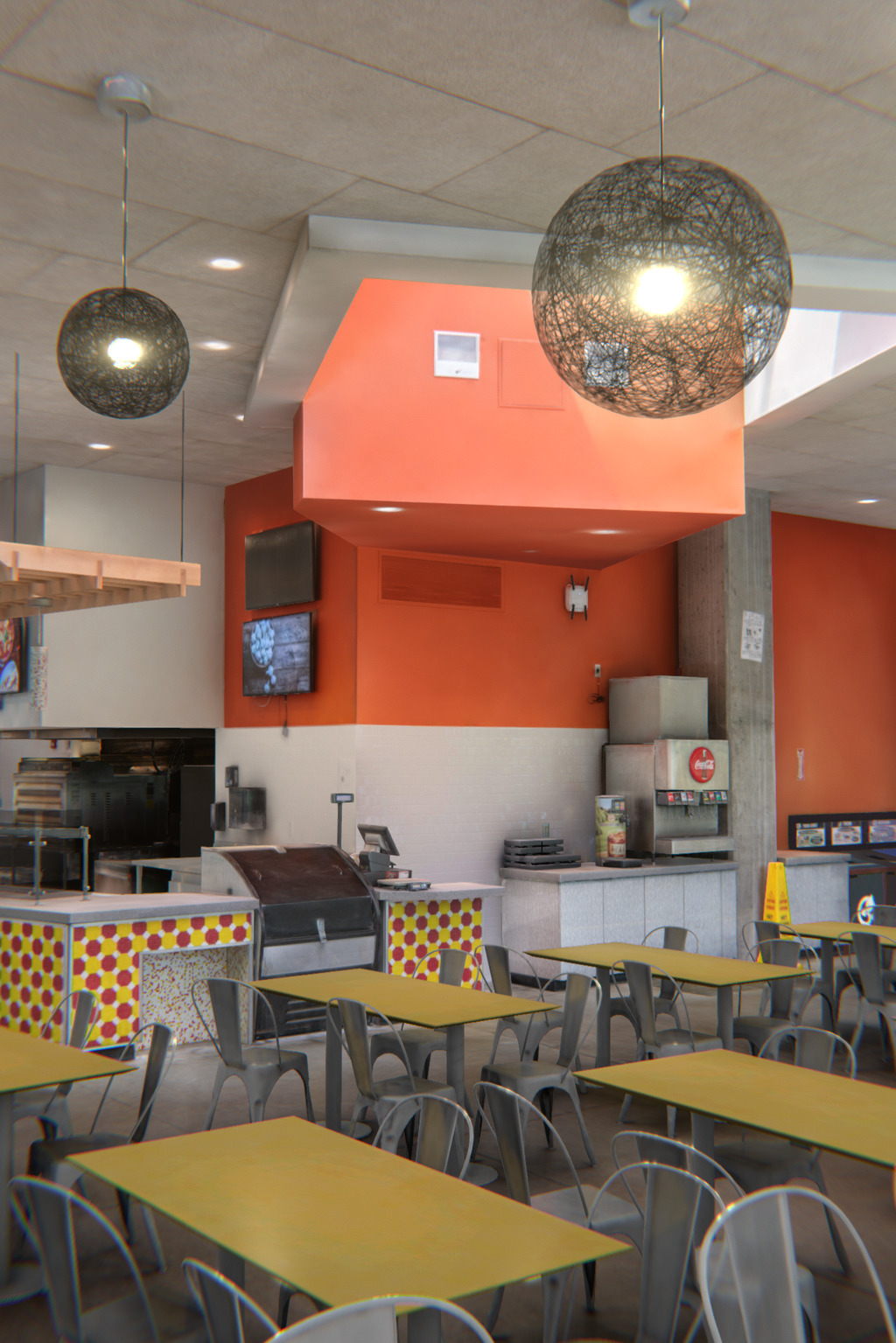}
		\caption{Rendered image of the Stata Center Cafe. \vspace{1em}}
		\label{fig:envasset_ground_floor_render}
	\end{subfigure}
	~ 
	\begin{subfigure}[t]{0.22\textwidth}
		\includegraphics[scale=0.25, width=\textwidth]{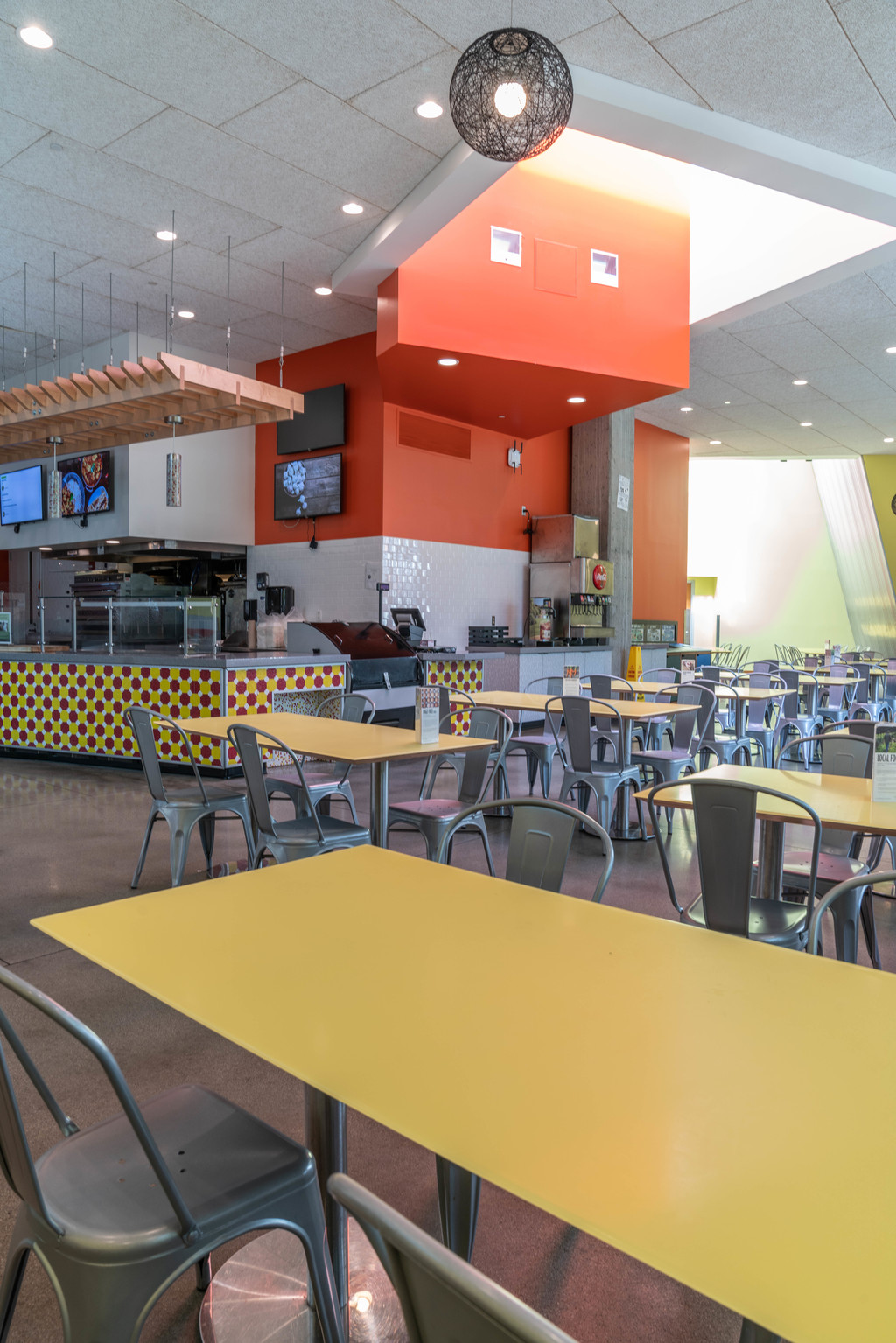}
		\caption{Photograph of the Stata Center Cafe. \vspace{1em}}
		\label{fig:envasset_ground_floor_real}
	\end{subfigure}   
~	
		\begin{subfigure}[t]{0.22\textwidth}
		\includegraphics[width=\textwidth]{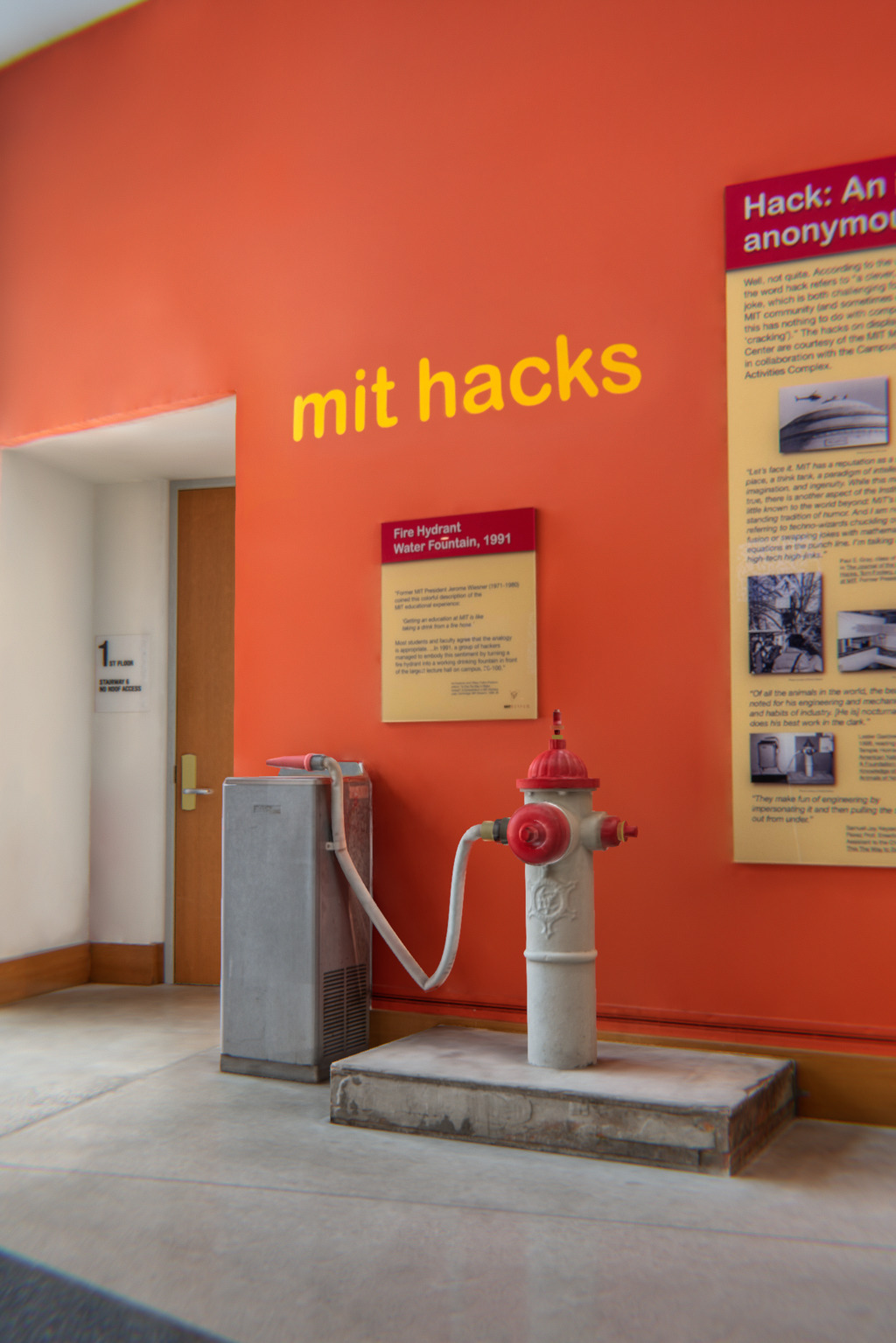}
		\caption{Rendered image of the Firehose. \vspace{1em}}
		\label{fig:envasset_ground_floor_render}
	\end{subfigure}
	~ 
	\begin{subfigure}[t]{0.22\textwidth}
		\includegraphics[scale=0.25, width=\textwidth]{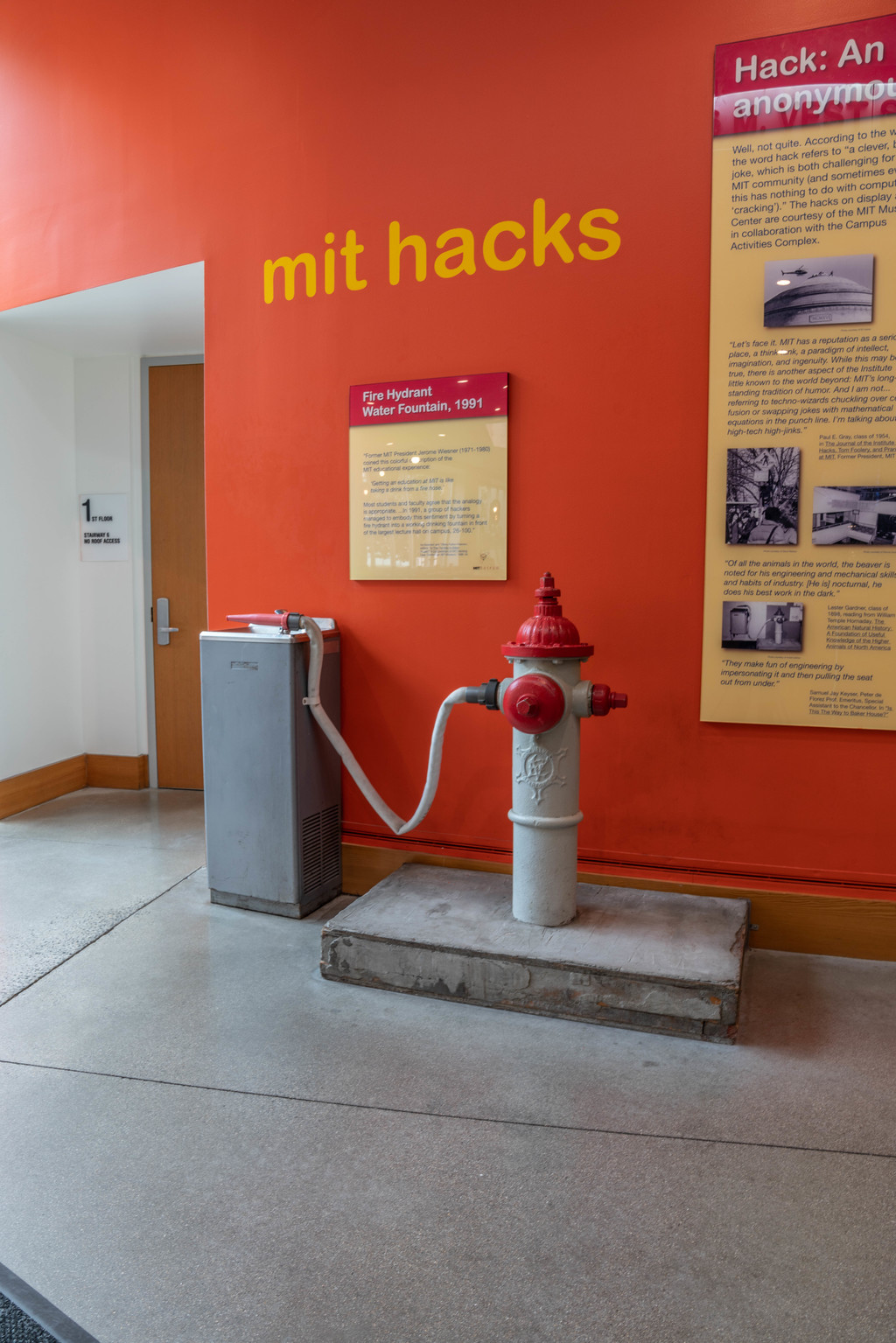}
		\caption{Photograph of the Firehose. \vspace{1em}}
		\label{fig:envasset_ground_floor_real}
	\end{subfigure} 
	~
	  	\begin{subfigure}[t]{0.22\textwidth}
		\includegraphics[width=\textwidth]{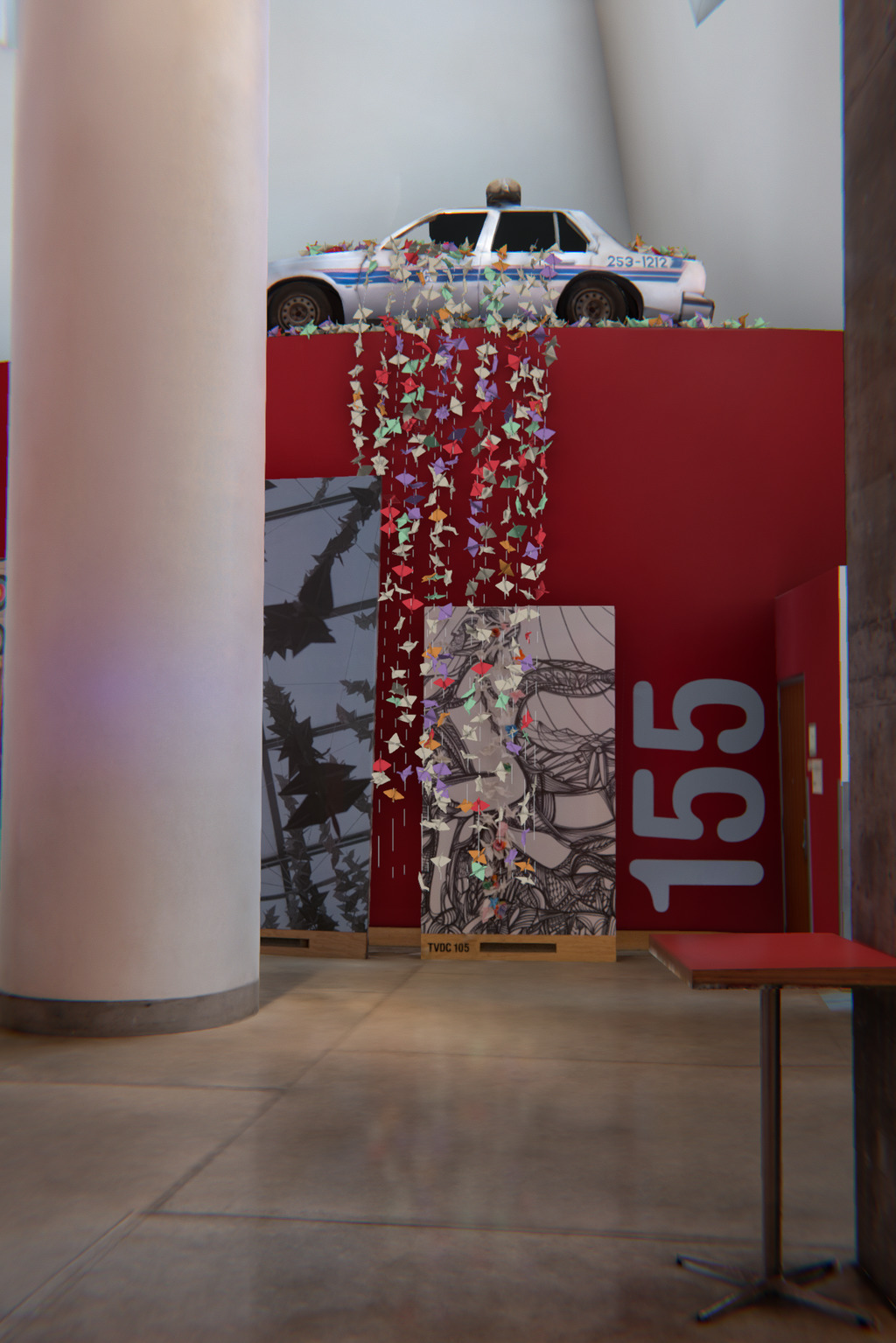}
		\caption{Rendered image of the car. \vspace{1em}}
		\label{fig:envasset_ground_floor_render}
	\end{subfigure}
	~ 
	\begin{subfigure}[t]{0.22\textwidth}
		\includegraphics[scale=0.25, width=\textwidth]{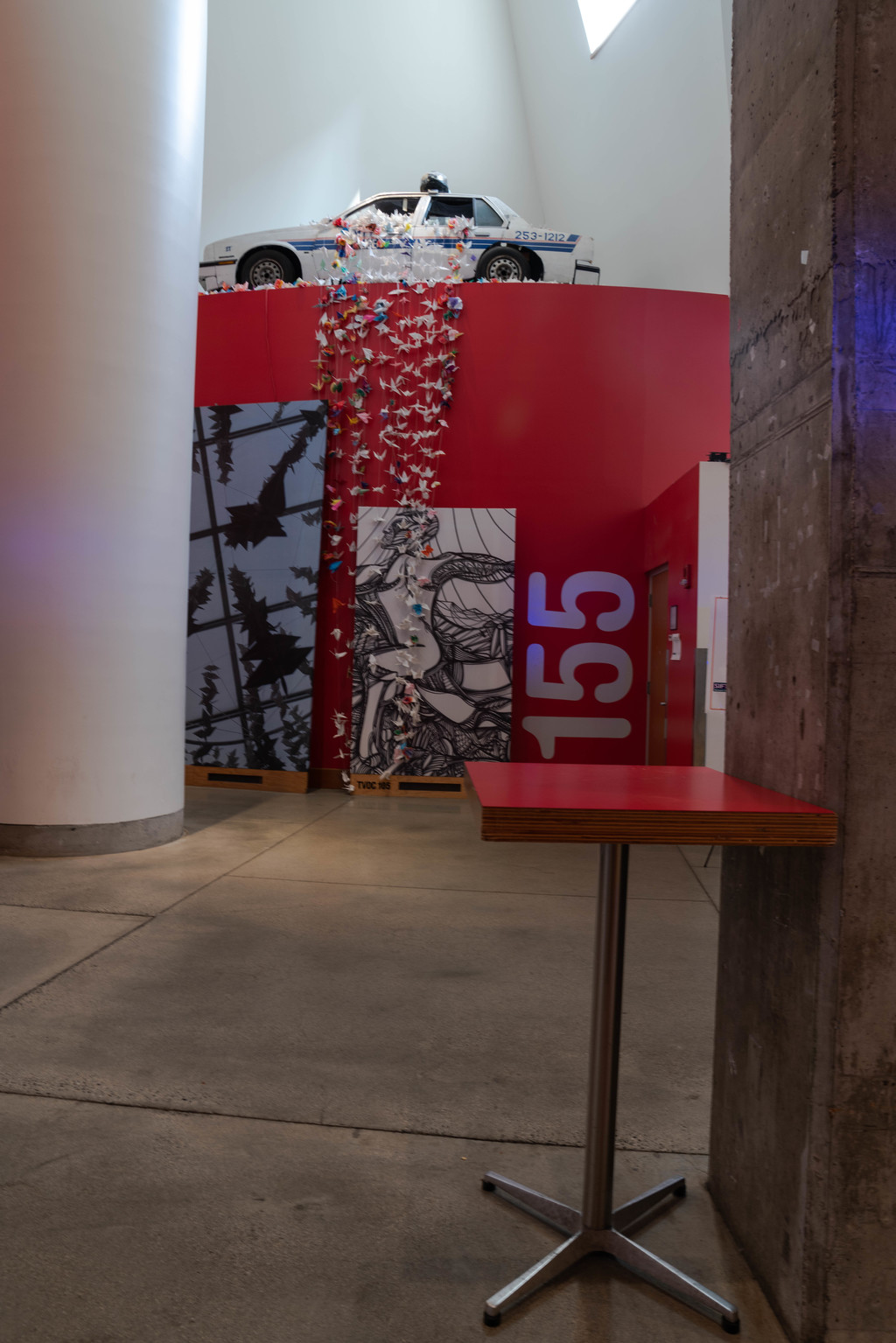}
		\caption{Photograph of the Car. \vspace{1em}}
		\label{fig:envasset_ground_floor_real}
	\end{subfigure}   
	~
		\begin{subfigure}[t]{0.22\textwidth}
		\includegraphics[width=\textwidth]{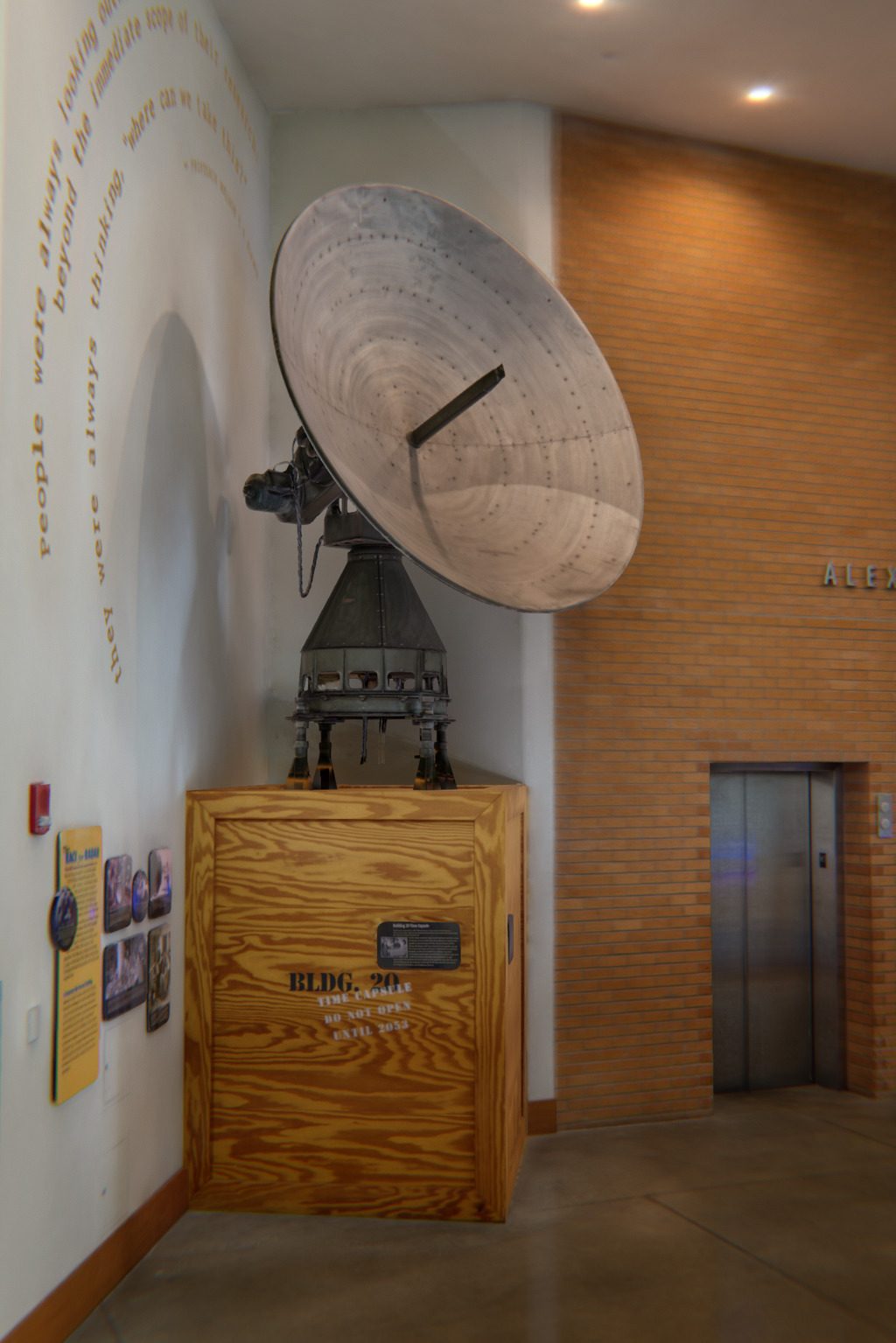}
		\caption{Rendered image of the Radar. \vspace{1em}}
		\label{fig:envasset_ground_floor_render}
	\end{subfigure}
	~ 
	\begin{subfigure}[t]{0.22\textwidth}
		\includegraphics[scale=0.25, width=\textwidth]{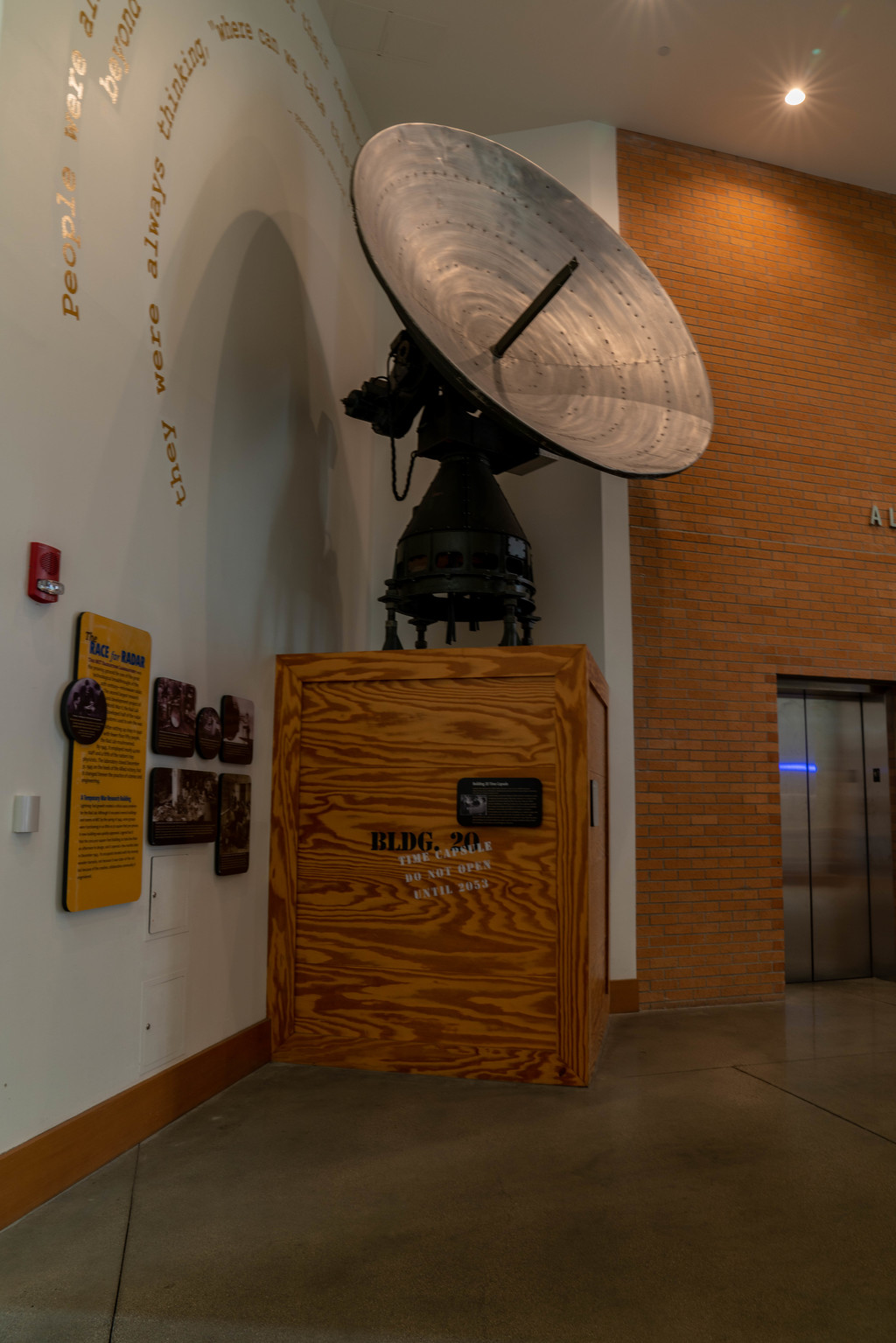}
		\caption{Photograph of the Radar. \vspace{1em}}
		\label{fig:envasset_ground_floor_real}
	\end{subfigure}
	~
		\begin{subfigure}[t]{0.22\textwidth}
		\includegraphics[width=\textwidth]{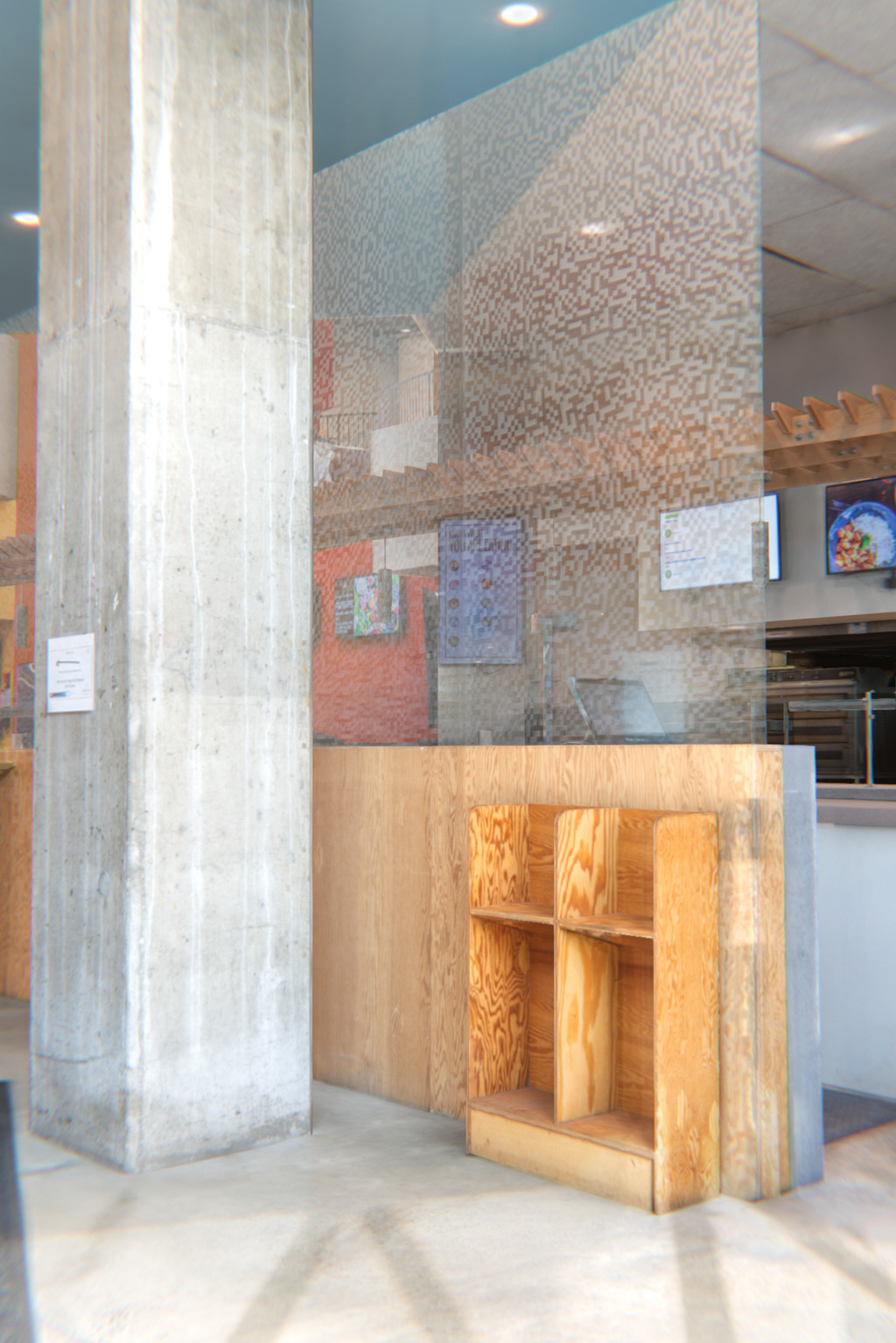}
		\caption{Rendered image of the Shelf. \vspace{1em}}
		\label{fig:envasset_ground_floor_render}
	\end{subfigure}
	~ 
	\begin{subfigure}[t]{0.22\textwidth}
		\includegraphics[scale=0.25, width=\textwidth]{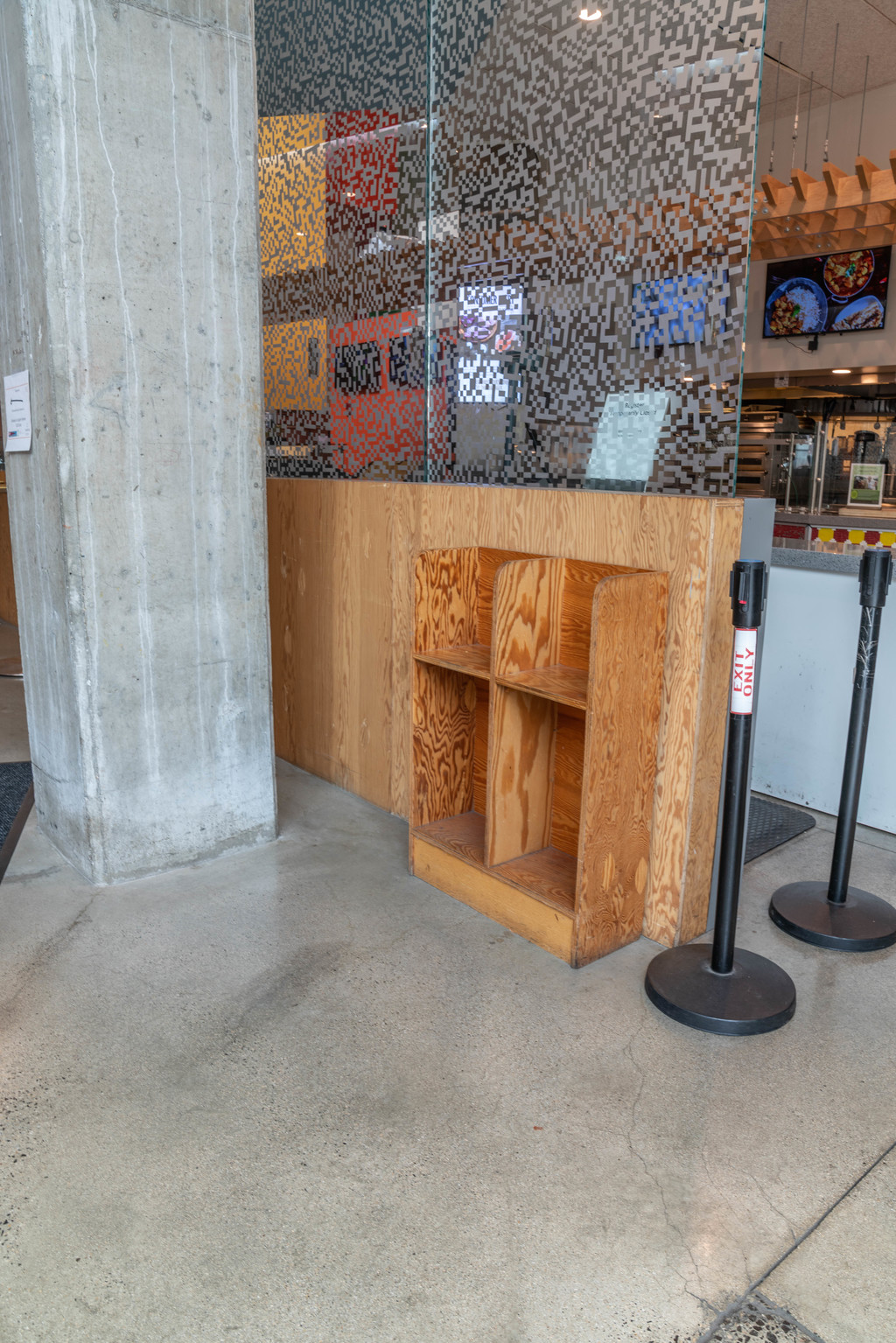}
		\caption{Photograph of the Shelf. \vspace{1em}}
		\label{fig:envasset_ground_floor_real}
	\end{subfigure}      
~
		\begin{subfigure}[t]{0.22\textwidth}
		\includegraphics[width=\textwidth]{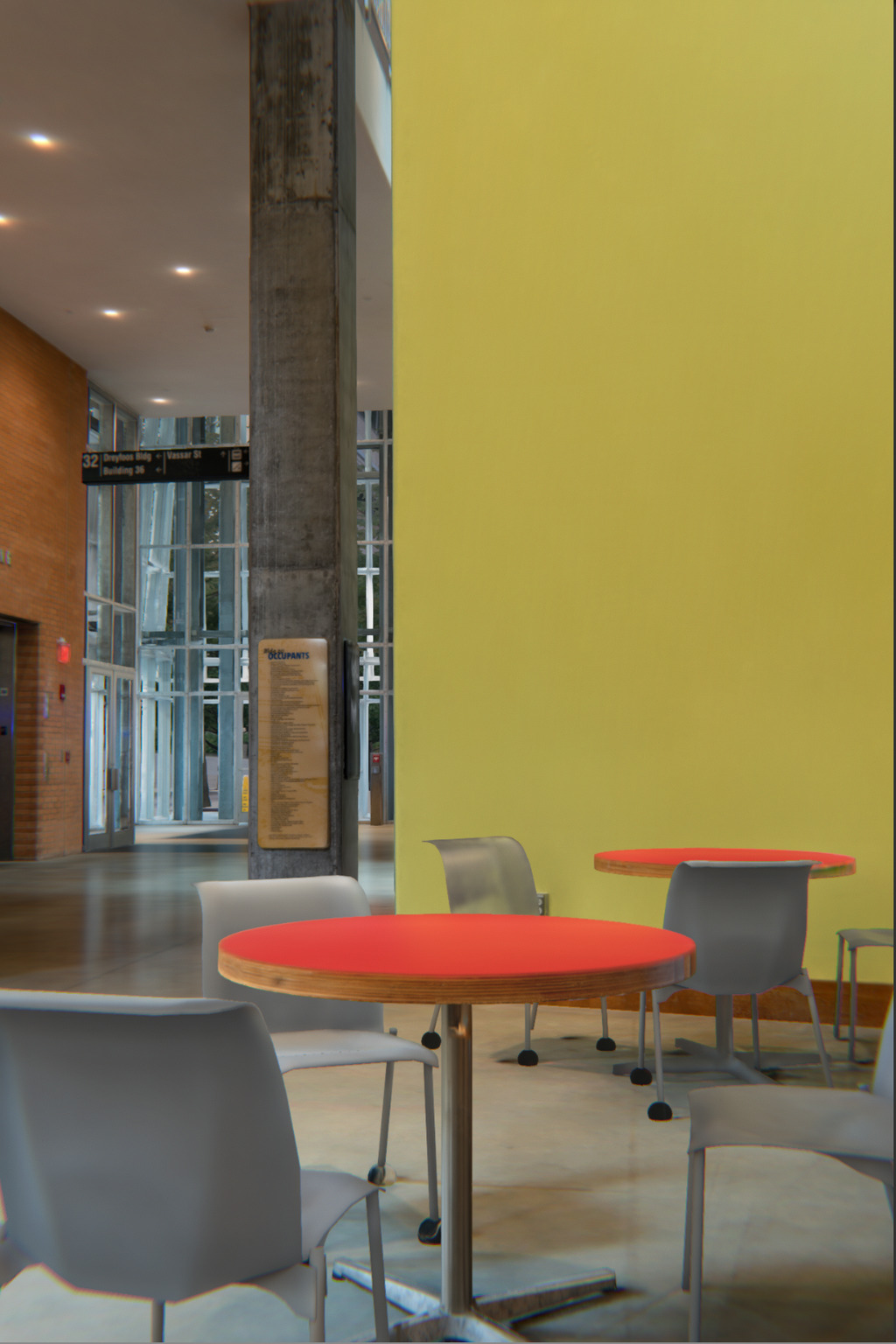}
		\caption{Rendered image of a study space. \vspace{1em}}
		\label{fig:envasset_ground_floor_render}
	\end{subfigure}
	~ 
	\begin{subfigure}[t]{0.22\textwidth}
		\includegraphics[scale=0.25, width=\textwidth]{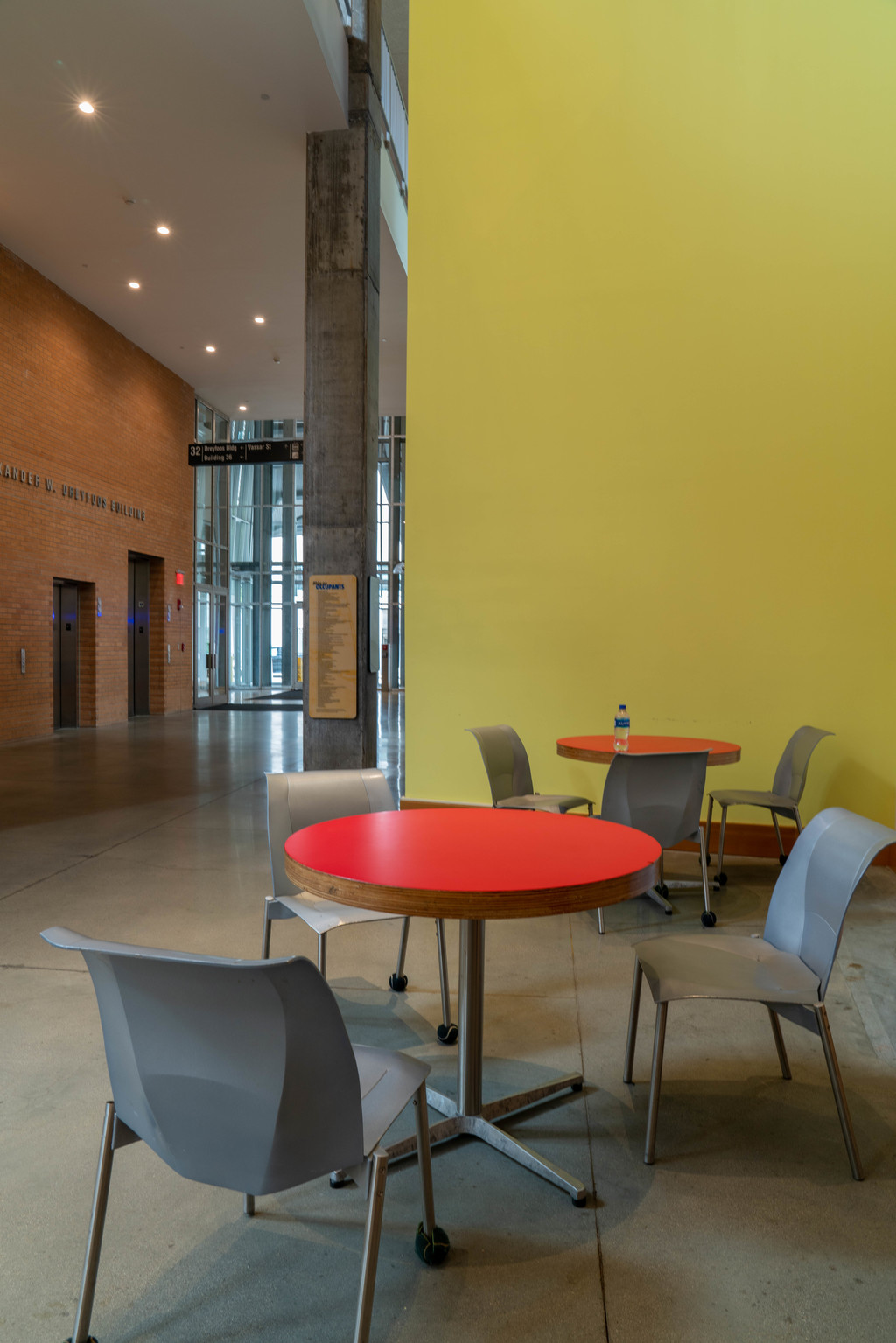}
		\caption{Photograph of a study space. \vspace{1em}}
		\label{fig:envasset_ground_floor_real}
	\end{subfigure}  
	\caption{Comparison images showing real images and a rendering from a similar viewpoint in the ground floor environment.}
	\label{fig:comp_ground_floor}
\end{figure*}

\begin{figure*}
\centering
	\begin{subfigure}[t]{0.44\textwidth}
		\includegraphics[width=\textwidth]{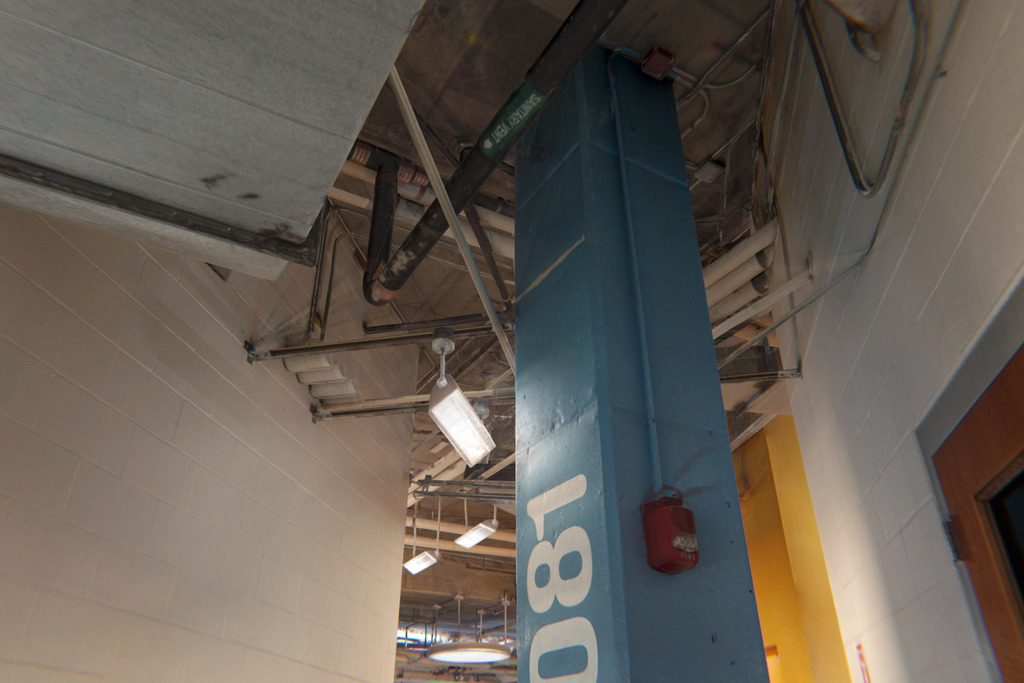}
		\caption{Rendered image of the pipes and ventilation system in the basement.\vspace{1em}}
		\label{fig:envasset_basement_render}
	\end{subfigure}
	~ 
	\begin{subfigure}[t]{0.44\textwidth}
		\includegraphics[width=\textwidth, scale=0.25]{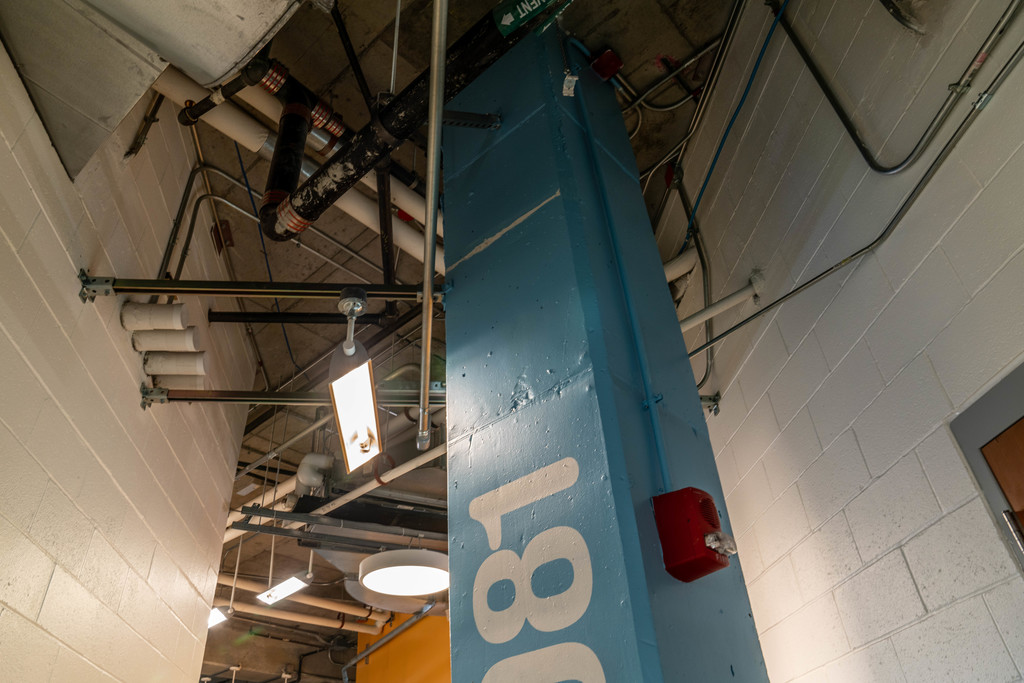}
		\caption{Photograph of the pipes and ventilation system in the basement.\vspace{1em}}
		\label{fig:envasset_basement_real}
	\end{subfigure}
~
		\begin{subfigure}[b]{0.44\textwidth}
		\includegraphics[width=\textwidth]{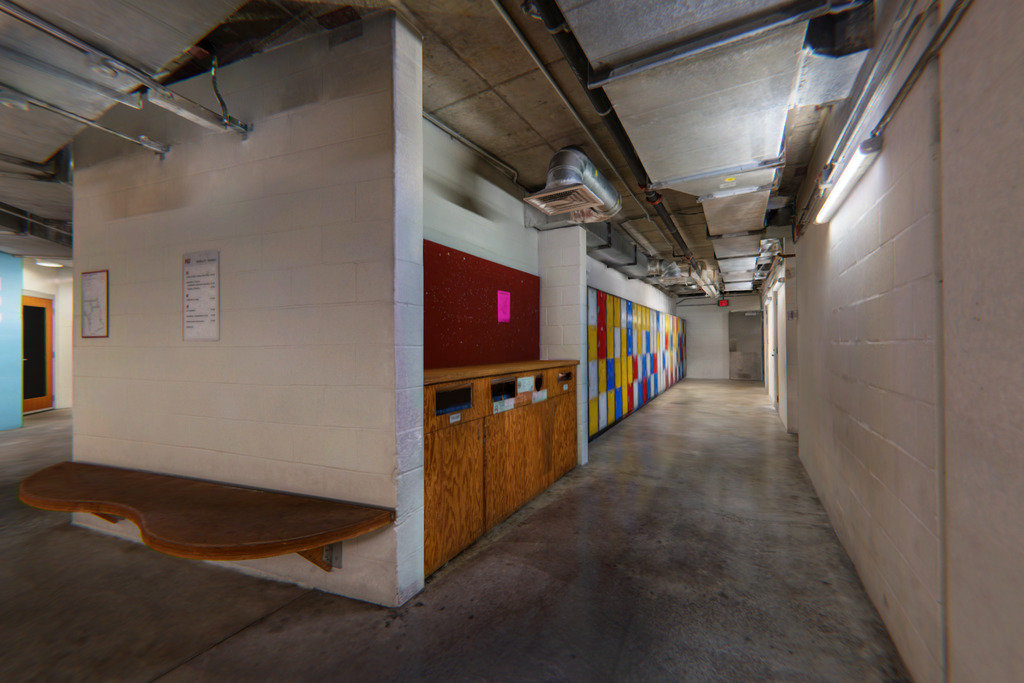}
		\caption{Rendered image of the corridor and lockers.\vspace{1em}}
		\label{fig:envasset_basement_2_render}
	\end{subfigure}
	~ 
	\begin{subfigure}[b]{0.44\textwidth}
		\includegraphics[width=\textwidth, scale=0.25]{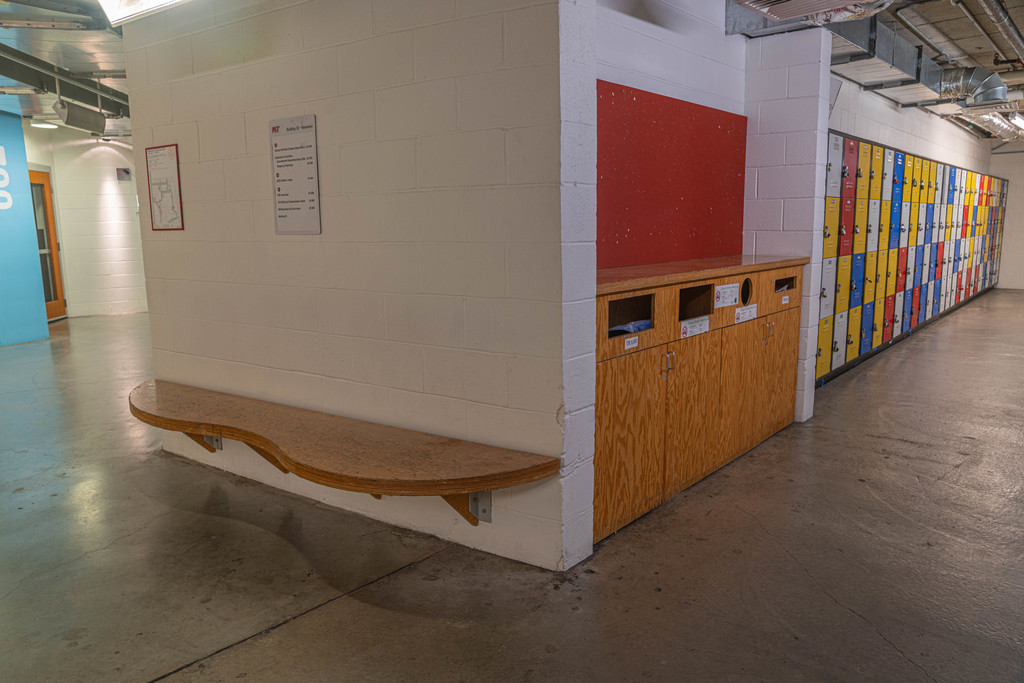}
		\caption{Photograph of the corridor and lockers.\vspace{1em}}
		\label{fig:envasset_basement_2_real}
	\end{subfigure}
	\caption{Comparison images of photographs and rendered images from similar viewpoints in the Basement environment.}
	\label{fig:comp_basement}
\end {figure*}

\begin{figure*}
\includegraphics[width=\textwidth]{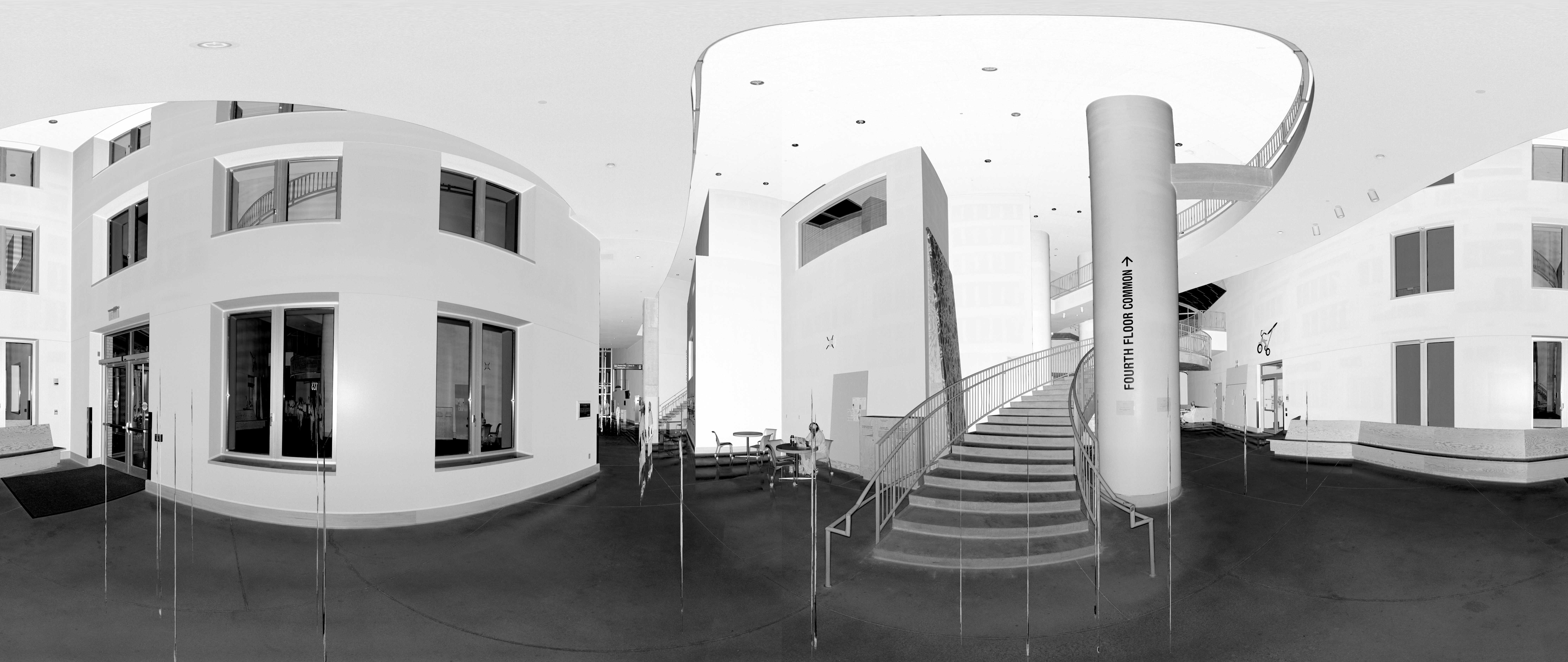}
\caption{Laser scan captured of the first floor of the Stata Center}
\label{fig:laser_scan}
\end{figure*}

FlightGoggles provides simulation environments with exceptional visual fidelity.
Its high level of photorealism is achieved using 84 unique 3D models in the Abandoned Factory environment and 75 3D models in the Stata Center Environments captured from real-world objects using photogrammetry. 
An example of the comparisons for the Abandoned Factory environment can be seen in Figure \ref{fig:envasset}.
Figure \ref{fig:comp_ground_floor} and Figure \ref{fig:comp_basement} show the comparisons of objects in the environment with photographs of those objects.

\subsection{Photorealistic Sensor Simulation using Photogrammetry}
Photogrammetry creates the 3D model of a real-world object from its photographs and (when available) data from laser scanners. 
For constructing the Stata Center assets, we used the FARO XS330 laser scanner.
An example of the laser scan is shown in \ref{fig:laser_scan}.
Multiple photographs from different viewpoints of a real-world object are used to construct a high-resolution 3D model for use in virtual environments.
For comparison, traditional 3D modeling techniques for creating photorealistic assets require hand-modeling and texturing, both of which require large amounts of artistic effort and time.
Firstly, modern photogrammetry techniques enable the creation of high-resolution assets in a much shorter time when compared to conventional modeling method. 
Secondly, the resulting renderings are visually far closer to the real-world object that is being modeled, which may be critical in robotics simulation. 
Due to its advantages over traditional modeling methods, photogrammetry is already widely used in the video game industry; however, photogrammetry has still not found traction within the robotics simulation community.
\subsubsection{Photogrammetry asset capture pipeline}
Photogrammetry was used to create open-source 3D assets for the FlightGoggles environment.
These assets are based on thousands of high-resolution digital photographs of real-world objects and environmental elements, such as walls and floors.
The digital images were first color-balanced, and then combined to reconstruct object meshes using the GPU-based reconstruction software \textit{Reality Capture}~\cite{realityCapture}.
After this step, the raw object meshes were manually cleaned to remove reconstruction artifacts.
Mesh baking was performed to generate base color, normal, height, metallic, ambient occlusion, detail, and smoothness maps for each object; which are then combined into one high-definition object material in Unity3D.

%% file: visualrendering.tex

\section{EXTEROCEPTIVE SENSOR SIMULATION}
\label{sec:exteroceptive}
This section describes each of the sensor models available and the performance optimizations performed to speed up the rendering and accommodate underpowered systems.

\subsection{Performance Optimizations and System Requirements}
To ensure that FlightGoggles is able to run on a wide spectrum of GPU rendering hardware with $\ge$ 2GB of video random access memory (VRAM), aggressive performance and memory optimizations were performed.
FlightGoggles VRAM and GPU computation requirements can be reduced using user-selectable
quality profiles based on three major settings:
real-time reflections,
maximum object texture resolution, and
maximum level of detail (\ie\, polygon count).

\subsubsection{Level of detail (LOD) optimization}
For each object mesh in the environment, 3 LOD meshes with different levels of detail (\ie\, polygon count and texture resolution) were generated: low, medium, and high. 
For meshes with lower levels of detail, textures were downsampled using subsampling and subsequent smoothing.
During simulation, the real-time render pipeline improves render performance by selecting the appropriate level of detail object mesh and texture based on the size of the object mesh in camera image space.
Users can also elect to decrease GPU VRAM usage by capping the maximum level of detail to use across all meshes in the environment using the quality settings.

\subsubsection{Render batching optimizations}
In order to increase render performance by reducing individual GPU draw calls, FlightGoggles leverages two different methods of render batching according to the capabilities available in the rendering machine. 
For Windows-based systems supporting DirectX11, FlightGoggles leverages Unity3D's experimental Scriptable Render Pipeline dynamic batcher, which drastically reduces GPU draw calls for all static and dynamic objects in the environment. 
For Linux and MacOS systems, FlightGoggles statically batches all static meshes in the environment. 
Static batching drastically increases render performance, but also increases VRAM usage in the GPU as all meshes must be combined and preloaded onto the GPU at runtime.
To circumvent this issue, FlightGoggles exposes quality settings to the user that can be used to lower VRAM usage for systems with low available VRAM.

\subsubsection{Dynamic clock scaling in ROS}
For rendering hardware that is incapable of reliable real-time frame rates despite the numerous rendering optimizations that FlightGoggles employs, FlightGoggles can perform automatic clock scaling to guarantee a nominal camera frame rate in simulation time when using the ROS framework.
When automatic clock scaling is enabled, FlightGoggles monitors the frame rate of the renderer output and dynamically adjusts ROS' sim-time rate to achieve the desired nominal frame rate in sim-time.
Since FlightGoggles uses the built-in ROS sim-time framework, changes in sim-time rate does not affect the relative timing of client nodes and helps to reduce simulation stochasticity across simulation runs. 

\subsection{Exteroceptive Sensor Models}
FlightGoggles is capable of high-fidelity simulation of various types of exteroceptive sensors, such as RGB-D cameras, time-of-flight distance sensors, and infrared radiation (IR) beacon sensors. Default noise characteristics, and intrinsic and extrinsic parameters are based on real sensor specifications, and can easily be adjusted. Moreover, users can instantiate multiple instances of each sensor type.
This capability allows quick prototyping and evaluation of distinct exteroceptive sensor arrangements.

\begin{table}[h]
\centering
\begin{tabular}{@{}ll@{}}
\toprule
Camera Parameter             & Default Value        \\ \midrule
Vertical Field of View ($fov$) & $70^{\circ}$           \\
Image Resolution ($w \times h$)       & $1024$\,px $\times\, 768$\,px \\
Stereo Baseline ($t$)          & $32$\,cm             \\ \bottomrule
\end{tabular}
\caption{Camera sensor parameters enabled by default in FlightGoggles along with their default values.}
\label{tab:camera_params}
\end{table}

\subsubsection{Camera}
The default camera model provided by FlightGoggles is a perfect, \ie, distortion-free, camera projection model with optional motion blur and lens dirt. Table \ref{tab:camera_params} lists the major camera parameters exposed by default in the FlightGoggles API along with their default values. These parameters can be changed via the FlightGoggles API. The camera extrinsics $T^{b}_{c}$ where $b$ is the vehicle fixed body frame and $c$ is the camera frame can also be changed in real-time for \eg to simulate a gimbaled camera.

\subsubsection{Infrared beacon sensor}

To facilitate the quick development of guidance, navigation, and control algorithms; an IR beacon sensor model is included.
This sensor provides image-space $u, v$ measurements of IR beacons in the camera's field of view.
The beacons can be placed at static locations in the environment or on moving objects.
Using realtime ray-casting from each RGB camera, simulated IR beacon measurements are tested for occlusion before being included in the IR sensor output.
Figure \ref{fig:ir_marker_sensor} shows a visual representation of the sensor output.

\begin{figure}[tbp]
	\centering	
	\includegraphics[width=0.48\textwidth, trim={1cm 5cm 7cm 5cm},clip]{images/ir_markers_faded_background}
	\caption{Rendered camera view (faded) with IR marker locations overlayed. 
		The unprocessed measurements and marker IDs from the simulated IR beacon sensor are indicated in red.
		The measurements are verified by comparison to image-space reprojections of ground-truth IR marker locations, which are indicated in green. Note that IR markers can be arbitrarily placed by the user, including on dynamic objects.}
	\label{fig:ir_marker_sensor}
\end{figure}

\subsubsection{Time-of-flight range sensor}

FlightGoggles is able to simulate (multi-point) time-of-flight range sensors using ray casts in any specified directions. In the default vehicle configuration, a downward-facing single-point range finder for altitude estimation is provided. The noise characteristics of this sensor are similar to the commercially available LightWare SF11/B laser altimeter~\cite{laserRangeFinder}. 

\subsection{Camera Types}

FlightGoggles allows the simulation of the following camera types to enable varied applications from real time guidance, navigation and control to machine learning to dataset generation.

\subsubsection{RGB Camera}
This mode is the default for a camera type requested from the FlightGoggles renderer and returns a 3 channel (24 bit) RGB image. 

\subsubsection{Grayscale Camera}
This mode returns a 1 channel (8 bit) grayscale image.
In this mode, the network usage is reduced by a factor of 3 in comparison to the RGB camera and is advantageous to application such as testing visual odometry where only grayscale images are required for many algorithms.

\subsubsection{Depth Camera}
This mode return the true depth performed using real time racing as a 1 channel (16 bit) image.
The maximum depth is 100m and the resolution of the depth image is 0.0015m.

\subsubsection{Optical Flow Camera}
This mode simulates the optical flow between frames and returns the optic flow as a 3 channel (24 bit) image colored as an Hue Saturation Value (HSV) encoding of the optic flow.

\subsubsection{Semantic Segmentation}
This mode returns the object category ID as a 3 channel (24 bit) image.

%% file: vehicledynamics.tex

\section{SIMULATED VEHICLES}\label{sec:sim_vehicles}
FlightGoggles can be applied to scenarios including both real-life and simulated vehicles.
This section details the models used for simulation of flight dynamics and inertial measurements of multicopter vehicles.
Additionally, FlightGoggles provides a simple car dynamics simulation, which is also described here.

\subsection{Motor Dynamics}
We model the dynamics of the $N$ motors using a first-order lag with time constant $\tau_m$, as follows:
\begin{equation}\label{eq:sim_motordyn}
{\dot \omega}_i = \frac{1}{\tau_m}\left({\omega}_{i,c} - {\omega_i}\right),
\end{equation}
where ${\omega_i}$ is the rotor speed of motor $i$, with $i = 1,2, \dots, N$ ($N=4$ in case of a quadcopter) and the subscript $c$ indicates the commanded value. Rotor speed is defined such that $\omega_i > 0$ corresponds to positive thrust in the motor frame z-axis.

\subsection{Forces and Moments}\label{sec:dyn:forcesmoments}
We distinguish two deterministic external forces and moments acting on the vehicle body: firstly, thrust force and control moment due to the rotating propellers; secondly, aerodynamic drag and moment due to the vehicle linear and angular speed.

\subsubsection{Thrust force and control moment}
We employ a summation of forces and moments over the propeller index $i$ in order to be able to be able to account for various vehicle and propeller configurations in a straightforward manner. The total thrust and control moment are given by
\begin{align}
\vect{f}_T &= \sum_{i} \vect{R}^b_{m_i} \vect{f}_{T,i},\label{eq:ft}\\
\vect{\mu}_T &= \sum_{i} \left(\vect{R}^b_{m_i}\vect{\mu}_{T,i} + \vect{r}_i \times \vect{R}^b_{m_i}\vect{f}_{T,i} \right)\label{eq:mut}
\end{align}
with $\vect{R}^b_{m_i} \in SO(3)$ the constant rotation matrix from the motor reference frame to the vehicle-fixed reference frame, and $\vect{r}_i$ the position of the motor in the latter frame. The force and moment vector in the motor reference frame are given by
\begin{align}
\vect{f}_{T,i} &= \left[\begin{array}{c c c}
0 & 0 & k_{\vect{f}_T} \omega_i\vert\omega_i\vert
\end{array}\right]^T,\\
\vect{\mu}_{T,i}  &= \left[\begin{array}{c c c}
0 & 0 & (-1)^{1_{\omega}(i)} \left(k_{\vect{\mu}_T}\omega_i \vert\omega_i\vert + J_m \dot\omega_i\right)
\end{array}\right]^T,
\end{align}
where $1_{\omega}(\cdot)$ is an indicator function for the set of propellers for which $\omega_i > 0$ corresponds to positive rotation rate around the motor frame z-axis; $J_m$ is the rotor and propeller mass moment of inertia; and $k_{\vect{f}_T}$ and $k_{\vect{\mu}_T}$ indicate the constant propeller thrust and torque coefficients, respectively.

\subsubsection{Aerodynamic drag and moment}
Aerodynamic drag has magnitude proportional to the vehicle speed squared and acts in direction opposite the vehicle motion according to
\begin{equation}\label{fd}
\vect{f}_D = - k_{\vect{f}_D} \|\vect{v}\|_2 \vect{v}
\end{equation}
with $\|\cdot\|_2$ the $l^2$-norm, $\vect{v}$ vehicle velocity relative to the world-fixed reference frame, and $k_{\vect{f}_D}$ the vehicle drag coefficient. Similarly, the aerodynamic moment is given by
\begin{equation}\label{md}
\vect{\mu}_D = - \vect{k}_{\vect{\mu}_D} \|\vect{\Omega}\|_2 \vect{\Omega}
\end{equation}
where $\vect{\Omega}$ is the angular rate in vehicle-fixed reference frame, and $\vect{k}_{\vect{\mu}_D}$ is a 3-by-3 matrix containing the aerodynamic moment coefficients.

\subsection{Vehicle Dynamics}
The vehicle translational dynamics are given by
\begin{align}
\vect{\dot x} &= \vect{v},\label{eq:xdot}\\
\vect{\dot v} &= \vect{g} + m^{-1}\left(\vect{R}_b^w \vect{f}_T + \vect{f}_D  + \vect{w}_{\vect{f}}\right),\label{eq:vdot}
\end{align}
where $m$ is the vehicle mass; and $\vect{x}$, $\vect{v}$, and $\vect{g}$ are the position, velocity, and gravitational acceleration in the world-fixed reference frame, respectively. The stochastic force vector $\vect{w}_{\vect{f}}$ captures unmodeled dynamics, such as propeller vibrations and atmospheric turbulence. It is modeled as a continuous white-noise process with auto-correlation function $\vect{W}_{\vect{f}}\delta(t)$ (with $\delta(\cdot)$ the Dirac delta function), and can thus be sampled discretely according to
\begin{equation}\label{eq:samplecontwn}
\vect{ w}_{\vect{f}}  \sim \mathcal{N}(\vect{0},\frac{\vect{W}_{\vect{f}}}{\Delta t}),
\end{equation}
where $\Delta t$ is the integration time step.
The rotation matrix from body-fixed to world frame is given by
\begin{equation}\label{key}
\vect{R}_b^w = \left[\begin{array}{ccc}1 - 2 (q_j^2 + q_k^2) &
2 (q_i q_j - q_k q_r) &
2 (q_i q_k + q_j q_r) \\
2 (q_i q_j + q_k q_r) &
1 - 2 (q_i^2 + q_k^2) &
2 (q_j q_k - q_i q_r) \\
2 (q_i q_k - q_j q_r) &
2 (q_j q_k + q_i q_r) &
1 - 2 (q_i^2 + q_j^2)\end{array}\right],
\end{equation}
where $\vect{q} = [q_r \; q_i \; q_j \; q_k]^T$ is the vehicle attitude unit quaternion vector. The corresponding attitude dynamics are given by
\begin{align}\label{eq:qdot}
\vect{\dot q} &= \frac{1}{2} \vect{q} \circ \vect{\Omega} = \frac{1}{2}\left[\begin{array}{ccc}
-q_i &-q_j &-q_k\\
 q_r &-q_k &q_j\\
q_k &q_r &-q_i\\
-q_j &q_i &q_r
\end{array}\right]\vect{\Omega},\\
\vect{\dot \Omega} &= \vect{J}^{-1}(\vect{\mu}_T + \vect{\mu}_{D}+\vect{w}_{\vect{\mu}}-\vect{\Omega} \times \vect{H})\label{eq:Omegadot}
\end{align}
with $\vect{J}$ the vehicle moment of inertia tensor, and $\vect{H}$ the angular momentum given by
\begin{equation}\label{eq:angular momentum}
\vect{H} = \vect{J}\vect{\Omega} - \sum_{i} \vect{R}^b_{m_i} \left[\begin{array}{c c c}
0 & 0 & (-1)^{1_{\omega}(i)} J_m \omega_i
\end{array}\right]^T.
\end{equation}
The stochastic moment contribution $\vect{w}_{\vect{\mu}}$ is modeled as a continuous white-noise process with auto-correlation function $\vect{W}_{\vect{\mu}}\delta(t)$, and sampled similarly to \eqref{eq:samplecontwn}.

\subsection{Physics Integration}
The vehicle state is updated at 960 Hz when using default settings, so that even high-bandwidth motor dynamics can be represented accurately. Both explicit Euler and 4th-order Runge-Kutta algorithms are provided for integration of \eqref{eq:sim_motordyn}, \eqref{eq:xdot}, \eqref{eq:vdot}, \eqref{eq:qdot}, and \eqref{eq:Omegadot}.

\subsection{Inertial Measurement Model}
Acceleration and angular rate measurements are obtained from a simulated IMU according to the following measurement equations:
\begin{align}\label{eq:imu}
\vect{a}_{IMU} &= \vect{R}_b^{IMU} m^{-1}\left(\vect{f}_T + \vect{R}_w^b \vect{f}_D  + \vect{R}_w^b\vect{f}_w\right) \notag\\
& {\;\;\;\;\;\;\;\;\;\;\;\;\;\;\;\;\;\;\;\;\;\;\;\;\;\;\;\;\;} + \vect{b}_{\vect{a}_{IMU}} + \vect{v}_{\vect{a}_{IMU}},\\
\vect{\Omega}_{IMU} &= \vect{R}_b^{IMU}\vect{\Omega} + \vect{b}_{\vect{\Omega}_{IMU}} + \vect{v}_{\vect{\Omega}_{IMU}}
\end{align}
where $\vect{b}_{\vect{a}_{IMU}}$ and $\vect{b}_{\vect{\Omega}_{IMU}}$ are the accelerometer and gyroscope measurement biases, respectively; and $\vect{v}_{\vect{a}_{IMU}} \sim \mathcal{N}(\vect{0},\vect{V}_{\vect{a}_{IMU}})$ and $\vect{v}_{\vect{\Omega}_{IMU}} \sim \mathcal{N}(\vect{0},\vect{V}_{\vect{\Omega}_{IMU}})$ the corresponding thermo-mechanical measurement noises. Brownian motion is used to model the bias dynamics, as follows:
\begin{align}
\vect{\dot b}_{\vect{a}_{IMU}} &= \vect{w}_{\vect{a}_{IMU}},\\
\vect{\dot b}_{\vect{\Omega}_{IMU}} &= \vect{w}_{\vect{\Omega}_{IMU}}
\end{align}
with $\vect{w}_{\vect{a}_{IMU}}$ and $\vect{w}_{\vect{\Omega}_{IMU}}$ continuous white noise processes with auto-correlation $\vect{W}_{\vect{a}_{IMU}}$ and $\vect{W}_{\vect{\Omega}_{IMU}}$, respectively, and sampled similar to \eqref{eq:samplecontwn}.

\subsection{Acro/Rate Mode Controller}
In order to ease the implementation of high-level guidance and control algorithms, a quadcopter rate mode controller can be enabled. This controller allows direct control of the vehicle thrust and angular rates, while maintaining accurate low-level dynamics in simulation.

\subsubsection{Low-pass filter (LPF)}
The rate mode controller employs gyroscope measurements for closed-loop angular rate control. A low-pass Butterworth filter is used to reduce the influence of IMU noise. The filter dynamics are as follows:
\begin{equation}
	\vect{\ddot \Omega}_{LPF} = - q_{LPF} \vect{\dot \Omega}_{LPF} + p_{LPF} \left(\vect{\Omega}_{IMU}-\vect{\Omega}_{LPF}\right),
\end{equation}
where the positive gains $q_{LPF}$ and $p_{LPF}$ represent the filter damping and stiffness, respectively.

\subsubsection{Proportional-integral-derivative (PID) control}
A standard PID control design is used to compute the commanded angular acceleration as a function of the angular rate command, as follows:
\begin{align}
\vect{\dot \Omega}_c &= \vect{K}_{P,PID} \Delta \vect{\Omega} + \vect{K}_{I,PID} \int \Delta \vect{\Omega} \operatorname{d}t\notag\\
& \;\;\;\;\;\;\;\;\;\;\;\;\;\;\;\;\;\;\;\;\;\;\;\;\;\;\;\;\;\;\;\;\;\;\;\;\;\;\;\; - \vect{K}_{D,PID} \vect{\dot \Omega}_{LPF} \\
\Delta \vect{\Omega} &= \vect{\Omega}_c - \vect{\Omega}_{LPF}
\end{align}
where $\vect{K}_{P,PID}$, $\vect{K}_{I,PID}$, and $\vect{K}_{D,PID}$ are diagonal gain matrices.

\subsubsection{Control allocation}
Given the angular acceleration command $\vect{\dot \Omega}_c$ and the collective thrust command $T_c$, the controller computes the motor speed commands that result in the appropriate control moment and thrust force.

First, the required control moment is found by inversion of the expressions for the angular acceleration $\vect{\dot \Omega}$ in \eqref{eq:Omegadot}, as follows:
\begin{equation}
\vect{\mu}_{T,c} = \vect{J}\vect{\dot \Omega}_c + \vect{\Omega}_{LPF} \times \vect{J}\vect{\Omega}_{LPF}.
\end{equation}
Note that the aerodynamic moment and the rotor and propeller inertia are neglected in the inversion. The stochastic moment is assumed zero.

Next, the expressions for $\vect{f}_T$ and $\vect{\mu}_T$ in respectively \eqref{eq:ft} and \eqref{eq:mut} are inverted to obtain the commanded motor speeds. For a quadcopter in standard configuration, this amounts to inversion of a constant full-rank 4-by-4 matrix, as follows:
\begin{multline}
\left[\begin{array}{c} 
\omega_{1,c}|\omega_{1,c}|\\
\omega_{2,c}|\omega_{2,c}|\\
\omega_{3,c}|\omega_{3,c}|\\
\omega_{4,c}|\omega_{4,c}|
\end{array} \right] = 
\\
\left[ \begin{array}{c c c c}
k_{\vect{f}_T} l&  k_{\vect{f}_T} l&  -k_{\vect{f}_T} l&-  k_{\vect{f}_T} l\\
-k_{\vect{f}_T} l&  k_{\vect{f}_T} l&  k_{\vect{f}_T} l& - k_{\vect{f}_T} l\\
-k_{\mu_T} & k_{\mu_T} & -k_{\mu_T} & k_{\mu_T} \\
k_{\vect{f}_T} & k_{\vect{f}_T}& k_{\vect{f}_T}&k_{\vect{f}_T}
\end{array}
\right]^{-1}
\left[\begin{array}{c}
\\
\vect{\mu}_{T,c}\\
\\
\hline
T_c\end{array} \right],
\end{multline}
where $l$ is the moment arm from motor to vehicle center of gravity, i.e., $\vect{r}_i = \left[\begin{array}{c c c}
\pm l & \pm l & 0 \end{array}\right]^T$ for $i=1,2,3,4$. Note that again the rotor and propeller inertia is neglected.

\subsection{Car Dynamics}
FlightGoggles provides simulation of basic car dynamics.
It uses a Dubins vehicle-like dynamics model that takes the vehicle speed and yaw rate as inputs.

The two-dimensional position dynamics are given by
\begin{equation}
\mathbf{\dot p} = \mathbf{R}_\psi \left(\left[\begin{array}{c} 
u_v\\
0\end{array}\right] + \mathbf{w}_v \right),
\end{equation}
where $u_v$ is the commanded speed, $\mathbf{R}_\psi$ is the vehicle-to-world rotation matrix, and $\mathbf{w}_v$ is stochastic velocity process noise. The transformation matrix is computed based on the heading angle of which the dynamics are
\begin{equation}
\dot \psi = u_{\dot \psi},
\end{equation}
where $u_{\dot \psi}$ is the commanded yaw rate. The velocity noise is sampled according to
\begin{equation}
\mathbf{ w}_v  \sim \mathcal{N}(\mathbf{0},\frac{\mathbf{W}_{{v}}}{\Delta t}),
\end{equation}
where ${\Delta t}$ is the simulation time interval and $\mathbf{W}_v \delta(t)$ is the noise auto-correlation function (with $\delta(t)$ the Dirac delta function).

%% file: experiments.tex

\section{APPLICATIONS}
\label{sec:useCases}

\begin{figure*}[tbp]
	\centering
	\begin{subfigure}[b]{0.45\textwidth}
		\includegraphics[width=\textwidth]{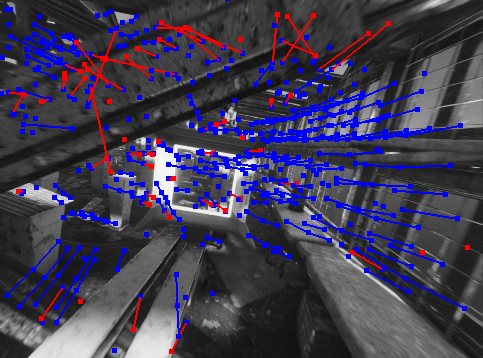}
	\end{subfigure}
	~
	\begin{subfigure}[b]{0.45\textwidth}
		\includegraphics[width=\textwidth]{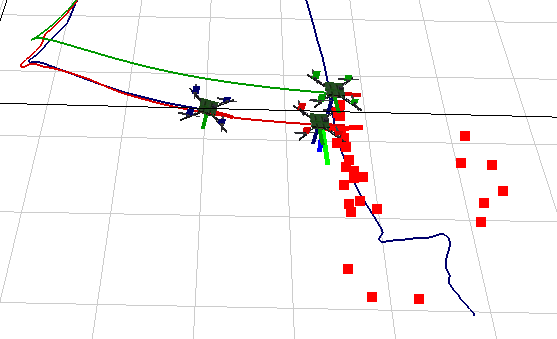}
	\end{subfigure}
	\caption {The figure on the left shows the visual features tracked using a typical visual inertial odometry pipeline on the simulated camera imagery. On the right, the plot shows three drones, the ground truth trajectory is in green, the high-rate estimate is in red, and the smoothed estimate is in blue. The red squares indicate triangulated features in the environment.}
	\label{fig:VIO_experiment}
\end{figure*}

In this section, we discuss application that FlightGoggles has been used for and potential future applications.
Potential applications of FlightGoggles include: human-vehicle interaction, active sensor selection, multi-agent systems, and visual inertial navigation research for fast and agile vehicles \cite{sayremccord2018visual, antonini2018blackbird, murali2019perception}.
The FlightGoggles simulator was used for the simulation part of the AlphaPilot challenge \cite{alphapilot}.

\subsection{Aircraft-in-the-Loop High-Speed Flight using Visual Inertial Odometry}

Camera-IMU sensor packages are widely used in both commercial and research applications, because of their relatively low cost and low weight. 
Particularly in GPS-denied environments, cameras may be essential for effective state estimation.
Visual inertial odometry (VIO) algorithms combine camera images with preintegrated IMU measurements to estimate the vehicle state~\cite{sayremccord2018visual,forster2015manifold}. 
While these algorithms are often critical for safe navigation, it is challenging to verify their performance in varying conditions. 
Environment variables, \eg, lighting and object placement, and camera properties may significantly affect performance, but generally cannot easily be varied in reality. 
For example, to show robustness in visual simultaneous localization and mapping may require data collected at different times of day or even across seasons~\cite{beall2014appearance, alcantarilla2018street}.
Moreover, obstacle-rich environments may increase the risk of collisions, especially in high-speed flight, further increasing the cost of extensive experiments.

FlightGoggles allows us to change environment and camera parameters and thereby enables us to quickly verify VIO performance over a multitude of scenarios, without the risk of actual collisions. 
By connecting FlightGoggles to a motion capture room with a real quadcopter in flight, we are able to combine its photo-realistic rendering with true flight dynamics and inertial measurements. 
This alleviates the necessity of complicated models including unsteady aerodynamics, and the effects of vehicle vibrations on IMU measurements.

Figure \ref{fig:VIO_experiment} gives an overview of a VIO flight in FlightGoggles. 
The quadcopter uses the trajectory tracking controller described in \cite{cdc2018indi} to track a predefined trajectory that was generated using methods from \cite{richter2016polynomial}. 
State estimation is based entirely on the pose estimate from VIO, which is using the virtual imagery from FlightGoggles and real inertial measurements from the quadcopter. 
In what follows, we briefly describe two experiments where the FlightGoggles simulator was used to verify developed algorithms for quadcopter state estimation and planning. 

\subsubsection {Visual inertial odometry development}
Sayre-McCord et al.~\cite{sayremccord2018visual} performed experiments using the FlightGoggles simulator in 2 scenarios to verify the use of  a simulator to perform the development of VIO algorithms using real-time exteroceptive camera simulation with aircraft-in-the-loop.
For the baseline experiment, they flew the quadcopter through a window without the assistance of a motion capture system first using a on-board camera and then using the simulated imagery from FlightGoggles.
Their experiments show that the estimation error of the developed VIO algorithm for the live camera is comparable to the simulated camera.

\subsubsection {Vision-aware planning}
During the performance of agile maneuvers by a quadcopter, visually salient features in the environment are often lost due to the limited field of view of the on-board camera.
This can significantly degrade the estimation accuracy.
To address this issue, \cite{murali2019perception} presented an approach to incorporate the perception objective of keeping salient features in view in the quadcopter trajectory planning problem.
This work was extended by \cite{spasojevic2020PATOPP} which plans time optimal trajectories while keeping chosen features in the field of view. 
The FlightGoggles simulation environment was used to perform experiments for these approaches. It allowed rapid experimentation with feature-rich objects in varying amounts and locations.
The experiments show that significant gains in estimation performance can be achieved by using the proposed vision aware planning algorithm as the speed of the quadcopter is increased.
These works left the problem of selecting the features open which was addressed in \cite{spasojevic2020feature}.
FlightGoggles enables the rapid prototyping of different camera modalities and rapid forward simulation of trajectories which makes experimentation easier.
For details, we refer the reader to \cite{murali2019perception, spasojevic2020PATOPP,spasojevic2020feature}.

\subsection{Collecting data using FlightGoggles Arcade}
FlightGoggles Arcade is a game developed using the FlightGoggles assets and provides a method to allow skilled players to enjoy FlightGoggles as a game while simultaneously generating data for applications such as imitation learning. 
The FlightGoggles Arcade is distributed as a prepackaged binary for ease of use and stores flown trajectories to the cloud as comma separated value files that can be easily accessed and used. 
There are 5 courses in the Ground Floor environment, 2 in the Basement environment and 1 course (the AlphaPilot course described in \ref{sec:AlphaPilot}) in the game at the time of writing.

\begin{figure}
\includegraphics[width=0.45\textwidth]{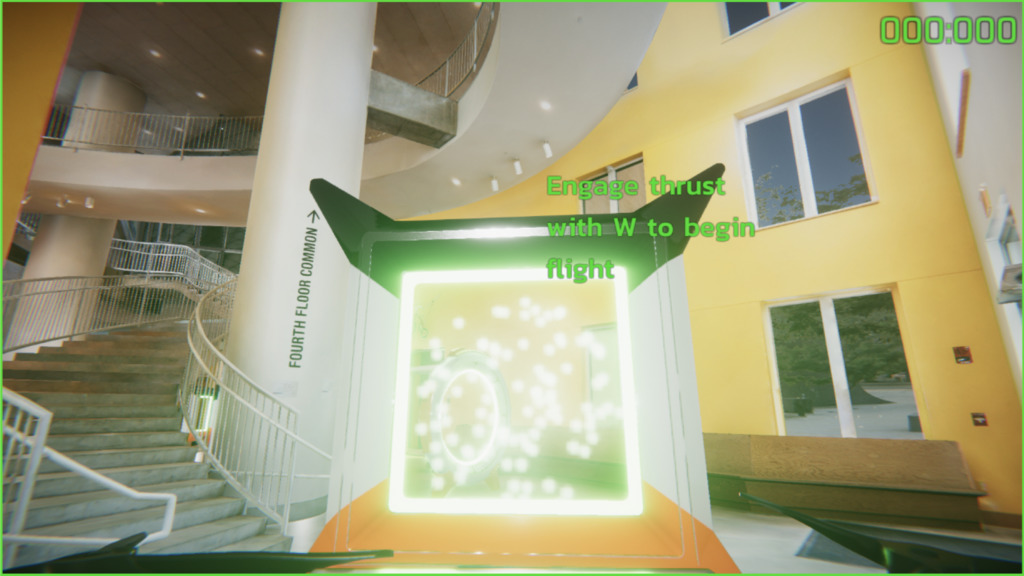}
\end{figure}

\begin{figure*}
\centering
\begin{subfigure}[t]{0.9\textwidth}
\includegraphics[width=\textwidth]{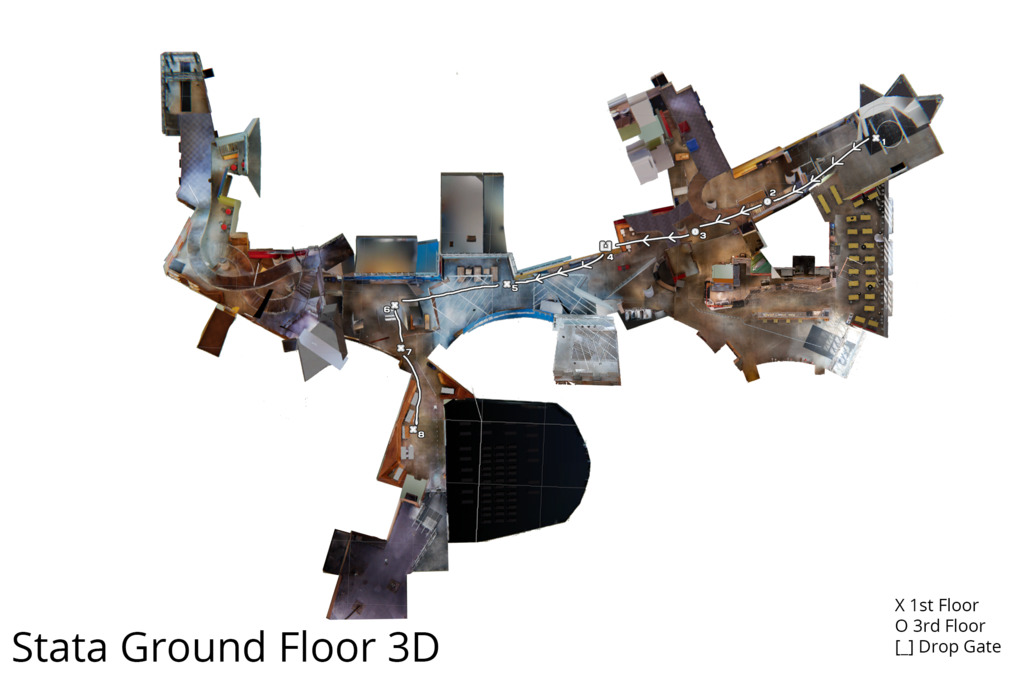}
\end{subfigure}
\caption{Example course layout in the FlightGoggles Arcade.}
\end{figure*}

\subsection{Interactions with Dynamic Actors}
\label{sec:arvr}

\begin{figure*}[tbp]
	\centering
	\begin{subfigure}[b]{0.3\textwidth}
		\includegraphics[width=\textwidth, trim={7.0cm 0.2 6.5cm 1.85cm}, clip]{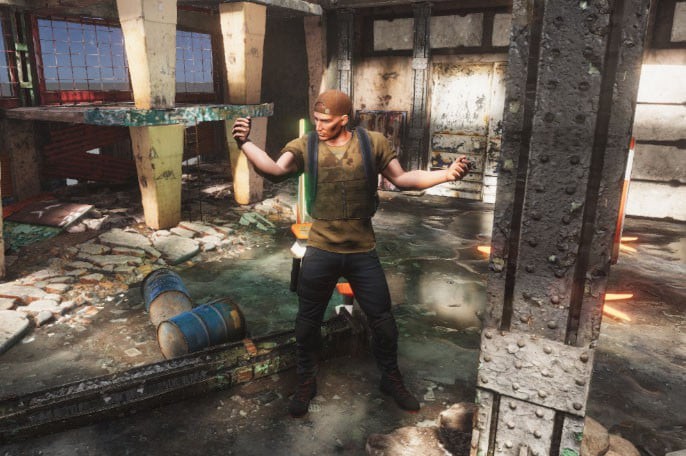}
	\end{subfigure}
	~
	\begin{subfigure}[b]{0.3\textwidth}
		\includegraphics[width=\textwidth, trim={2cm 1cm 9.2cm 4.6cm}, clip]{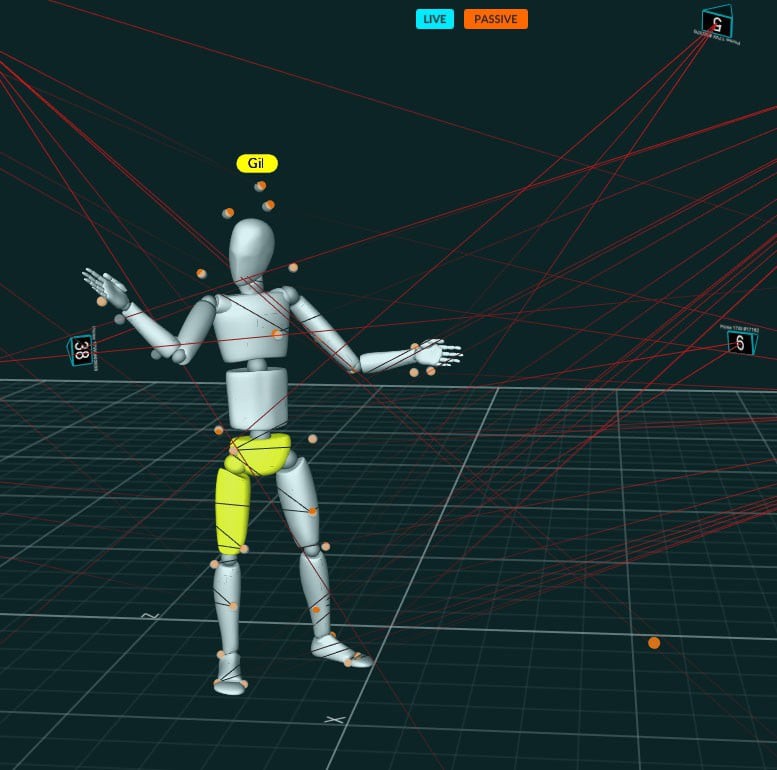}
	\end{subfigure}
	~
	\begin{subfigure}[b]{0.3\textwidth}
		\includegraphics[width=\textwidth]{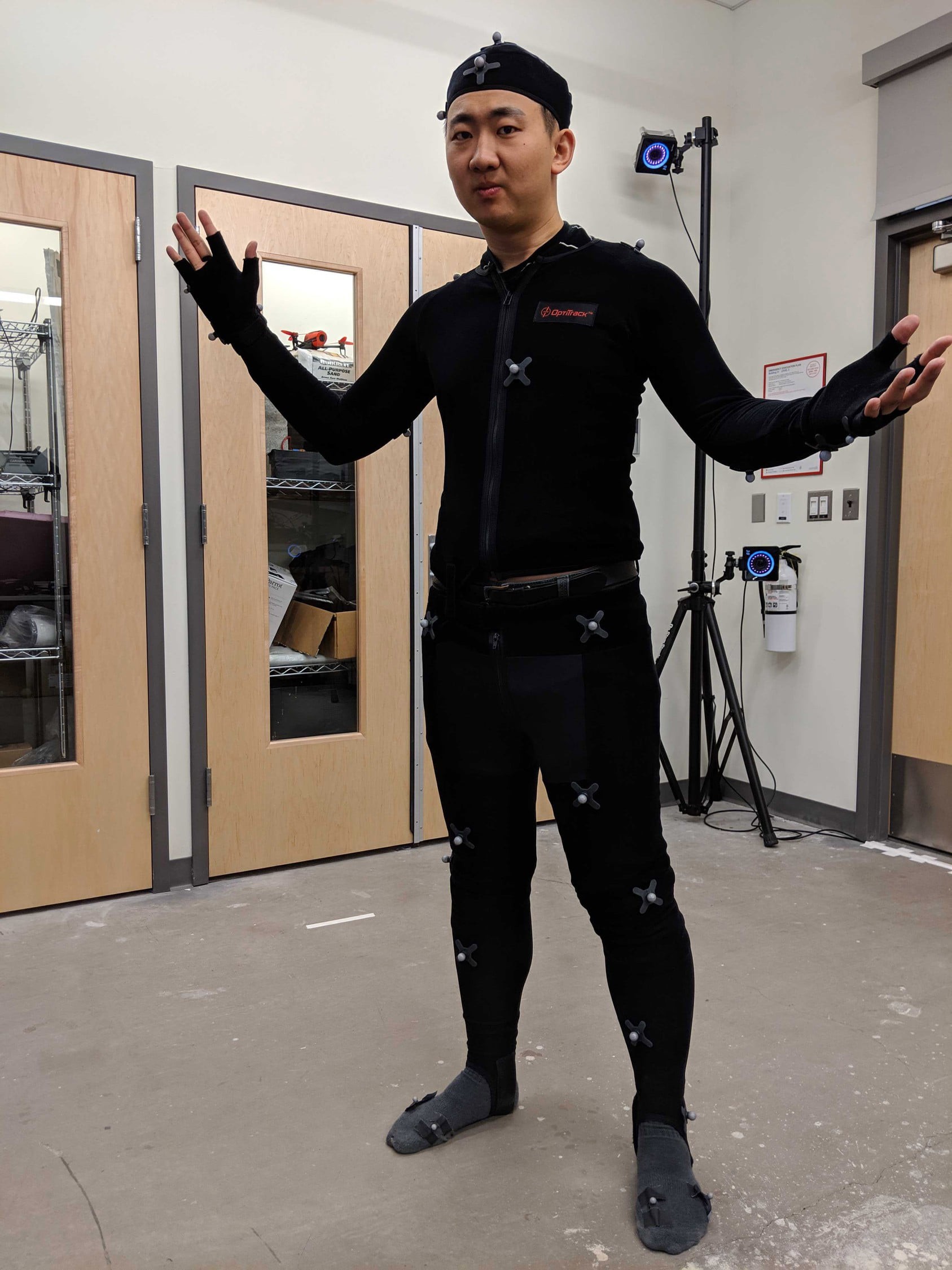}
	\end{subfigure}
	\caption {A dynamic human actor in the FlightGoggles virtual environment is rendered in real-time, based on skeleton tracking data of a human in a motion capture suit with markers.}
	\label{fig:human_render}
\end{figure*}

\begin{figure*}[tbp]
\centering
\includegraphics[width=\textwidth]{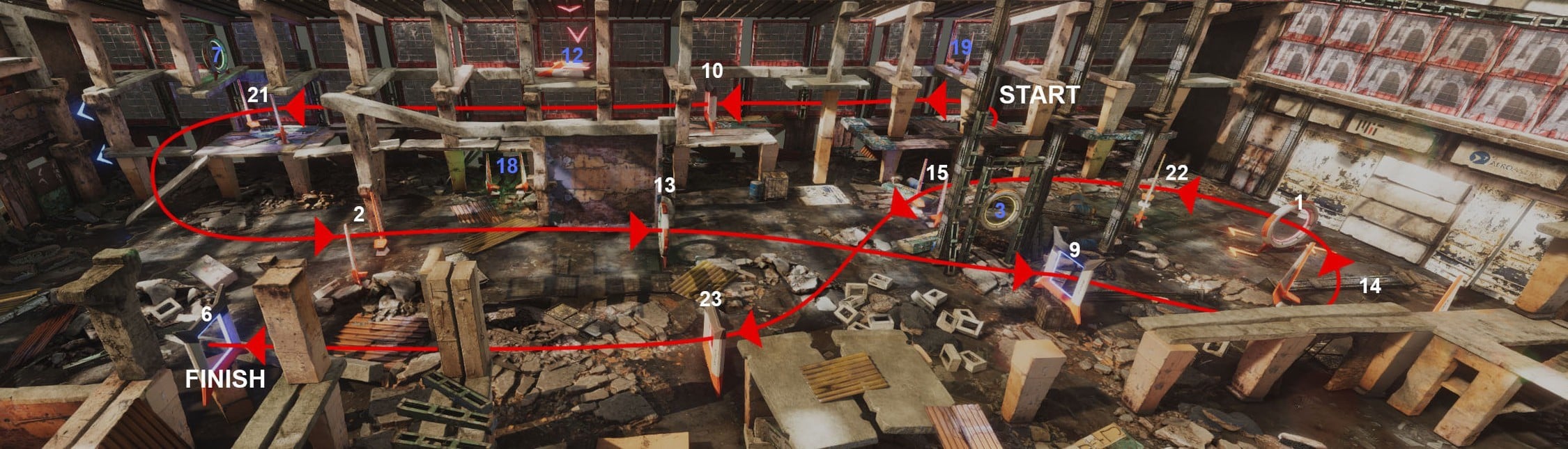}
\caption{Racecourse layout for the AlphaPilot simulation challenge. Gates along the racecourse have unique IDs labeled in white. Gate IDs in blue are static and not part of the race. The racecourse has 11 gates, with a total length of $\sim$240m.}
\label{fig:racecourse}
\end{figure*}

FlightGoggles is able to render dynamic actors, \eg, humans or vehicles, in real time from real-world models with ground-truth movement.
Figure \ref{fig:human_render} gives an overview of a simulation scenario involving a human actor.
In this scenario, the human is rendered in real time based on skeleton tracking motion capture data, while 
a quadcopter is simultaneously flying in a separate motion capture room.
While both dynamic actors (\ie\, human and quadcopter) are physically in separate spaces, they are both in the same virtual FlightGoggles environment.
Consequently, both actors are visible to each other and can interact through simulated camera imagery. This imagery can for example be displayed on virtual reality goggles, or used in vision-based autonomy algorithms.
FlightGoggles provides the capability to simulate these realistic and versatile human-vehicle interactions in an inherently safe manner.

\input{ar.tex}

%% file: ar.tex
\subsection{Augmented Reality}
\label{sec:ar}

\begin{figure*}
\centering
~
\begin{subfigure}[t]{0.3\textwidth}
\includegraphics[width=\textwidth]{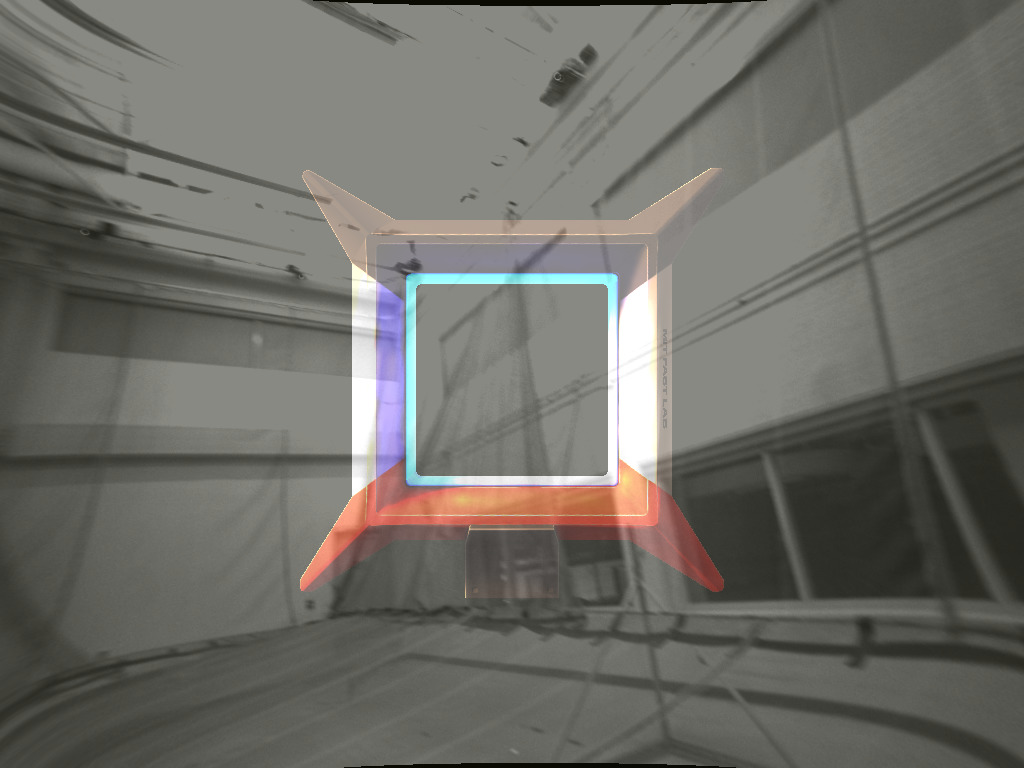}
\caption{No compensation.}
\end{subfigure}
~
\begin{subfigure}[t]{0.3\textwidth}
\includegraphics[width=\textwidth]{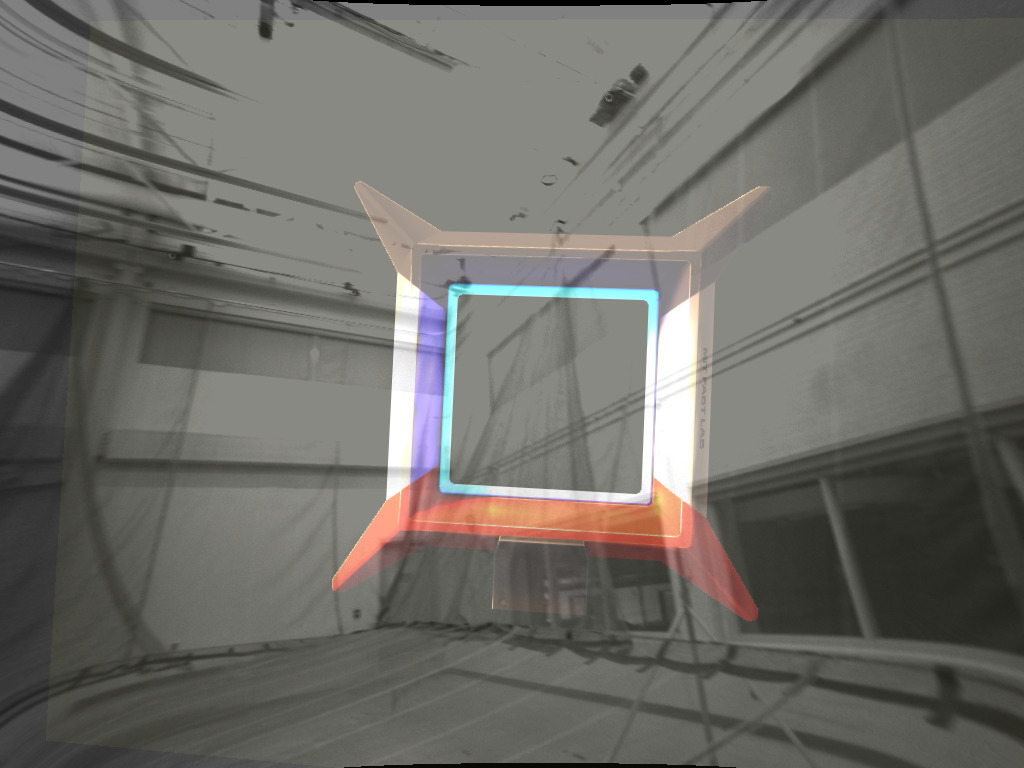}
\caption{56ms delay.}
\end{subfigure}
~
\begin{subfigure}[t]{0.3\textwidth}
\includegraphics[width=\textwidth]{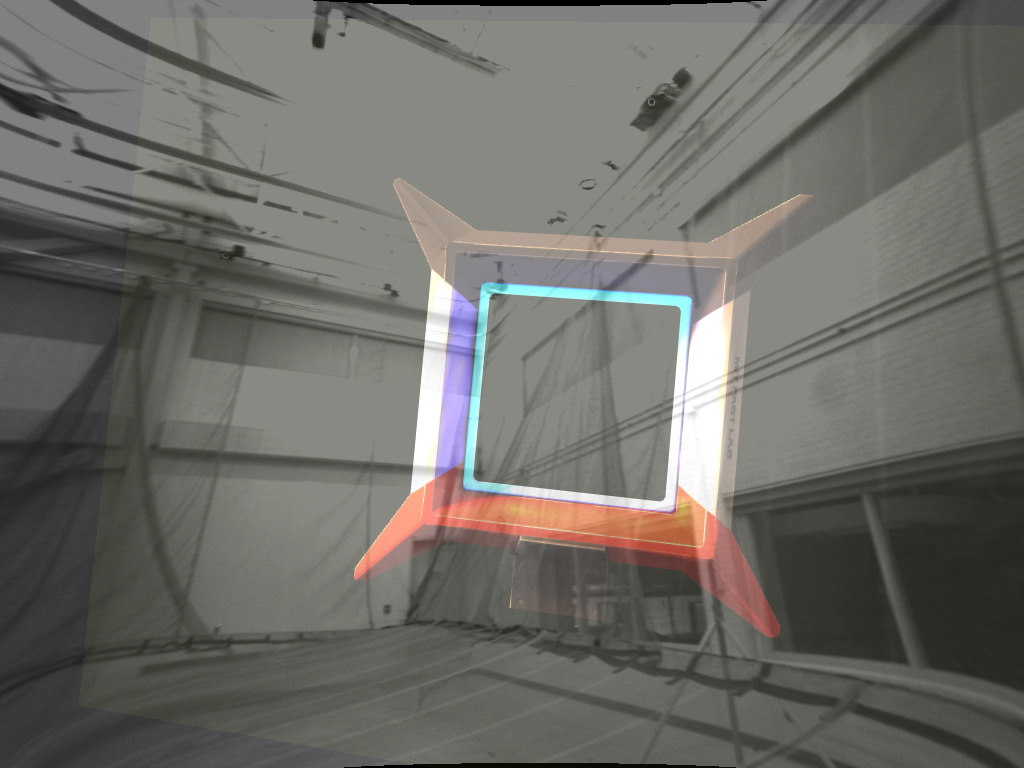}
\caption{112ms delay.}
\end{subfigure}
~
\begin{subfigure}[t]{0.3\textwidth}
\includegraphics[width=\textwidth]{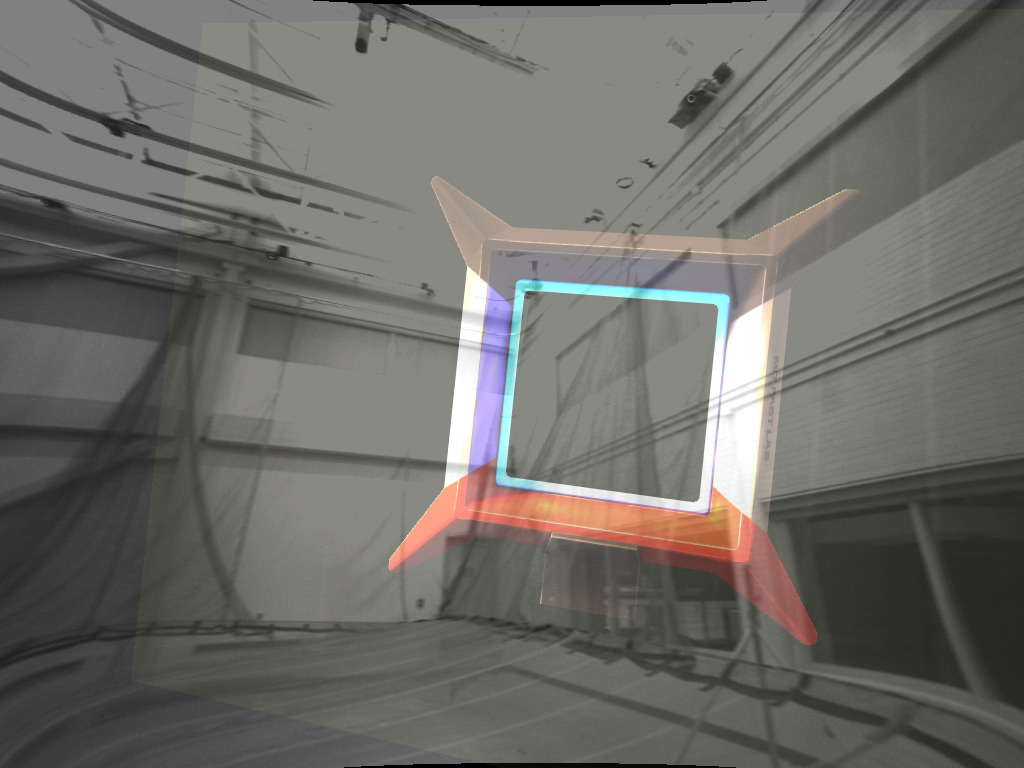}
\caption{168ms delay.}
\end{subfigure}
~
\begin{subfigure}[t]{0.3\textwidth}
\includegraphics[width=\textwidth]{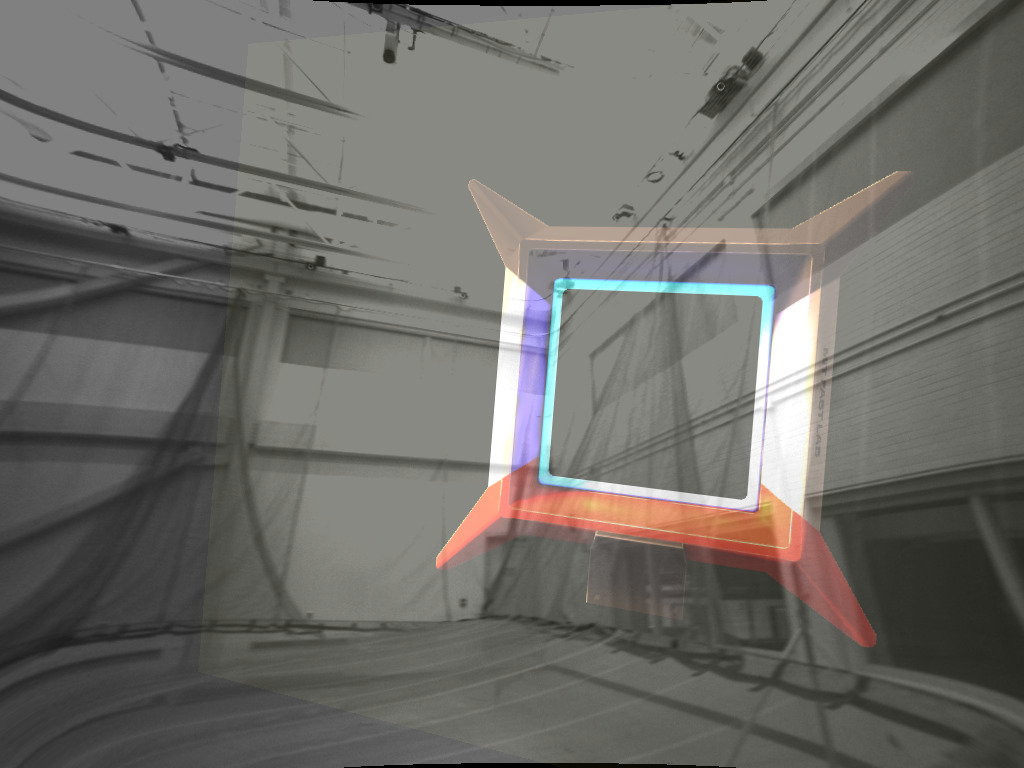}
\caption{224ms delay.}
\end{subfigure}
~
\begin{subfigure}[t]{0.3\textwidth}
\includegraphics[width=\textwidth]{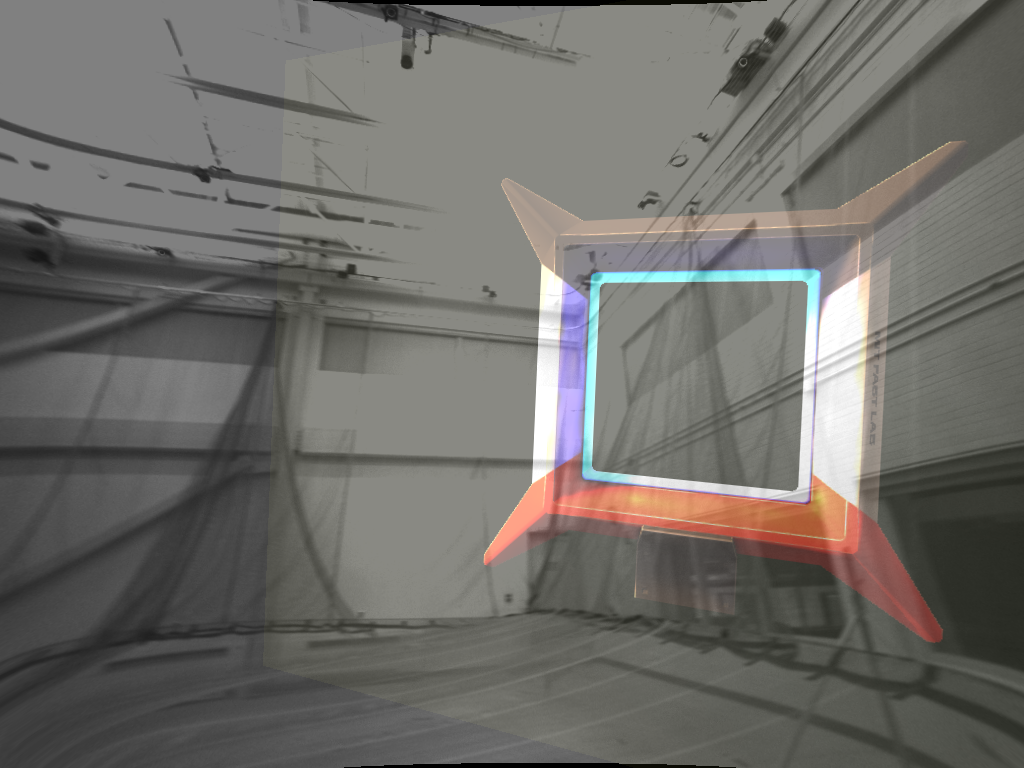}
\caption{280ms delay.}
\end{subfigure}
\caption{Augmented reality rendering of the gate in real imagery with delay compensation using homography estimated from the inertial measurement unit.}
\label{fig:latency_compensation}
\end{figure*}

In this section we describe the augemented reality applications from FlightGoggles. 
One of the major challenges of augmented reality simulation in a hardware-in-the-loop simulation is the latency between rendering and the availabilty of the image on the platform. 
Even in a desktop setting this latency is not insignificant since rendering could be more than a few frames behind a high-rate camera. 
To address this issue, we use the IMU available on the platform to compute a local difference in translation and rotation to compute the forward projection of the image. 
We assume that the object is at a nominal distance $d$ and the homography can be computed as 
\[
H = K \cdot ( R^T   + \frac{1}{d} \cdot t \cdot e_{3}^T) \cdot K^{-1}
\]
where $R$ is the rotation between the render timestamp and the current time, $t$ is the translation and $K$ is the camera matrix.  
The results of this forward simulation is shown in Fig \ref{fig:latency_compensation}.
For details we refer the reader to \cite{guerra2019photorealistic}.

%% file: alphapilot.tex

\subsection{AlphaPilot Challenge}
\label{sec:AlphaPilot}
The AlphaPilot challenge \cite{alphapilot} is an autonomous drone racing challenge organized by Lockheed Martin, NVIDIA, and the Drone Racing League (DRL).
The challenge is split into two stages, with a simulation phase open to the general public and a real-world phase where nine teams selected from the simulation phase will compete against each other by programming fully-autonomous racing drones built by the DRL. 
The FlightGoggles simulation framework was used as the main qualifying test in the simulation phase for selecting nine teams that would progress to the next stage of the AlphaPilot challenge. 
To complete the test, contestants had to submit code to Lockheed Martin that could autonomously pilot a simulated quadcopter with simulated sensors through the 11-gate race track shown in Figure \ref{fig:racecourse} as fast as possible.
Test details were revealed to all contestants on February 14th, 2019 and final submissions were due on March 20th, 2019. 
This section describes the AlphaPilot qualifying test and provides an analysis of anonymized submissions. 

\begin{table*}[p]                                                                                    
\centering                                                                                 
\begin{tabular}{|*{4}{c}|c|*{3}{c}|*{3}{c}|*{3}{c}|*{6}{R}|}                                                          
\toprule
\multicolumn{4}{|c|}{Sensors}  & &\multicolumn{3}{c|}{Estimation} & \multicolumn{3}{c|}{Planning} & \multicolumn{3}{c|}{Control} & & & & & & \\  
\hline
\rotatebox{90}{\parbox{2cm}{Camera}} &  \rotatebox{90}{\parbox{2cm}{IMU}}     & \rotatebox{90}{\parbox{2cm}{Ranger}}     &\rotatebox{90}{\parbox{2cm}{Infra-Red}}  & \rotatebox{90}{\parbox{2cm}{Learning}}  &  \rotatebox{90}{\parbox{2cm}{VIO}}  & \rotatebox{90}{\parbox{2cm}{Filter}}  &\rotatebox{90}{\parbox{2cm}{Smoother}}  & \rotatebox{90}{\parbox{2cm}{Polynomial}}  & \rotatebox{90}{\parbox{2cm}{Visual Servo}} & \rotatebox{90}{\parbox{2cm}{Other}}  &\rotatebox{90}{\parbox{2cm}{Linear}}  & \rotatebox{90}{\parbox{2cm}{MPC}} & \rotatebox{90}{\parbox{2cm}{Other}}  & Score & Score 1 & Score 2 & Score 3 & Score 4 & Score 5 \\
\midrule                                                                                      
 & \check & & \check & & & \check&  & & & \check& & \check & & 91.391 & 91.523 & 91.495 & 91.377 & 91.315 & 91.244  \\ 
 & \check & & \check & & &  & \check& \check& & &\check  & & \check &84.517 & 85.354 & 84.624 & 84.326 & 84.186 & 84.096  \\  
  & \check & & \check & & & \check & &\check & & &  & \check & & 81.044 & 81.468 & 81.048 & 80.936 & 80.922 & 80.848  \\  
   & \check & & \check & & & \check && &\check & & \check &  & &80.560 & 80.993 & 80.859 & 80.395 & 80.294 & 80.257  \\  
 \check & \check & & \check & & \check & \check && & & \check & &\check & &78.613 & 78.777 & 78.645 & 78.562 & 78.549 & 78.531  \\ 
  & \check &  & \check & & & \check & &\check & & && & \check & 78.552 & 78.693 & 78.592 & 78.500 & 78.497 & 78.478  \\  
\check & \check &  & \check & & \check & \check & & & & & &\check & & 76.078 & 76.595 & 76.173 & 75.947 & 75.902 & 75.774  \\ 
  & \check & & \check & & & \check & &\check & & &  \check& & &74.225 & 74.509 & 74.173 & 74.168 & 74.142 & 74.131 \\  
\check & \check & & \check & & \check & \check & & &\check & & \check& & & 71.443 & 71.503 & 71.471 & 71.446 & 71.437 & 71.359  \\ 
 & \check & & \check & & & \check & & \check & & & & & \check & 71.096 & 71.278 & 71.101 & 71.074 & 71.024 & 71.002  \\  
 & \check & \check & \check & & & \check & &  & \check & & \check&  & &  70.873 & 73.580 & 72.797 & 72.784 & 72.704 & 62.497  \\ 
\check & \check & \check & \check & \check & & \check & & & &\check &\check&  & &  70.456 & 71.025 & 70.791 & 70.222 & 70.211 & 70.032  \\
  & \check & & \check & & & \check & & & & \check& \check&  & & 69.908 & 71.419 & 70.699 & 69.294 & 69.167 & 68.962  \\  
& \check & \check & \check & \check & & & & & & &   & & &57.264 & 76.958 & 76.345 & 66.540 & 66.477 & 0.000  \\   
& \check & & \check & & &\check & & \check & & &  \check&  & &56.275 & 56.484 & 56.359 & 56.206 & 56.169 & 56.156  \\  
& \check & \check & \check & & & \check & & & \check & & \check&  & & 55.875 & 57.496 & 56.152 & 55.708 & 55.289 & 54.731 \\ 
\check & \check & & \check & & \check & \check & & \check & & & \check&  & &29.843 & 74.758 & 74.457 & 0.000 & 0.000 & 0.000 \\     
& \check & & \check & & & \check & & &\check & &  \check&  & &12.986 & 25.192 & 19.887 & 19.853 & 0.000 & 0.000 \\     
 & \check & & \check & & & \check & & &\check & & \check&  & &12.576 & 41.047 & 21.835 & 0.000 & 0.000 & 0.000 \\      
 & \check &  & \check & & & \check & &  & \check & & \check&  & &11.814 & 59.068 & 0.000 & 0.000 & 0.000 & 0.000  \\  
\bottomrule                                                                                          
\end{tabular}                                                                               
\caption{Sensor usage, algorithm choices and five highest scores in AlphaPilot simulation challenge.}
\label{table:AlphaPilotScores}
\end{table*} 

\begin{figure}[!tbp]
	\centering
	\begin{subfigure}[b]{\linewidth}
		\centering
		\includegraphics[width=0.95\linewidth]{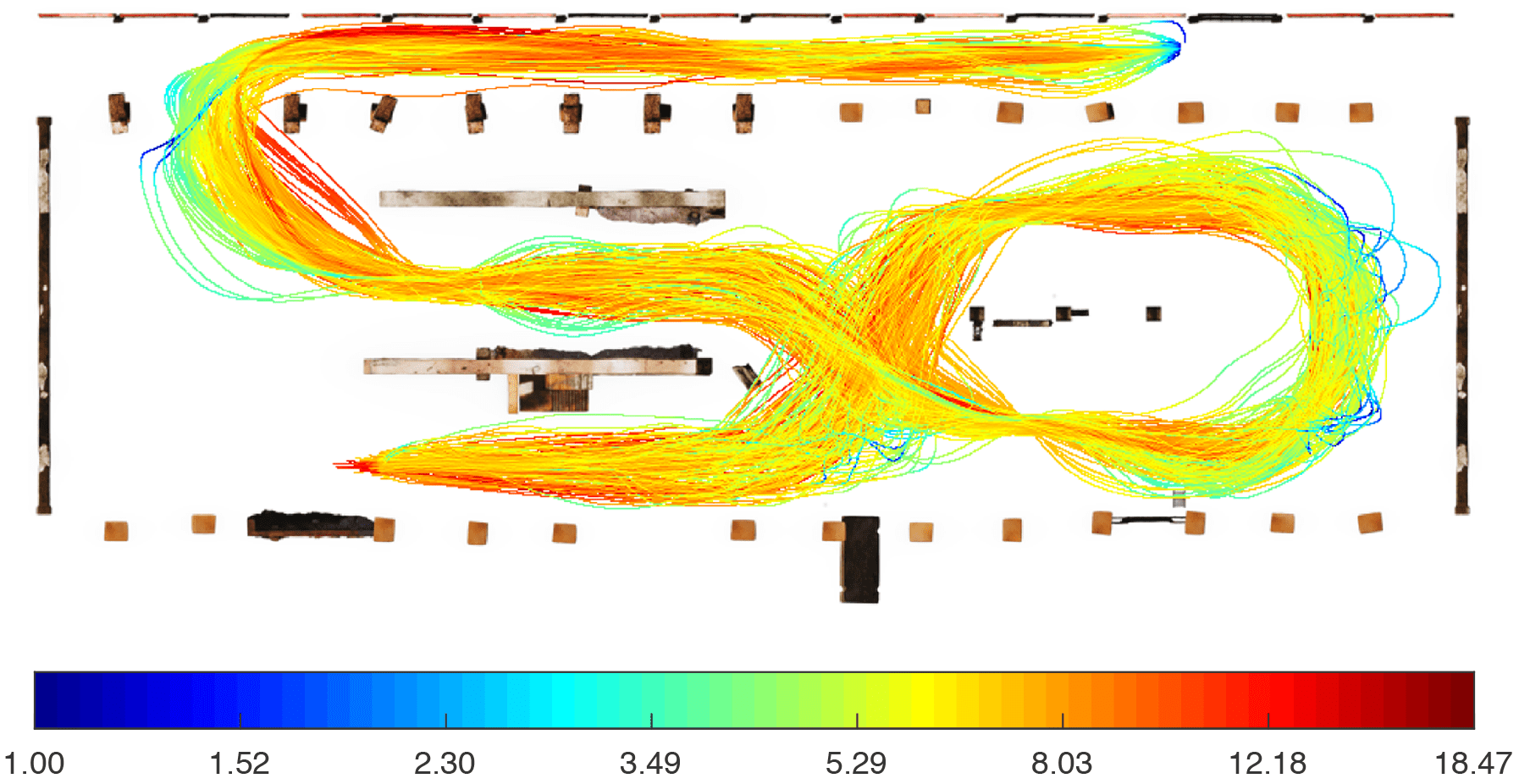}
		\caption{Visualization of speed profiles across \textit{successful} runs.}
		\label{fig:speedmap_sucessful}
	\end{subfigure}
	~
	\begin{subfigure}[b]{\linewidth}
		\centering
		\includegraphics[width=0.95\linewidth]{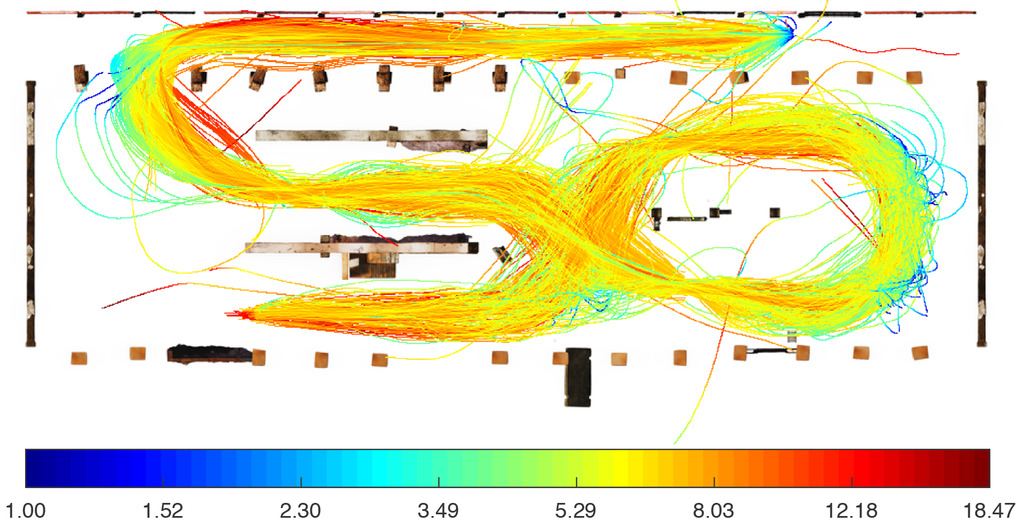}
		\caption{Visualization of speed profiles across \textit{all} runs.}
		\label{fig:speedmap}
	\end{subfigure}
	~
	\begin{subfigure}[b]{\linewidth}
		\centering
		\includegraphics[trim={0cm 0cm 0.75cm 0cm}, clip,width=0.95\linewidth]{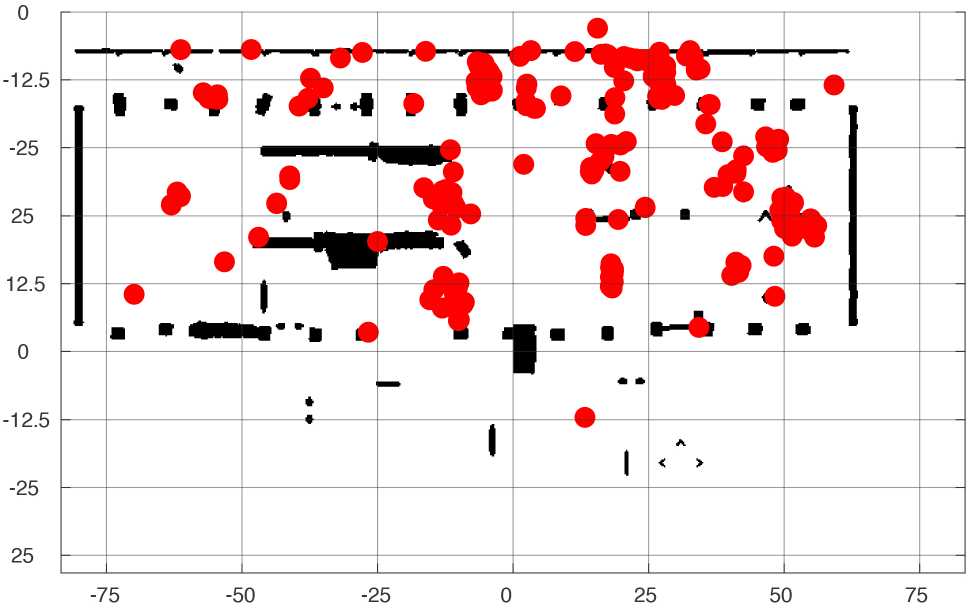}
		\caption{Crash locations for all teams across all runs. Note that most crashes occur near gates, obstacles, or immediately post-takeoff.}
		\label{fig:crashMap}
	\end{subfigure}
\caption{Overhead visualizations of speed profiles and crash locations for top 20 AlphaPilot teams across all 25 runs.}
\label{fig:overheadMaps}
\end{figure}

\begin{table*}[tbp]
\centering
\begin{tabular}{ccccc}
\toprule
Sensor Package Selection & Number of Teams & Completed Runs ($\%$) & Mean Score & Std. Deviation \\ 
\midrule
IMU + IR & 12 & 48.67 & 35.32 & 37.72  \\
IMU + IR + Camera & 4  & 36 & 26.72 &  35.87 \\
IMU + IR + Ranger & 3 & 24 & 15.04 & 27.43 \\ 
IMU + IR + Ranger + Camera & 1 & 60 & 41.39 & 34.55\\
\bottomrule
\end{tabular}
\caption {Sensor combinations used by the top AlphaPilot teams, percentage of completed runs, mean score and standard deviation across all 25 evaluation courses.}
\label{table:SensorUsageImpact}
\end{table*}

\begin{figure*}[h]
    \centering
    \begin{subfigure}[b]{0.45\textwidth}
        \includegraphics[trim={2.8cm 7.6cm 2.5cm 7.2cm}, clip, width=\textwidth]{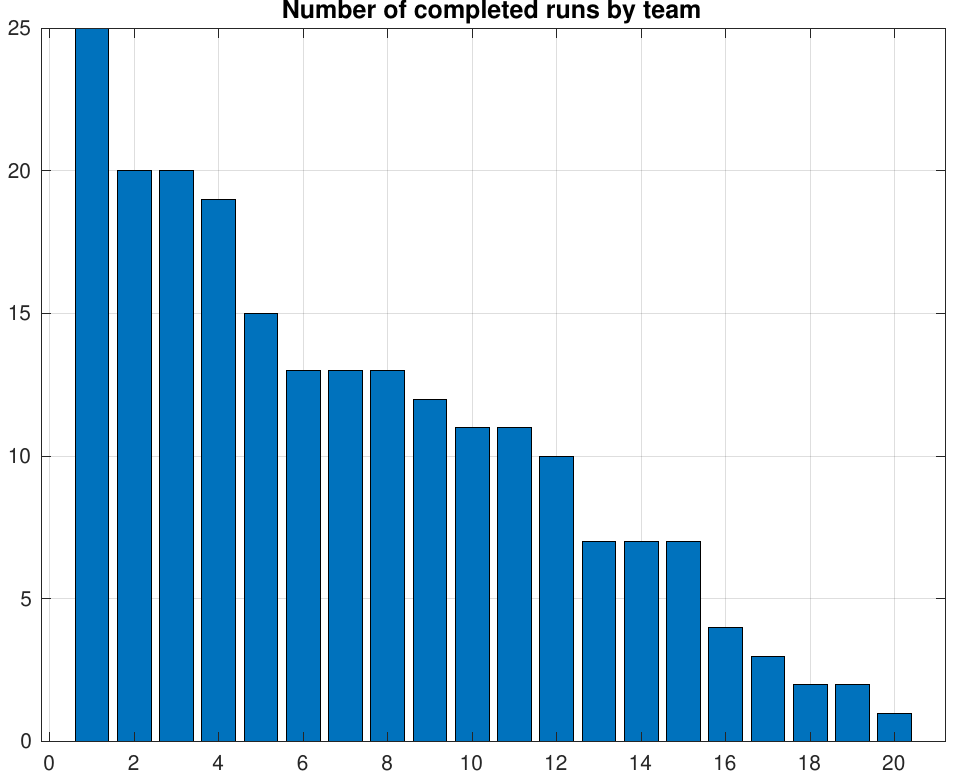}
        \caption{The number of completed runs for the top 20 teams.}
        \label{fig:sensorPkgs}
    \end{subfigure}
    ~
    \begin{subfigure}[b]{0.45\textwidth}
        \includegraphics[trim={2.8cm 7.6cm 2.5cm 7.2cm}, clip,width=\textwidth]{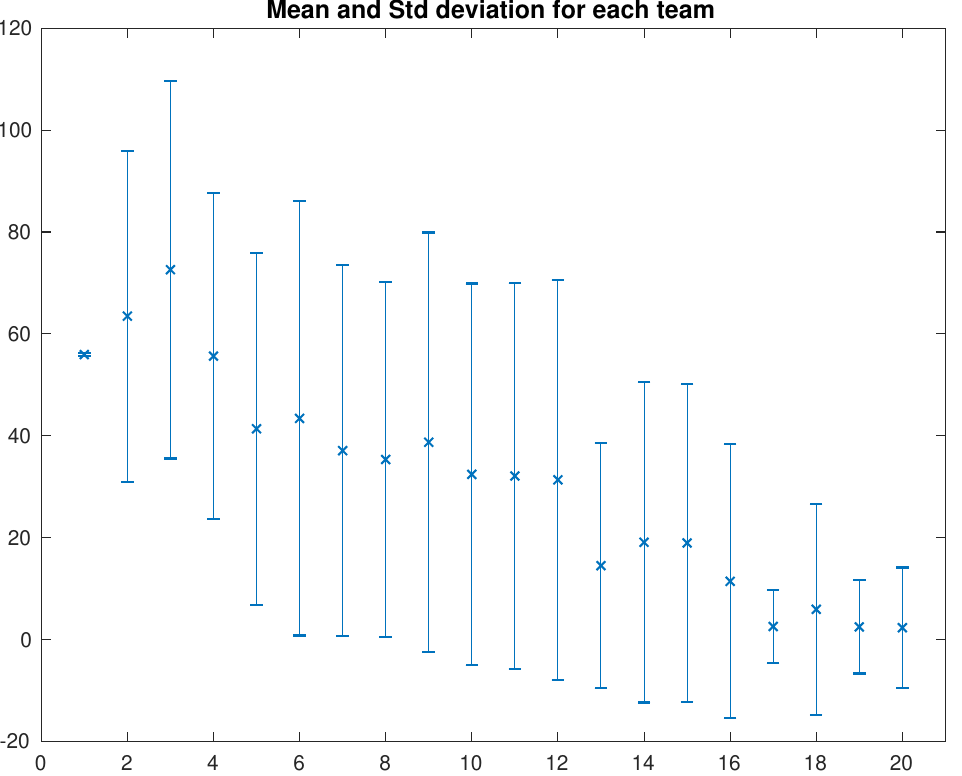}
        \caption{The mean and standard deviation of scores.}
        \label{fig:sensorErrorBars}
    \end{subfigure}
\caption{The figures above show the number of completed runs by each of the top 20 AlphaPilot teams along with the mean and standard deviation of their scores for all $25$ runs.}
\label{fig:individualPerformance}
\end{figure*}

\subsubsection{Challenge outline}
The purpose of the AlphaPilot simulation challenge was for teams
to demonstrate their autonomous guidance, navigation, and control capability in a realistic simulation environment.
The participants' aim was to make a simulated quadcopter based on the multicopter dynamics model described in Section \ref{sec:sim_vehicles}
complete the track as fast as possible.
To accomplish this, measurements from four simulated sensors were provided:
(stereo) cameras, IMU, downward-facing time-of-flight range sensor, and infrared gate beacons.
Through the FlightGoggles ROS API, autonomous systems could obtain sensor measurements and provide collective thrust and attitude rate inputs to the quadcopter's low-level acro/rate mode controller.

The race track was located in the FlightGoggles \textit{Abandoned Factory} environment and consisted of 11 gates.
To successfully complete the entire track, the quadcopter had to pass all gates in order.
Points were deducted for missed gates, leading to the following performance metric
\begin{equation}
score=10\cdot gates-time
\end{equation}
where $gates$ is the number of gates passed in order and $time$ is the time taken in seconds to reach the final gate.
If the final gate was not reached within the race time limit
or the quadcopter collided with an environment object,
a score of zero was recorded.
To discourage memorization of the course, there were 25 courses for each contestant to complete during evaluation.
For each course, the exact gate locations were subject to random unknown perturbations.
These perturbations were large enough to require adapting the vehicle trajectory,
but did not change the track layout in a fundamental way.
For development and verification of their algorithms,
participants were also provided with another set of 25 courses with
identically distributed gate locations.
The final score for the challenge was the mean of the five highest scores among the 25 evaluation courses.


\subsubsection{FlightGoggles sensor usage}

Table \ref{table:AlphaPilotScores} shows the usage of provided sensors, the algorithm choices, and final and five highest scores for the 20 top teams (sorted by final score). 
All of these 20 teams used both the simulated IMU sensor and the infrared beacon sensors. 
Several teams chose to also incorporate the camera and the time-of-flight range sensor.
A more detailed overview of the sensor combinations used by the teams is shown in Table \ref{table:SensorUsageImpact}. 
This table shows the number of teams that employed a particular combination of sensors, the percentage of runs completed, and the mean and standard deviation of the scores across all 25 attempts.

\subsubsection{Algorithm choices}

The contestants were tasked with developing guidance, navigation, and control algorithms.
Table \ref{table:AlphaPilotScores} tabulates the general estimation, planning, and control approaches used for each team alongside the sensor choices and their scores.
 
Of the top 20 teams, only one used an end-to-end learning-based method. 
The other 19 teams relied on more traditional pipelines (estimation, planning, and control) to complete the challenge.
One of those teams used learning to determine the pose of the camera from the image.

\textbf{Estimation:} For state estimation, all but one team used a filtering algorithm such as the extended Kalman filter \cite{smith1962ekf}, unscented Kalman filter \cite{julier1997ukf}, particle filter \cite{liu1998smc}, or the Madgwick filter \cite{madgwick2010efficient} with the other team using a smoothing based technique \cite{kaess2012isam2}.
The teams that chose to use a visual inertial odometry algorithm opted to use off-the-shelf solutions such as ROVIO \cite{bloesch2015robust, bloesch2017iterated} or VINS-Mono \cite{qin2018vins} for state estimation.

\textbf{Planning:} The most common methods used for planning involved visual servo using infrared beacons or polynomial trajectory planning such as \cite{mellinger2011minimum, richter2016polynomial}.
Other methods used for planning either used manually-defined waypoints or used sampling-based techniques for building trajectory libraries.
5 of the 19 teams to use model based techniques also incorporated some form of perception awareness to their planning algorithms.

\textbf{Control:} The predominant methods for control were linear control techniques and model predictive control \cite{kamel2017model}. 
The other algorithms that were used were geometric and backstepping control methods \cite{bouabdallah2005backstepping}.

\subsubsection{Analysis of trajectories}

To visualize the speed along the trajectory, we discretize the trajectories on a $x-y$ grid and the image is colored by the logarithm of the average speed in the grid.
Figure \ref{fig:speedmap_sucessful} shows the trajectories of all successful course traversals colored by the speed. 
From the figure, we can observe that most teams chose to slow down for the two sharp turns that are required. 
We can also observe that the the average speed around gates is lower than other portions of the environment which can be attributed to the need to search for the `next gate'.
Figure \ref{fig:speedmap} shows the trajectories of all course traversals including trajectories that eventually crash colored by the speed.
Figure \ref{fig:crashMap} shows the crash locations of all the failed attempts. 
Since many of the teams relied on visual-servo based techniques, the gate beacons are harder to observe close to the gates and many of the crash locations are close to gates.

\subsubsection{Individual performance of top teams}
For individual performance, we analyze the number of completed runs for each of the top teams and the mean and standard deviations of the scores. 
This is shown in Figure \ref{fig:individualPerformance}.
Given, that the scoring function for the competition only used the top $5$ scores, the teams were encouraged to take significant risk and only one team consistently completed all of the $25$ challenges provided successfully. 
As a result of the risk taking to achieve faster speeds, $75\%$ of the contestants failed to complete the challenge half the time. 
This risk taking behavior is also observed in the large standard deviation for all the teams.